\documentclass[11pt]{article} 
\usepackage{geometry}
\usepackage{float,amsmath, amssymb, graphicx, url, algorithm2e,amsthm,bm}

\usepackage[breaklinks=false,backref=false,colorlinks=false]{hyperref}
\usepackage{breakurl}

\floatstyle{ruled}
\newfloat{algorithm}{tbp}{loa}
\providecommand{\algorithmname}{Algorithm}
\floatname{algorithm}{\protect\algorithmname}

\global\long\def\x{\bm{x}}
\global\long\def\X{\bm{X}}
\global\long\def\Y{\bm{Y}}
\global\long\def\AA{\mathcal{A}}
\global\long\def\BB{\mathcal{B}}

\global\long\def\K{\bm{K}}
\global\long\def\B{\bm{B}}
\global\long\def\Z{\bm{Z}}
\global\long\def\A{\bm{A}}
\global\long\def\J{\bm{J}}

\global\long\def\D{\bm{D}}
\global\long\def\E{\bm{E}}

\global\long\def\g{\bm{g}}
\global\long\def\cc{\bm{c}}
\global\long\def\y{\bm{y}}
\global\long\def\z{\bm{z}}
\global\long\def\e{\bm{e}}

\global\long\def\dd{\bm{d}}

\global\long\def\u{\bm{u}}
\newcommand{\vv}{\ensuremath{\bm{v}}}
\global\long\def\w{\bm{w}}
\global\long\def\h{\bm{h}}
\global\long\def\H{\bm{H}}
\global\long\def\I{\bm{I}}
\global\long\def\U{\bm{U}}

\global\long\def\G{\bm{G}}

\global\long\def\V{\bm{V}}
\global\long\def\P{\bm{P}}

\global\long\def\zero{\bm{0}}

\global\long\def\aalpha{\bm{\alpha}}
\global\long\def\bbeta{\bm{\beta}}
\global\long\def\ddelta{\bm{\delta}}
\global\long\def\cchi{\bm{\chi}}

\global\long\def\ttheta{\bm{\theta}}
\global\long\def\ugamma{\underline{\gamma}}
\global\long\def\xxi{\bm{\xi}}
\global\long\def\eeps{\bm{\epsilon}}

\global\long\def\sech{\operatorname{sech}}

\newtheorem{theorem}{Theorem}
\newtheorem{corollary}{Corollary}
\newtheorem{lemma}{Lemma}

\newtheorem{remark}{Remark}

\title{A Convex Formulation for Mixed Regression with Two Components:  Minimax Optimal Rates}

\author{
Yudong Chen \\
{The University of California, Berkeley}\\
{yudong.chen@eecs.berkeley.edu}
\and
Xinyang Yi\\
{The University of Texas at Austin}\\
{yixy@utexas.edu}
\and
Constantine Caramanis\\
{The University of Texas at Austin}\\
{constantine@utexas.edu} }

\date{}

\begin{document}

\maketitle

\begin{abstract}
We consider the mixed regression problem with two components, under adversarial and stochastic noise. We give a convex optimization formulation that provably recovers the true solution, and provide upper bounds on the recovery errors for both arbitrary noise and stochastic noise settings. We also give {\em matching minimax lower bounds} (up to log factors), showing that under certain assumptions, our algorithm is information-theoretically optimal. Our results represent the first tractable algorithm guaranteeing successful recovery with tight bounds on recovery errors and sample complexity.
\end{abstract}

\section{Introduction}
\label{sec:intro}

This paper considers the problem of {\em mixed linear regression}, where the output variable we see comes from one of two unknown regressors. Thus we see data $(\x_i,y_i) \in \mathbb{R}^p\times \mathbb{R}$, where
$$
y_i = z_i \cdot \langle \x_i, \bbeta_1^* \rangle + (1-z_i) \cdot \langle \x_i, \bbeta_2^* \rangle + e_i, \quad i=1,\dots,n,
$$
where $z_i \in \{0,1\}$ can be thought of as a hidden label, and $ e_i $ is the noise. Given the label for each sample, the problem decomposes into two standard regression problems, and can be easily solved. Without it, however, the problem is significantly more difficult. 
The main challenge of mixture models, and in particular mixed regression falls in the intersection of the {\em statistical} and {\em computational} constraints: the problem is difficult when one cares both about an efficient algorithm, and about near-optimal ($n = O(p)$) sample complexity. Exponential-effort brute force search typically results in statistically near-optimal estimators; on the other hand, recent tensor-based methods give a polynomial-time algorithm, but at the cost of $O(p^6)$ sample complexity (recall $\bbeta^*_1,\bbeta^*_2 \in \mathbb{R}^p$) instead of the optimal rate, $O(p)$.\footnote{It should be possible to improve the tensor rates to $O(p^4)$ for the case of Gaussian design.}

The Expectation Maximization (EM) algorithm is computationally very efficient, and widely used in practice. However, its behavior is poorly understood, and in particular, no theoretical guarantees on global convergence are known.

\paragraph{Contributions.}
In this paper, we tackle both statistical and algorithmic objectives at once. The algorithms we give are efficient, specified by solutions of convex optimization problems; in the noiseless, arbitrary noise and stochastic noise regimes, they provide the best known sample complexity results; in the balanced case where nearly half the samples come from each of $\bbeta_1^{\ast}$ and $\bbeta_2^{\ast}$, we provide matching minimax lower bounds, showing our results are optimal.

Specifically, our contributions are as follows:
\begin{itemize}
\item In the arbitrary noise setting where the noise $ \e = (e_1, \ldots, e_n)^\top $ can be adversarial, we show that under certain technical conditions, as long as the number of observations for each regressor satisfy $ n_1,n_2 \gtrsim p $, our algorithm produces an estimator $ (\hat{\bbeta}_1, \hat{\bbeta}_2) $ which satisfies
\[
\Vert \hat{\bbeta}_b - \bbeta_b^* \Vert_2 \lesssim \frac{\Vert \e \Vert_2}{\sqrt{n}},\; b=1,2.
\]
Note that this immediately implies exact recovery in the noiseless case with $ O(p) $ samples. 

\item In the stochastic noise setting with sub-Gaussian noise and balanced labels, we show under the necessary assumption $ n_1,n_2 \gtrsim p $ and a Gaussian design matrix, our estimate satisfies the following (ignoring polylog factors):
\begin{align*}
\Vert \hat{\bbeta}_b - \bbeta_b^* \Vert_2  & \lesssim
\begin{cases}
\sigma\sqrt{\frac{p}{n}} , & \textrm{if } \gamma\ge\sigma,\\
\frac{\sigma^{2}}{\gamma}\sqrt{\frac{p}{n}}, & \textrm{if } \sigma\left(\frac{p}{n}\right)^{\frac{1}{4}}\le \gamma\le\sigma,\\
\sigma\left(\frac{p}{n}\right)^{\frac{1}{4}}, & \textrm{if } \gamma\le\sigma\left(\frac{p}{n}\right)^{\frac{1}{4}}
\end{cases}
\end{align*}
where $ b=1,2 $ and $ \gamma$ is any lower bound of $ \Vert \bbeta_1^* \Vert_2 + \Vert \bbeta_2^*\Vert_2 $ and $ \sigma^2 $ is the variance of the noise~$ e_i $. 
\item In both the arbitrary and stochastic noise settings, we provide minimax lower bounds that match the above upper bounds up to at most polylog factors, thus showing that the results obtained by our convex optimization solution are information-theoretically optimal. Particularly in the stochastic setting, the situation is a bit more subtle: the minimax rates in fact depend on the signal-to-noise and exhibit several phases, thus showing a qualitatively different behavior than in standard regression and many other parametric problems (for which the scaling is $ \sqrt{1/n} $). 
\end{itemize}

\section{Related Work and Contributions}
Mixture models and latent variable modeling are very broadly used in a wide array of contexts far beyond regression. Subspace clustering \cite{elhamifar2009sparse, soltanolkotabi2013robust,wang2013subspace}, Gaussian mixture models~\cite{Hsu2012Gussian,azizyan2013sparseMixture} and $k$-means clustering are popular examples of unsupervised learning for mixture models. 
The most popular and broadly implemented approach to mixture problems, including mixed regression, is the so-called Expectation-Maximization (EM) algorithm~\cite{dempster1977maximum,geoffrey2004finite}.
In fact, EM has been used for mixed regression for various application domains~\cite{Viele2002, grun2007applications}. 
Despite its wide use, still little is known about its performance beyond local convergence~\cite{wu1983convergence,balakrishnan2014EM}. 

One exception is the recent work in~\cite{YiCaramanisSanghavi2013}, which considers mixed regression in the noiseless setting, where they propose an alternating minimization approach initialized by a grid search and show that it recovers the regressors in the noiseless case with a sample complexity of $ O(p\log^2 p) $.  Extension to the noisy setting is very recently considered in~\cite{balakrishnan2014EM}. Focusing on the stochastic noise setting and the high SNR regime (i.e., when $ \gamma \gtrsim \sigma $; cf. Section~\ref{sec:intro}), they show that the EM algorithm with good initialization achieves the error bound $ \Vert \hat{\bbeta}_b - \bbeta_b^* \Vert_2  \lesssim \sqrt{\gamma^2 + \sigma^2} \sqrt{\frac{p}{n}}$. Another notable exception is the work in~\cite{Buhlmann2010}. There, EM is adapted to the high-dimensional sparse regression setting, where the regressors are known to be sparse. The authors use EM to solve a penalized (for sparsity) likelihood function. A generalized EM approach achieves support-recovery, though once restricted to that support where the problem becomes a standard mixed regression problem, only convergence to a local optimum can be guaranteed. 

Mixture models have been recently explored using the recently developed technology of tensors in~\cite{Anandkumar2012Tensor,Hsu2012Gussian}. In~\cite{chaganty13}, the authors consider a tensor-based approach, regressing $\mathbf{x}^{\otimes 3}$ against $y_i^3$, and then using the tensor decomposition techniques to efficiently recover each $\bbeta_b^*$. These methods are not limited to the mixture of only two models, as we are. Yet, the tensor approach requires $ O(p^6) $ samples, which is  several orders of magnitude more  than the $O(p \cdot \textrm{polylog}(p))$ that our work requires. As noted in their work, the higher sampling requirement for using third order tensors seems intrinsic.

In this work we consider the setting with two mixture components. Many interesting applications have binary latent factors: gene mutation present/not, gender, healthy/sick individual, children/adult, etc.; see also the examples in~\cite{Viele2002}. Theoretically, the minimax rate was previously unknown even in the two-component case. Extension to more than two components is of great interest.

Finally, we note that our focus is on estimating the regressors $ (\bbeta_1^*, \bbeta_2^*) $ rather than identifying the hidden labels $ \{z_i\} $ or predicting the response $ y_i $ for future data points. The relationship between covariates and response is often equally (some times more) important as prediction. For example, the regressors may correspond to unknown signals or molecular structures, and the response-covariate pairs are linear measurements; here the regressors are themselves the object of interest. For many mixture problems, including clustering, identifying the labels accurately for all data points may be (statistically) impossible. Obtaining the regressors allows for an estimate of this label (see~\cite{sun2013classifier} for a related setting).

\section{Main Results}
In this section we present this paper's main results. In addition, we present the precise setup and assumptions, and introduce the basic notation we use.

\subsection{Problem Set Up}\label{sec:setup}
Suppose there are two unknown vectors $\bbeta_{1}^{*}$ and $\bbeta_{2}^{*}$ in $\mathbb{R}^{p}$. We observe $n$ \emph{noisy} linear measurements $\{(\x_{i},y_{i})\}_{i=1}^{n}$ which satisfy the following: for $b\in\{1,2\}$ and $i\in\mathcal{I}_{b}\subseteq[n]$, 
\begin{equation}
y_{i}=\left\langle \x_{i},\bbeta_{b}^{*}\right\rangle +e_{i},\label{eq:observation}
\end{equation}
where $\mathcal{I}_1$ with $n_{1}=\left|\mathcal{I}_{1}\right|$ and $\mathcal{I}_2$ with $n_{2}=\left|\mathcal{I}_{2}\right|$ denote the subsets of the measurements corresponding to $\bbeta_{1}^{*}$ and $\bbeta_{2}^{*}$, respectively. Given $\{(\x_{i},y_{i})\}_{i=1}^{n}$, the goal is to recover $\bbeta_{1}^{*}$ and $\bbeta_{2}^{*}$. In particular, for the true regressor pair $ \ttheta^* = (\bbeta^*_1,\bbeta^*_2) $ and an estimator $ \hat{\ttheta} = (\hat{\bbeta}_1,\hat{\bbeta}_2) $ of it, we are interested in bounding the recovery error
\[
\rho(\hat{\ttheta},\ttheta^*):=\min\left\{ \left\Vert \hat{\bbeta}_{1}-\bbeta_{1}^*\right\Vert _{2}+\left\Vert \hat{\bbeta}_{2}-\bbeta_{2}^*\right\Vert _{2},\left\Vert \hat{\bbeta}_{1}-\bbeta_{2}^*\right\Vert _{2}+\left\Vert \hat{\bbeta}_{2}-\bbeta_{1}^*\right\Vert _{2}\right\},
\] 
i.e., the total error in both regressors up to permutation. Unlike the noiseless setting, in the presence of noise, the correct labels are in general irrecoverable.

The key high-level insight that leads to our optimization formulations, is to work in the lifted space of $p \times p$ matrices, {\em yet without lifting to $3$-tensors}. Using basic matrix concentration results not available for tensors, this ultimately allows us to provide optimal statistical rates. In this work, we seek to recover the following:

\begin{equation}
\begin{aligned}\K^{*} & :=\frac{1}{2}\left(\bbeta_{1}^{*}\bbeta_{2}^{*\top}+\bbeta_{2}^{*}\bbeta_{1}^{*\top}\right) \in \mathbb{R}^{p \times p},\\
\g^{*} & :=\frac{1}{2}\left(\bbeta_{1}^{*}+\bbeta_{2}^{*}\right) \in \mathbb{R}^p.
\end{aligned}
\label{eq:Kstar}
\end{equation}
Clearly $\bbeta_{1}^{*}$ and $\bbeta_{2}^{*}$ can be recovered from $\K^{*}$ and $\g^{*}$. Indeed, note that
\[
\J^{*}:=\g^{*}\g^{*\top}-\K^{*}=\frac{1}{4}\left(\bbeta_{1}^{*}-\bbeta_{2}^{*}\right)\left(\bbeta_{1}^{*}-\bbeta_{2}^{*}\right)^{\top}.
\]
Let $\lambda^{*}$ and $\vv^{*}$ be the first eigenvalue-eigenvector pair of $\J^{*}$. We have $\sqrt{\lambda^{*}}\vv^{*}:=\pm\frac{1}{2}\left(\bbeta_{1}^{*}-\bbeta_{2}^{*}\right)$; together with $\g^{*}$ we can recover $\bbeta_{1}^{*}$ and $\bbeta_{2}^{*}$. Given approximate versions $\hat{\K}$ and $\hat{\g}$ of $\K^{*}$ and $\g^{*}$, we obtain estimates $\hat{\bbeta}_1$ and $\hat{\bbeta}_2$ using a similar approach, which we give in Algorithm~\ref{alg:get_beta}.
\begin{algorithm}
\caption{\label{alg:get_beta}Estimate $\bbeta^{*}$'s}
Input: $(\hat{\K},\hat{\g})\in\mathbb{R}^{p\times p}\times\mathbb{R}^{p}$.
Compute the matrix $\hat{\J}=\hat{\g}\hat{\g}^{\top}-\hat{\K}$, and its first eigenvalue-eigenvector pair $\hat{\lambda}$ and $\hat{\vv}$.
Compute $\hat{\bbeta}_{1},\hat{\bbeta}_{2}=\hat{\g}\pm\sqrt{\hat{\lambda}}\hat{\vv}.$
Output: $(\hat{\bbeta}_{1},\hat{\bbeta}_{2})$
\end{algorithm}
We show below that in fact this recovery procedure is stable, so that if $\hat{\K}$ and $\hat{\g}$ are close to $\K^{*}$ and $\g^*$, Algorithm \ref{alg:get_beta} outputs $(\hat{\bbeta}_{1},\hat{\bbeta}_{2})$ that are close to $(\bbeta_{1}^{*},\bbeta_{2}^{*})$.

We now give the two formulations for arbitrary and stochastic noise, and we state the main results of the paper. For the arbitrary noise case, while one can use the same quadratic objective as we do in arbitrary case, it turns out that the analysis is more complicated than considering a similar objective -- an $\ell_1$ objective. In the noiseless setting, our results immediately imply exact recovery with an optimal number of samples, and in fact remove the additional log factors in the sample complexity requirements in~\cite{YiCaramanisSanghavi2013}. 
In both the arbitrary/adversarial noise setting and the stochastic noise setting, our results are information-theoretically optimal, as they match (up to at most a polylog factor) the minimax lower bounds we derive in Section~\ref{ssec:lowerbounds}.

\paragraph{Notation.} We use lower case bold letters to denote vectors, and capital bold-face letters for matrices. For a vector $ \ttheta $, $ \theta_i $ and $ \theta(i) $ both denote its $ i $-th coordinate. We use standard notation for matrix and vector norms, e.g., $\|\cdot\|_*$ to denote the nuclear norm (as known as the trace norm, which is the sum of the singular values of a matrix), $\|\cdot\|_F$ the Frobenius norm, and $\|\cdot\|$  the operator norm. We define a quantity we use repeatedly. Let
\begin{align}\label{eq:alpha}
\alpha:=&\frac{\left\Vert \bbeta_{1}^{*}-\bbeta_{2}^{*}\right\Vert _{2}^{2}}{\left\Vert \bbeta_{1}^{*}\right\Vert _{2}^{2}+\left\Vert \bbeta_{2}^{*}\right\Vert ^{2}}.
\end{align}
Note that $ \alpha>0 $ when $ \bbeta_1^* \neq\bbeta_2^* $, and is always bounded by $ 2 $. We say a number $c$ is a \emph{numerical constant} if $c$ is independent of the dimension $p$, the number of measurements $n$ and the quantity $ \alpha $. For ease of parsing, we typically use $c$ to denote a \emph{large} constant, and $\frac{1}{c}$ for a \emph{small} constant.

\subsection{Arbitrary Noise}
We consider first the setting of arbitrary noise, with the following specific setting. We take $\{\x_{i}\}$ to have i.i.d., zero-mean and sub-Gaussian entries\footnote{Recall that, as shown in~\cite{YiCaramanisSanghavi2013}, the general deterministic covariate mixed regression problem is NP-hard even in the noiseless setting.} with sub-Gaussian norm bounded by a numeric constant, $\mathbb{E}\left[(\x_{i}(l))^{2}\right]=1$, and $\mathbb{E}\left[(\x_{i}(l))^{4}\right]=\mu$ for all $ i\in[n] $ and $ l\in[p] $. We assume that $\mu$ is a fixed constant and independent of $p$ and $ \alpha $. If $ \{\x_{i}\} $ are standard Gaussian vectors, then these assumptions are satisfied with sub-Gaussian norm $ 1 $ and $ \mu=3 $. The only assumption on the noise $\e=(e_1,\;\cdots\;e_n)^{\top}$ is that it is bounded in $\ell_{2}$ norm. The noise $\e$ is otherwise arbitrary, possibly adversarial, and even possibly depending on $\left\{ \x_{i}\right\} $ and $\bbeta_{1}^{*},\bbeta_{2}^{*}$.

We consider the following convex program:
\begin{eqnarray}
\min_{\K,\g} && \left\Vert \K\right\Vert _{*}\label{eq:cvx_opt}\\
\mbox{s.t.} && \sum_{i=1}^{n}\left|-\left\langle \x_{i}\x_{i}^{\top},\K\right\rangle +2y_{i}\left\langle \x_{i},\g\right\rangle -y_{i}^{2}\right|\le\eta.\label{eq:constraint}
\end{eqnarray}
The intuition is that in the noiseless case with $ \e=\zero $, if we substitute the desired solution $ (\K^*,\g^*)$ given by~\eqref{eq:Kstar} into the above program, the LHS of~\eqref{eq:constraint} becomes zero; moreover, the rank of $ \K^* $ is $ 2 $, and minimizing the nuclear norm term in~\eqref{eq:cvx_opt} encourages the optimal solution to have low rank. Our theoretical results give a precise way to set the right hand side, $\eta$, of the constraint.
The next two theorems summarize our results for arbitrary noise. Theorem \ref{thm:determ} provides guarantees on how close the optimal solution $(\hat{\K},\hat{\g})$ is to $(\K^*,\g^*)$; then the companion result, Theorem \ref{thm:determ_beta}, provides quality bounds on $(\hat{\bbeta}_1,\hat{\bbeta}_2)$, produced by using Algorithm \ref{alg:get_beta} on the output $(\hat{\K},\hat{\g})$.

\begin{theorem}[Arbitrary Noise]
\label{thm:determ} There exist numerical positive constants $c_{1},\ldots, c_{6}$
such that the following holds. Assume $\frac{n_{1}}{n_{2}},\frac{n_{2}}{n_{1}}=\Theta(1).$
Suppose, moreover, that (1) $\mu>1$ and $\alpha>0$; (2) $\min\left\{ n_{1},n_{2}\right\} \ge c_{3}\frac{1}{\alpha}p$; (3) the parameter $\eta$ satisfies
\[
\eta\ge c_{4}\sqrt{n}\left\Vert \e\right\Vert _{2}\left\Vert \bbeta_{2}^{*}-\bbeta_{1}^{*}\right\Vert _{2};
\]
and (4) the noise satisfies 
\[
\left\Vert \e\right\Vert _{2}\le\frac{\sqrt{\alpha}}{c_{5}}\sqrt{n}\left(\left\Vert \bbeta_{1}^{*}\right\Vert_2 +\left\Vert \bbeta_{2}^{*}\right\Vert_2 \right).
\]
Then, with probability at least $1-c_{1}\exp(-c_{2}n)$, any optimal solution
$(\hat{\K},\hat{\g})$ to the program (\ref{eq:cvx_opt})--(\ref{eq:constraint})
satisfies
\begin{align*}
\left\Vert \hat{\K}-\K^{*}\right\Vert _{F} & \le c_6\frac{1}{\sqrt{\alpha}n}\eta,\\
\left\Vert \hat{\g}-\g^{*}\right\Vert _{2} & \le c_6\frac{1}{\sqrt{\alpha}n\left(\left\Vert \bbeta_{1}^{*}\right\Vert _{2}+\left\Vert \bbeta_{2}^{*}\right\Vert _{2}\right)}\eta.
\end{align*}

\end{theorem}

We then use Algorithm~\ref{alg:get_beta} to estimate $(\bbeta_1^*,\bbeta_2^*)$, which is stable as shown by the theorem below.

\begin{theorem}[Estimating $\bbeta^{*}$, arbitrary noise] \label{thm:determ_beta} Suppose conditions 1--4 in Theorem \ref{thm:determ} hold, and $\eta\asymp\sqrt{n}\left\Vert \e\right\Vert _{2}\left\Vert \bbeta_{2}^{*}-\bbeta_{1}^{*}\right\Vert _{2}$. Then with probability at least $1-c_{1}\exp(-c_{2}n)$, the output $ \hat{\ttheta} = (\hat{\bbeta}_1, \hat{\bbeta}_2) $ of Algorithm \ref{alg:get_beta} satisfies
\[
\rho(\hat{\ttheta}, \ttheta^*)\le\frac{1}{c_{3}\sqrt{\alpha}}\frac{\left\Vert \e\right\Vert _{2}}{\sqrt{n}},\quad b=1,2.
\]
\end{theorem}

Theorem \ref{thm:determ_beta} immediately implies exact recovery in the noiseless case. 
\begin{corollary}
[Exact Recovery]\label{cor:exact}Suppose $\e=\zero$, the conditions 1 and 2 in Theorem \ref{thm:determ} hold, and $\eta=0$. Then with probability at least $1-c_{1}\exp(-c_{2}n)$, Algorithm \ref{alg:get_beta} returns the true $\{\bbeta_1^{*},\bbeta_2^{*} \}$.
\end{corollary}

\noindent\textbf{Discussion of Assumptions:} 

(1) In Theorem \ref{thm:determ}, the condition $\mu>1$ is satisfied, for instance, if $\left\{ \x_{i}\right\}$ is Gaussian (with $\mu=3$). Moreover, this condition is in general necessary. To see this, suppose each $\x_{i}(l)$ is a Rademacher $\pm1$ variable, which has $\mu=1$, and $\bbeta_{1}^{*},\bbeta_{2}^{*}\in\mathbb{R}^{2}$. The response variable $ y_i $ must have the form 
\begin{align*}
y_{i} & =\pm(\bbeta_{b}^{*})_{1}\pm(\bbeta_{b}^{*})_{2}.
\end{align*}
Consider two possibilities: $\bbeta_{1}^{*}=-\bbeta_{2}^{*}=(1,0)^{\top}$ or $\bbeta_{1}^{*}=-\bbeta_{2}^{*}=(0,1)^{\top}$. In both cases, $(\x_{i},y_{i})$ may take any one of the values in $\left\{ \pm1\right\} ^{2}\times\left\{ \pm1\right\} $ with equal probabilities. Thus, it is impossible to distinguish between these two possibilities.

(2) The condition $\alpha>0$ holds if $\bbeta_{1}^{*}$ and $\bbeta_{2}^{*}$ are not equal. Suppose $\alpha$ is lower-bounded by a constant. The main assumption on the noise, namely, $\left\Vert \e\right\Vert _{2}\lesssim\sqrt{n}\left(\left\Vert \bbeta_{1}^{*}\right\Vert _{2}+\left\Vert \bbeta_{2}^{*}\right\Vert _{2}\right)$ (the condition 4 in Theorem~\ref{thm:determ}) cannot be substantially relaxed if we want a bound on $\left\Vert \hat{\g}-\g^{*}\right\Vert _{2}$. Indeed, if $\left|e_{i}\right|\gtrsim\left\Vert \bbeta_{b}^{*}\right\Vert _{2}$ for all $i$, then an adversary may choose $e_{i}$ such that
\[
y_{i}=\x_{i}^{\top}\bbeta_{b}^{*}+e_{i}=0, \quad \forall i,
\]
in which case the convex program (\ref{eq:cvx_opt})--(\ref{eq:constraint}) becomes independent of $\g$. That said, the case with condition 4 violated can be handled trivially. Suppose $\left\Vert \e\right\Vert _{2}\ge c_{4}\sqrt{\alpha n}\left(\left\Vert \bbeta_{1}^{*}\right\Vert _{2}+\left\Vert \bbeta_{2}^{*}\right\Vert _{2}\right)$ for any constant $ c_4 $. A standard argument for ordinal linear regression shows that the blind estimator $\hat{\bbeta}:=\min_{\bbeta}\sum_{i\in\mathcal{I}_1\cup\mathcal{I}_2}\left|\x_{i}^{\top}\bbeta-y_{i}\right|$ satisfies w.h.p.
\begin{align*}
\max\left\{\left\Vert \hat{\bbeta}-\bbeta_{1}^{*}\right\Vert_2, \left\Vert \hat{\bbeta}-\bbeta_{2}^{*}\right\Vert_2 \right\} \lesssim\frac{\left\Vert \e\right\Vert _{2}}{\sqrt{n}},
\end{align*}
and this bound is optimal (see the minimax lower bound in Section~\ref{ssec:lowerbounds}). Therefore, the condition 4 in Theorem \ref{thm:determ} is not really restrictive, i.e., the case when it holds is precisely the interesting setting.

(3) Finally, note that if $n_1/n_2 = o(1)$ or $n_2/n_1 = o(1)$, then a single $\bbeta^*$ explains 100\% (asymptotically) of the observed data. Moreover, the standard least squares solution recovers this $\bbeta^*$ at the same rates as in standard (not mixed) regression. 

\paragraph*{Optimality of sample complexity.}

The sample complexity requirements of Theorem~\ref{thm:determ_beta} and Corollary~\ref{cor:exact} are  optimal. The results require the number of samples $n_{1},n_{2}$ to be $\Omega(p)$. Since we are estimating two $p$ dimensional vectors without any further structure, this result cannot be improved.

\subsection{Stochastic Noise and Consistency}
We now consider the stochastic noise setting. We show that for Gaussian covariate in the balanced setting, we have asymptotic consistency and the rates we obtain match information-theoretic bounds we give in Section \ref{ssec:lowerbounds}, and hence are minimax optimal. Specifically, our setup is as follows. We assume the covariates $\{ \x_i\} $ have i.i.d. Gaussian entries with zero mean and unit variance . For the noise, we assume $\{e_{i}\}$ are i.i.d., zero-mean sub-Gaussian with  $\mathbb{E}\left[e_{i}^{2}\right]=\sigma^{2}$ and their sub-Gaussian norm $\left\Vert e_{i}\right\Vert _{\psi_{2}}\le c\sigma$ for some absolute constant $ c $, and are independent of $ \{\x_i\} $.

Much like in standard regression, the independence assumption on $ \{e_i\}$ makes the least-squares objective analytically convenient. In particular, we consider a Lagrangian formulation, regularizing the squared loss objective with the nuclear norm of $\K$. Thus, we solve the following:

\begin{equation}
\begin{aligned}\min_{\K,\g}\; & \sum_{i=1}^{n}\left(-\left\langle \x_{i}\x_{i}^{\top},\K\right\rangle +2y_{i}\left\langle \x_{i},\g\right\rangle -y_{i}^{2}+\sigma^{2}\right)^{2}+\lambda\left\Vert \K\right\Vert _{*}. \end{aligned}
\label{eq:L1_regularized_LS}
\end{equation}
We assume the noise variance $\sigma^{2}$ is known and can be estimated.\footnote{We note that similar assumptions are made in~\cite{chaganty13}. It might be possible to avoid the dependence on $ \sigma $ by using a symmetrized error term (see, e.g.,~\cite{cai2013rop}).} 
As with the arbitrary noise case, our first theorem guarantees $(\hat{\K},\hat{\g})$ is close to $(\K^*,\g^*)$, and then a companion theorem gives error bounds on estimating $\bbeta_b^*$.
\begin{theorem}
\label{thm:rand} For any constant $ 0<c_3<2 $, there exist numerical positive constant $ c_1,c_2,c_4,c_5,c_6 $, which might depend on $ c_3 $, such that the following hold. Assume $\frac{n_{1}}{n_{2}},\frac{n_{2}}{n_{1}}=\Theta(1).$
Suppose: (1) $ \alpha \ge c_3 $; (2) $\min\left\{ n_{1},n_{2}\right\} \ge c_4 p$; (3) $\left\{ \x_{i}\right\} $ are Gaussian; and (4) $\lambda$ satisfies \[\lambda\ge c_5 \sigma\left(\left\Vert \bbeta_{1}^{*}\right\Vert_2 +\left\Vert \bbeta_{2}^{*}\right\Vert_2 +\sigma\right)\left(\sqrt{np}+\left|n_{1}-n_{2}\right|\sqrt{p}\right)\log^{3}n. \]
With probability at least $1-c_{1}n^{-c_{2}}$, any optimal solution
$(\hat{\K},\hat{\g})$ to the regularized least squares program (\ref{eq:L1_regularized_LS})
satisfies
\begin{align*}
\left\Vert \hat{\K}-\K^{*}\right\Vert _{F} & \le c_6\frac{1}{n}\lambda,\\
\left\Vert \hat{\g}-\g^{*}\right\Vert _{2} & \le c_6\frac{1}{n\left(\left\Vert \bbeta_{1}^{*}\right\Vert +\left\Vert \bbeta_{2}^{*}\right\Vert +\sigma\right)}\lambda.
\end{align*}

\end{theorem}

The bounds in the above theorem depend on $ |n_1 - n_2| $. This appears as a result of the objective function in the formulation~\eqref{eq:L1_regularized_LS} and not an artifact of our analysis.\footnote{Intuitively, if the majority of the observations are generated by one of the $ \bbeta_b^* $, then the objective produces a solution that biases toward this $ \bbeta_b^* $ since this solution fits more observations. It might be possible to compensate for such bias by optimizing a different objective.}
Nevertheless, in the balanced setting with $|n_1- n_2|$ small, we have consistency with optimal convergence rate.
In this case, running Algorithm \ref{alg:get_beta} on the optimal solution $ (\hat{\K},\hat{\g}) $ of the program~\eqref{eq:L1_regularized_LS} to estimate the $\bbeta^{*}$'s, we have the following guarantees.
\begin{theorem}[Estimating $\bbeta^{*}$, stochastic noise] \label{thm:random_beta}
Suppose $\left|n_{1}-n_{2}\right|=O(\sqrt{n\log n})$,
the conditions 1--3 in Theorem \ref{thm:rand} hold,  $\lambda\asymp\sigma\left(\left\Vert \bbeta_{1}^{*}\right\Vert +\left\Vert \bbeta_{2}^{*}\right\Vert +\sigma\right)\sqrt{np}\log^{3}n$, and  $ n \ge c_3 p \log^8 n $. Then with probability
at least $1-c_{1}n^{-c_{2}}$, the output $ \hat{\ttheta} = (\hat{\bbeta}_1, \hat{\bbeta}_2) $ of Algorithm \ref{alg:get_beta}
satisfies 
\[
\rho(\hat{\ttheta}, \ttheta^*)
\le c_4 \sigma\sqrt{\frac{p}{n}}\log^4n+c_4\min\left\{ \frac{\sigma^{2}}{\|\bbeta^*_1\|_2+\|\bbeta^*_2\|_2 }\sqrt{\frac{p}{n}},  \sigma\left(\frac{p}{n}\right)^{1/4}\right\}\log^4n.
\]
\end{theorem}
Notice the error bound has three terms which are proportional to $\sigma\sqrt{\frac{p}{n}}$, $ \frac{\sigma^{2}}{\|\bbeta^*_b\|_2}\sqrt{\frac{p}{n}}$ and $\sigma\left(\frac{p}{n}\right)^{1/4}$, respectively (ignoring log factors). We shall see that these three terms match well with the information-theoretic lower bounds given in Section~\ref{ssec:lowerbounds}, and represent three phases of the error rate.
\paragraph*{Discussion of Assumptions.}
The theoretical results in this sub-section assume Gaussian covariate distribution in addition to sub-Gaussianity of the noise. This assumption can be relaxed, but using our analysis, it comes at a cost in terms of convergence rate (and hence sample complexity required for bounded error). It can be shown that $n = \tilde{O}(p\sqrt{p})$ suffices under a general sub-Gaussian assumption on the covariate. We believe this additional cost is an artifact of our analysis.

\subsection{Minimax Lower Bounds}
\label{ssec:lowerbounds}
In this subsection, we derive minimax lower bounds on the estimation errors for both the arbitrary and stochastic noise settings.  Recall that $\ttheta^{*}:=\left(\bbeta_{1}^{*},\bbeta_{2}^{*}\right)\in\mathbb{R}^{p}\times\mathbb{R}^{p}$
is the true regressor pairs, and we use $\hat{\ttheta}\equiv\hat{\ttheta}\left(\X,\y\right)=\left(\hat{\bbeta}_{1},\hat{\bbeta}_{2}\right)$
to denote any estimator, which is a measurable function of the observed
data $\left(\X,\y\right)$. 
For any $\ttheta=(\bbeta_{1},\bbeta_{2})$
and $\ttheta'=\left(\bbeta_{1}',\bbeta_{2}'\right)$ in $ \mathbb{R}^p \times \mathbb{R}^p $, we have defined the error (semi)-metric 
\[
\rho\left(\ttheta,\ttheta'\right):=\min\left\{ \left\Vert \bbeta{}_{1}-\bbeta_{1}'\right\Vert _{2}+\left\Vert \bbeta_{2}-\bbeta_{2}'\right\Vert _{2},\left\Vert \bbeta_{1}-\bbeta_{2}'\right\Vert _{2}+\left\Vert \bbeta_{2}-\bbeta_{1}'\right\Vert _{2}\right\}.
\]

\begin{remark}
\label{rem:metric} We show in the appendix that $\rho(\cdot,\cdot)$ satisfies the triangle inequality.
\end{remark}
We consider the following class of parameters:
\begin{equation}\label{eq:target}
\Theta(\underline{\gamma}):=\left\{ \ttheta =  (\bbeta_{1},\bbeta_{2})\in\mathbb{R}^{p}\times\mathbb{R}^{p}:2\left\Vert \bbeta_{1}-\bbeta_{2}\right\Vert \ge\left\Vert \bbeta_{1}\right\Vert +\left\Vert \bbeta_{2}\right\Vert \ge\ugamma\right\},
\end{equation}
i.e., pairs of regressors whose norms and separation are lower bounded.

We first consider the arbitrary noise setting, where the noise $ \e $ is assumed to lie in the $ \ell_2 $-ball $ \mathbb{B}(\epsilon) := \{\aalpha \in \mathbb{R}^n: \Vert \aalpha \Vert_2 \le \epsilon\}  $ and otherwise arbitrary. We have the following theorem.
\begin{theorem}[Lower bound, arbitrary noise]
\label{thm:lower_bound_determ}There exist universal constants $c_{0},c_{1}>0$
such that the following is true. If $n\ge c_{1}p$, then for any $\ugamma>0$ and any hidden labels $\z\in\left\{ 0,1\right\} ^{n}$,
we have 
\begin{equation}
\inf_{\hat{\ttheta}}\sup_{\ttheta^{*}\in\Theta(\ugamma)}\sup_{\e\in\mathbb{B}(\epsilon)}\rho(\hat{\ttheta},\ttheta^{*})\ge c_{0}\frac{\epsilon}{\sqrt{n}}\label{eq:determ_lower}
\end{equation}
with probability at least $1-n^{-10}$, where the probability is w.r.t.
the randomness in $\X$.
\end{theorem}
The lower bound above matches the upper bound given in Theorem~\ref{thm:determ_beta}, thus showing that our convex formulation is minimax optimal and cannot be improved. Therefore, Theorems~\ref{thm:determ_beta} and~\ref{thm:lower_bound_determ} together establish
the following minimax rate of the arbitrary noise setting
\[
\rho(\hat{\ttheta},\ttheta^{*})\asymp\frac{\Vert \e\Vert _{2}}{\sqrt{n}},
\]
which holds when $n\gtrsim p$.

For the stochastic noise setting, we further assume the two components have equal mixing weights. Recall that $ z_i \in \{0,1\}$ is the $ i $-th hidden label, i.e., $ z_i = 1 $ if and only if $ i\in \mathcal{I}_1 $ for $ i=1,\ldots,n $.  We have the following theorem.
\begin{theorem}[Lower bound, stochastic noise]
\label{thm:lower_bound_stochastic}
Suppose $n\ge p\ge 64$, $\X\in\mathbb{R}^{n\times p}$
has i.i.d. standard Gaussian entries, $\e$ has i.i.d. zero-mean Gaussian
entries with variance $\sigma^{2},$ and $z_{i}\sim\text{Bernoulli}(1/2)$.
The following holds for some absolute constants $0<c_{0},c_{1}<1$.
\begin{enumerate}
\item For any $\ugamma>\sigma$, we have
\begin{equation}
\inf_{\hat{\ttheta}}\sup_{\ttheta^{*}\in\Theta(\underline{\gamma})}\mathbb{E}_{\X,\z,\e}\left[\rho(\ttheta^{*},\hat{\ttheta})\right]
\ge c_{0}\sigma\sqrt{\frac{p}{n}}.\label{eq:bound_standard}
\end{equation}
\item For any $c_{1}\sigma\left(\frac{p}{n}\right)^{1/4}\le\ugamma\le\sigma$, we have
\begin{equation}
\inf_{\hat{\ttheta}}\sup_{\ttheta^{*}\in\Theta(\underline{\gamma})}\mathbb{E}_{\X,\z,\e}\left[\rho(\ttheta^{*},\hat{\ttheta})\right]
\ge c_{0}\frac{\sigma^{2}}{\ugamma}\sqrt{\frac{p}{n}}.\label{eq:bound_norm}
\end{equation}
\item For any $0<\ugamma\le c_{1}\sigma\left(\frac{p}{n}\right)^{1/4}$,
we have
\begin{equation}
\inf_{\hat{\ttheta}}\sup_{\ttheta^{*}\in\Theta(\underline{\gamma})}\mathbb{E}_{\X,\z,\e}\left[\rho(\ttheta^{*},\hat{\ttheta})\right]
\ge c_{0}\sigma\left(\frac{p}{n}\right)^{1/4}.\label{eq:bound_quarter}
\end{equation}

\end{enumerate}
Here $\mathbb{E}_{\X,\z,\e}\left[\cdot\right]$ denotes the expectation
w.r.t. the covariate $\X$, the hidden labels $\z$ and the
noise $\e$.
\end{theorem}
We see that the  three lower bounds in the above theorem match the three terms in the upper bound given in Theorem~\ref{thm:random_beta} respectively up to a polylog factor, proving the minimax optimality of the error bounds of our convex formulation. Therefore, Theorems~\ref{thm:random_beta} and~\ref{thm:lower_bound_stochastic} together establish
the following minimax error rate (up to a
polylog factor) in the stochastic noise setting:
\begin{align*}
\rho(\ttheta^{*},\hat{\ttheta}) & \asymp
\begin{cases}
\sigma\sqrt{\frac{p}{n}}, & \text{if } \ugamma\gtrsim\sigma,\\
\frac{\sigma^{2}}{\ugamma}\sqrt{\frac{p}{n}}, & \text{if } \sigma\left(\frac{p}{n}\right)^{\frac{1}{4}}\lesssim\ugamma\lesssim\sigma,\\
\sigma\left(\frac{p}{n}\right)^{\frac{1}{4}}, & \text{if } \ugamma\lesssim\sigma\left(\frac{p}{n}\right)^{\frac{1}{4}},
\end{cases}
\end{align*}
where $\ugamma$ is any lower bound on $\left\Vert \bbeta_{1}^{*}\right\Vert +\left\Vert \bbeta_{2}^{*}\right\Vert .$
Notice how the scaling of the minimax error rate exhibits three phases depending on the Signal-to-Noise Ratio (SNR) $ {\ugamma}/{\sigma} $. (1) In the high SNR regime with $ \ugamma \gtrsim \sigma $, we see a fast rate -- proportional to $ 1/\sqrt{n} $ -- that is dominated by the error of estimating a single $ \bbeta_b^* $ and is the same as the rate for standard linear regression. (2) In the low SNR regime with $ \ugamma \lesssim \sigma \left(\frac{p}{n}\right)^{\frac{1}{4}} $,  we have a slow rate that is proportional to $ 1/n^{\frac{1}{4}} $ and is associated with the demixing of the two components $ \bbeta_1^*,\bbeta_2^* $. (3) In the medium SNR regime, the error rate transitions between the fast and slow phases and depends in a precise way on the SNR. For a related phenomenon, see~\cite{azizyan2013sparseMixture,chen1995mixture}.

\subsection{Implications for Phase Retrieval}
As an illustration of the power of our results, we discuss an application to the \emph{Phase Retrieval} problem, which has recently received much attention (e.g., \cite{CandesStrohmerVoroninski2011,CandesEldarStrohmerVoroninski2011,CandesLi2012,chen2013covariance,NetrapalliJainSanghavi2013,cai2013rop}). Recall that in the real setting, the phase retrieval problem is essentially a regression problem without sign information. Most recent work has focused on the noiseless case. Here, the problem is as follows: we observe $(\x_{i},z_{i})\in\mathbb{R}^{p}\times\mathbb{R}$,
$i=1,2,\ldots n$, where
\[
z_{i}=\left|\x_{i}^{\top}\bbeta^{*}\right|.
\]
The goal is to recover the unknown vector $ \bbeta^* \in\mathbb{R}^{p}$.
The stability of recovery algorithms has also been considered. Most work has focused on the setting where noise is added to the phase-less measurements, that is,
\begin{equation}\label{eq:noisy_mag}
z_{i}=\left|\x_{i}^{\top}\bbeta^{*}\right| +e_{i}.
\end{equation}
In many applications, however, it is also natural to consider the setting where the measurement noise is added before the phase is lost. This corresponds to the model:
\begin{equation}\label{eq:noisy_phase}
z_{i}=\left|\x_{i}^{\top}\bbeta^{*}  +e_{i} \right|.
\end{equation}
We may call~\eqref{eq:noisy_phase} the \emph{Noisy Phase Model}, as opposed to the \emph{Noisy Magnitude Model}~\eqref{eq:noisy_mag} considered by previous work on phase retrieval. This problem can be reduced to a mixed regression problem and solved
by our algorithm. The reduction is as follows. We generate $n$ independent Rademacher random variables
$\epsilon_{i},i=1,\ldots,n$. For each 
$i$, we set $y_{i}=\epsilon_{i}z_{i}$.
Let $s_{i}:=\text{sign}\left(\x_{i}^{\top}\bbeta^{*}+e_{i}\right)$ and $e_{i}'=\epsilon_{i}s_{i}e_{i}$, where we use the convention that $ \text{sign}(0)=1 $. Then we have
\[
y_{i}=\epsilon_{i}\left|\x_{i}^{\top}\bbeta^{*}+e_{i}\right|=\epsilon_{i}s_{i}\left(\x_{i}^{\top}\bbeta^{*}+e_{i}\right)=\x_{i}^{\top}(\epsilon_{i}s_{i}\bbeta^{*})+e_{i}'.
\]
If we let $\bbeta_{1}^{*}=\bbeta^{*}$, $\bbeta_{2}^{*}=-\bbeta^{*}$,
$\mathcal{I}_{1}=\{i:\epsilon_{i}s_{i}=1\}$ and $\mathcal{I}_{2}=\left\{ i:\epsilon_{i}s_{i}=-1\right\} $,
then the model becomes
\[
y_{i}=\x_{i}^{\top}\bbeta_{b}^{*}+e_{i}',\quad\forall i\in\mathcal{I}_{b},
\]
which is precisely the mixed regression model we consider. 

Note that with probability at least $ 1-n^{-3} $, $\frac{n}{2}-\sqrt{10n\log n}\le n_{b}=\left|\mathcal{I}_{b}\right|\le\frac{n}{2}+\sqrt{10n\log n}$ for $b=1,2$, so $\left|n_{1}-n_{2}\right|=O\left(\sqrt{n\log n}\right)$. Also note that $\left\Vert \e'\right\Vert _{2}=\left\Vert \e\right\Vert _{2}$. Conditioned on $\{\mathcal{I}_b\}$, the distribution of $ \{\x_i\} $ is the same as its unconditional distribution. Therefore, applying our arbitrary-noise result from Theorem \ref{thm:determ_beta}, we immediately get the following guarantees for phase retrieval under the Noisy Phase Model.
\begin{corollary}
[Phase retrieval, arbitrary noise] \label{cor:determ_phase} Consider the Noisy Phase Model in~\eqref{eq:noisy_phase}. Suppose the $\{\x_{i}\}$ are i.i.d., zero-mean sub-Gaussian with bounded sub-Gaussian norm, unit variance and fourth moment $\mu>1$, $n\gtrsim p$, $\eta\asymp c_{4}\sqrt{n}\left\Vert \e\right\Vert _{2}\left\Vert \bbeta^{*}\right\Vert _{2}$ and the noise is arbitrary, but bounded in magnitude: $\left\Vert \e\right\Vert _{2}\lesssim\sqrt{n}\left\Vert \bbeta^{*}\right\Vert_2$. Then using the reduction described above, the output of the program (\ref{eq:cvx_opt})--(\ref{eq:constraint}) followed by Algorithm \ref{alg:get_beta} satisfies
\[
\min_{b=1,2}\left\Vert \hat{\bbeta}_b-\bbeta^{*}\right\Vert _{2}\lesssim\frac{\left\Vert \e\right\Vert _{2}}{\sqrt{n}}
\]
 with probability at least $ 1-n^{-2} $.
\end{corollary}

The error bound above is again order-wise optimal, as we cannot achieve a smaller error even if the phase is not lost. 
Similarly as before, the large noise case with $\left\Vert \e\right\Vert _{2}\ge c_{4}\sqrt{n}\left\Vert \bbeta^{*}\right\Vert _{2}$ can be handled trivially using the blind estimator $\hat{\bbeta}:=\min_{\bbeta}\sum_{i\in[n]}\left|\x_{i}^{\top}\bbeta-z_{i}\right|$, which in this case satisfies the optimal error bound $ \left\Vert \hat{\bbeta}-\bbeta^{*}\right\Vert _{2}\lesssim\left\Vert \e\right\Vert _{2}/\sqrt{n} $.

Next, consider the stochastic noise  case where  $e_{i}$ is i.i.d., zero-mean {symmetric} sub-Gaussian with variance $\sigma^{2}$. Conditioned on $\{\mathcal{I}_{b}\}$, the conditional distributions of $\{e_{i}'\}$ and $\{\x_{i}\}$ inherit the properties of $e_i$ and the unconditional $ \x_i $, and are independent of each other. Applying Theorem~\ref{thm:random_beta}, we have the following.
\begin{corollary}
[Phase retrieval, stochastic noise] \label{cor:random_phase} Consider the Noisy Phase Model in~\eqref{eq:noisy_phase}. Suppose the $\{\x_{i}\}$ are i.i.d., zero-mean Gaussian with unit variance, and suppose that the noise $e_{i}$ is i.i.d., zero-mean symmetric sub-Gaussian with sub-Gaussian norm bounded by $ c_3\sigma $ and variance equal to $\sigma^{2}$. Suppose further that $n\gtrsim p$ and $\lambda\asymp\sigma\left(\left\Vert \bbeta^{*}\right\Vert_2 +\sigma\right)\sqrt{np}\log^{4}n$. Then using the reduction described above, the output of the program (\ref{eq:L1_regularized_LS}) followed by Algorithm \ref{alg:get_beta} satisfies (up to the sign of $\bbeta^{*}$)
\[
\min_{b=1,2}\left\Vert \hat{\bbeta}_b-\bbeta^{*}\right\Vert _{2}\lesssim\sigma\sqrt{\frac{p}{n}}\log^4n+\min\left\{ \frac{\sigma^{2}\sqrt{\frac{p}{n}}}{\left\Vert \bbeta^{*}\right\Vert_2 },\sigma\left(\frac{p}{n}\right)^{\frac{1}{4}}\right\}\log^4n
\]
with probability at least $1-n^{-2}$.
\end{corollary}
Phase retrieval is most interesting in the complex setting. Extension to this case is an interesting future direction.

\subsection{Scalability}

Finally, we make a comment on the scalability of the approach illustrated here. Both formulations~(\ref{eq:cvx_opt})--(\ref{eq:constraint}) and~\eqref{eq:L1_regularized_LS} are Semidefinite Programs (SDP).  In the arbitrary noise setting, the constraint in the convex program (\ref{eq:cvx_opt})--(\ref{eq:constraint}) can be rewritten as a collection of linear constraints through the standard transformation of convex $\ell_1$ constraints. The Lagrangian formulation (\ref{eq:L1_regularized_LS}) in the setting of stochastic noise, involves minimizing the sum of a trace norm term and a smooth quadratic term. The complexity of solving this regularized quadratic in the matrix space has similar complexity to problems such as matrix completion and PhaseLift, and first order methods can easily be adapted, thus allowing solution of large scale instances of the mixed regression problem.

\section{Proof Outline}
\label{sec:outline}
In this section, we provide the outline and the key ideas in the proofs of Theorems~\ref{thm:determ}, ~\ref{thm:rand}, ~\ref{thm:lower_bound_determ} and~\ref{thm:lower_bound_stochastic}. The complete proofs, along with the perturbation results of Theorems~\ref{thm:determ_beta}, \ref{thm:random_beta},
are deferred to the appendix.

The main hurdle is proving strict curvature near the desired solution $ (\K^*,\g^*) $ in the allowable directions. This is done by demonstrating that a linear operator related to the $ \ell_1/\ell_2 $ errors satisfies a restricted-isometry-like condition, and that this in turn implies a strict convexity condition along the cone centered at $(\K^*,\g^*)$ of all directions defined by potential optima. 

\subsection{Notation and Preliminaries\label{sub:notation}}

We use $\bbeta_{-b}^{*}$ to denote $\bbeta_{2}^{*}$ if $b=1$ and $\bbeta_{1}^{*}$ if $b=2$. Let $\ddelta_{b}^{*}:=\bbeta_{b}^{*}-\bbeta_{-b}^{*}.$ Without loss of generality, we assume $\mathcal{I}_{1}=\left\{ 1,\ldots,n_{1}\right\} $ and $\mathcal{I}_{2}=\left\{ n_{1}+1,\ldots,n\right\} $. For $i=1,\ldots,n_{1}$, we define $\x_{1,i}:=\x_{i}$, $y_{1,i}=y_{i}$ and $e_{1,i} = e_i$; correspondingly, for $i=1,\ldots,n_{2}$, we define $\x_{2,i}:=\x_{n_{1}+i}$, $y_{2,i}:=y_{n_1+i}$ and $e_{2,n+i}$. For each $b=1,2$, let $\X_{b}\in\mathbb{R}^{n_{b}\times p}$ be the matrix with rows $\{\x_{b,i}^{\top},i=1,\ldots,n_{b}\}$. For $b=1,2$ and $j=1,\ldots,\left\lfloor n_{b}/2\right\rfloor$, define the matrix $\B_{b,j}:=\x_{b,2j}\x_{b,2j}^{\top}-\x_{b,2j-1}\x_{b,2j-1}^{\top}$. Also let $\e_{b}:=[e_{b,1}\;\cdots\;e_{b,n_{b}}]^{\top}\in\mathbb{R}^{n_{b}}.$

For $b\in\left\{ 1,2\right\} $, define the mapping $\BB_{b}:\mathbb{R}^{p\times p}\mapsto\mathbb{R}^{\left\lfloor n_{b}/2\right\rfloor }$
by
\[
\left(\BB_{b}\Z\right)_{j}=\frac{1}{\left\lfloor n_{b}/2\right\rfloor }\left\langle \B_{b,j},\Z\right\rangle ,\quad\textrm{for each }j=1,\ldots,\left\lfloor n_{b}\right\rfloor .
\]
Since $y_{b,i}=\x_{b,i}^{\top}\bbeta_{b}^{*}+e_{b,i}$,
$i\in[n_{b}]$, we have for any $\Z\in\mathbb{R}^{p\times p}$,
$\z\in\mathbb{R}^{p}$ and for all $ j=1,\ldots,\left\lfloor n_{b}\right\rfloor $,
\begin{align*}
\frac{1}{\left\lfloor n_{b}/2\right\rfloor }\left(\left\langle \B_{b,j},\Z\right\rangle -2\dd_{b,j}^{\top}\z\right)
=&\frac{1}{\left\lfloor n_{b}/2\right\rfloor }\left\langle \B_{b,j},\Z-2\bbeta_{b}^{*}\z^{\top}\right\rangle +\left(e_{b,2j}\x_{b,2j}-e_{b,2j-1}\x_{b,2j}\right)^{\top}\z \\
=& \left(\BB_{b}\left(\Z-2\bbeta_{b}^{*}\z^{\top}\right)\right)_{j}+\left(e_{b,2j}\x_{b,2j}-e_{b,2j-1}\x_{b,2j}\right)^{\top}\z,.
\end{align*}
For each $b=1,2$, we also define the matrices $\A_{b,i}:=\x_{b,i}\x_{b,i}^{\top}$,
$i\in[n_{b}]$ and the mapping $\AA_{b}:\mathbb{R}^{p\times p}\mapsto\mathbb{R}^{n_{b}}$ given by
\[
\left(\AA_{b}\Z\right)_{i}=\frac{1}{n_{b}}\left\langle \A_{b,i},\Z\right\rangle ,\quad\text{for each }i\in[n_{b}].
\]

The following notation and definitions are standard. Let the rank-$2$ SVD of $\K^{*}$
be $\U\bm{\Sigma}\V^{\top}$. Note that $\U$ and $\V$ have the same
column space, which equals $\text{span}(\bbeta_{1}^{*},\bbeta_{2}^{*})$.
Define the projection matrix $\bm{P}_{\U}:=\U\U^{\top}=\V\V^{\top}$ 
and the subspace $T :=\left\{ \P_{\U}\Z+\Y\P_{\U}:\Z,\Y\in\mathbb{R}^{p\times p}\right\}$.
Let $T^{\bot}$ be the orthogonal subspace of $T$. The projections
to $T$ and $T^{\bot}$ are given by
$$
\mathcal{P}_{T}\Z  :=\P_{\U}\Z+\Z\P_{\U}-\P_{\U}\Z\P_{\U}, \qquad
\mathcal{P}_{T^{\bot}}\Z :=\Z-\mathcal{P}_{T}\Z.
$$
Denote the optimal solution to the optimization problem of interest (either~\eqref{eq:cvx_opt} or~\eqref{eq:L1_regularized_LS}) as $(\hat{\K},\hat{\g})=(\K^{*}+\hat{\H},\g^{*}+\hat{\h})$.  Let $\hat{\H}_{T}:=\mathcal{P}_{T}\hat{\H}$ and $\hat{\H}_{T}^{\bot}:=\mathcal{P}_{T^{\bot}}\hat{\H}$.

\subsection{Upper Bounds for Arbitrary Noise: Proof Outline}
\label{ssec:determoutline}
The proof follows from three main steps. 
\begin{enumerate}
\item[(1)] First, the $ \ell_1 $ error term that in this formulation appears in the LHS of the constraint (\ref{eq:constraint}) in the optimization, is naturally related to the operators $\AA_b$. Using the definitions above, for any feasible $(\K,\g) = (\K^*+\H,\g^*+\h)$, the constraint (\ref{eq:constraint}) in the optimization program can be rewritten as
$$
\textstyle{\sum_{b}}\left\Vert n_{b}\AA_{b}\left(-\H+2\bbeta_{b}^{*}\h^{\top}\right)+2\e_{b}\circ\left(\X_{b}\h\right)-\e_{b}\circ\left(\X_{b}\ddelta_{b}^{*}\right)-\e_{b}^{2}\right\Vert _{1}\le\eta.
$$
This inequality holds in particular for $ \H=\zero $ and $ \h=\zero$ under the conditions of the theorem, as well as for $ \hat{\H} $ and $ \hat{\h} $ associated with the optimal solution since it is feasible.
Now, using directly the definitions for $\AA_b$ and $\BB_b$, and a simple triangle inequality, we obtain that
\begin{align*}
\left\lfloor n_{b}/2\right\rfloor \left\Vert \BB_{b}\left(-\hat{\H}+2\bbeta_{b}^{*}\hat{\h}^{\top}\right)\right\Vert _{1}  & \le n_{b}\left\Vert \AA_{b}\left(-\hat{\H}+2\bbeta_{b}^{*}\hat{\h}^{\top}\right)\right\Vert _{1}.
\end{align*}
From the last two display equations, and using now the assumptions on $\eta$ and on $\e$, we obtain an upper bound for $\BB$ using the error bound $\eta$:
$$
\sum_{b}n\left\Vert \BB_{b}\left(-\hat{\H}+2\bbeta_{b}^{*}\hat{\h}^{\top}\right)\right\Vert _{1}-c_{2}\sum_{b}\sqrt{n}\left\Vert \e_{b}\right\Vert _{2}\|\hat{\h}\| _{2} \leq 2 \eta .
$$
\item[(2)] Next, we obtain a lower-bound on the last LHS by showing the operator $\BB$ is an approximate isometry on low-rank matrices. Note that we want to bound the $\|\cdot\|_2$ norm of $\hat{\h}$ and the Frobenius norm of $\hat{\H}$, though we currently have an $\ell_1$-norm bound on $\BB$ in terms of $\eta$, above. Thus, the RIP-like condition we require needs to relate these two norms. We show that with high probability, for low-rank matrices,
$$
\underbar{\ensuremath{\delta}}\left\Vert \Z\right\Vert _{F}\le\left\Vert \BB_{b}\Z\right\Vert _{1}\le\bar{\delta}\left\Vert \Z\right\Vert _{F},\quad\forall\Z\in\mathbb{R}^{p\times p}\textrm{ with rank}(\Z)\le\rho.
$$
Proving this RIP-like result is done using concentration and an $ \epsilon $-net argument, and requires the assumption $ \mu>1 $. We then use this and the optimality of $ (\hat{\K},\hat{\g}) $ to obtain the desired lower-bounds 
\begin{align*}
\sum_{b}\left\Vert \BB_{b}\left(\hat{\H}-2\bbeta_{b}^{*}\hat{\h}^{\top}\right)\right\Vert _{1} &\ge \frac{\sqrt{\alpha}}{c''}\left\Vert \hat{\H}_{T}\right\Vert _{F}\overset{(d)}{\ge}\frac{\sqrt{\alpha}}{c'}\left\Vert \hat{\H}\right\Vert _{F},\\
\sum_{b}\left\Vert \BB_{b}\left(\hat{\H}-2\bbeta_{b}^{*}\hat{\h}^{\top}\right)\right\Vert _{1} &\ge \frac{\sqrt{\alpha}}{c'}\left(\left\Vert \bbeta_{1}^{*}\right\Vert _{2}+\left\Vert \bbeta_{2}^{*}\right\Vert _{2}\right)\|\hat{\h}\| _{2}.
\end{align*}

\item[(3)] The remainder of the proof involves combining the upper and lower bounds obtain in the last two steps. After some algebraic manipulations, and use of conditions in the assumptions of the theorem, we obtain the desired recovery error bounds
\[
\|\hat{\h}\| _{2}\lesssim\frac{1}{\sqrt{\alpha}n\left(\left\Vert \bbeta_{1}^{*}\right\Vert _{2}+\left\Vert \bbeta_{2}^{*}\right\Vert _{2}\right)}\eta, \qquad
\left\Vert \hat{\H}\right\Vert _{F}\lesssim\frac{1}{n\sqrt{\alpha}}\eta.
\]

\end{enumerate}

\subsection{Upper Bounds for Stochastic Noise: Proof Outline}
\label{ssec:stochasticoutline}
The main conceptual flow of the proof for the stochastic setting is quite similar to the deterministic noise case, though some significant additional steps are required, in particular, the proof of a second RIP-like result. 

\begin{enumerate}
\item[(1)] For the deterministic case, the starting point is the constraint, which allows us to bound $\AA_b$ and $ \BB_b $ in terms of $\eta$ using feasibility of $ (\K^*,\g^*) $ and $(\K^*+\hat{\H},\g^*+\hat{\h})$. In the stochastic setup we have a Lagrangian (regularized) formulation, and hence we obtain the analogous result from optimality. Thus, the first step here involves showing that as a consequence of optimality, the solution $(\hat{\K},\hat{\g})=(\K^{*}+\hat{\H},\g^{*}+\hat{\h})$ satisfies:
\begin{equation*}
\sum_{b}\left\Vert n_{b}\AA_{b}\left(\!-\hat{\H}\!+\!2\bbeta_{b}^{*}\hat{\h}^{\top}\right)+2\e_{b}\!\circ\!(\X_{b}\hat{\h})\right\Vert _{2}^{2}
\le\lambda\!\left(\frac{3}{2}\left\Vert \hat{\H}_{T}\right\Vert _{*}\!-\!\frac{1}{2}\left\Vert \hat{\H}_{T}^{\bot}\right\Vert _{*}\right)+\lambda\left(\gamma\!+\!\sigma\right)\|\hat{\h}\| _{2},
\end{equation*}
where we have defined the parameter  $\gamma :=\left\Vert \bbeta_{1}^{*}\right\Vert _{2}+\left\Vert \bbeta_{2}^{*}\right\Vert _{2}$.
The proof of this inequality involves carefully bounding several noise-related terms using concentration. A consequence of this inequality is that $ \hat{\H} $ and $ \hat{\h} $ cannot be arbitrary, and must live in a certain cone. 
\item[(2)] The RIP-like condition for $\BB_b$ in the stochastic case is more demanding. We  prove a second RIP-like condition for $\|\BB_b \mathbf{Z} - \D_b \mathbf{z}\|_1$, using the Frobenius norm of $\Z$ and the $\ell_2$-norm of $\mathbf{Z}$:
\begin{align*}
\underbar{\ensuremath{\delta}}\left(\left\Vert \Z\right\Vert _{F}+\sigma\left\Vert \z\right\Vert _{2}\right)\le\left\Vert \BB_{b}\Z-\D_{b}\z\right\Vert _{1} &\le\bar{\delta}\left(\left\Vert \Z\right\Vert _{F}+\sigma\left\Vert \z\right\Vert _{2}\right),\\
&\forall\z\in\mathbb{R}^{p},\forall\Z\in\mathbb{R}^{p\times p}\textrm{ with rank}(\Z)\le r.
\end{align*}
We then bound $\AA$ by terms involving $\BB$, and then invoke the above RIP condition and the cone constraint to obtain the following lower bound:
$$ 
\sum_{b}\left\Vert n_{b}\AA_{b}\left(-\hat{\H}+2\bbeta_{b}^{*}\hat{\h}^{\top}\right)+2\e_{b}\circ\left(\X_{b}\hat{\h}\right)\right\Vert _{2}^{2} \gtrsim \frac{1}{8}n\left(\left\Vert \hat{\H}_{T}\right\Vert _{F}+\left(\gamma+\sigma\right)\|\hat{\h}\| _{2}\right)^{2}.
$$
\item[(3)] We now put together the upper and lower bounds in Step (1) and Step (2). This gives
\begin{align*}
n\left(\left\Vert \hat{\H}_{T}\right\Vert _{F}+\left(\gamma+\sigma\right)\|\hat{\h}\| _{2}\right)^{2} & \lesssim\lambda\left\Vert \H_{T}\right\Vert _{F}+\lambda(\gamma+\sigma)\|\hat{\h}\| _{2},
\end{align*}
from which it eventually follows that 
$$
\|\hat{\h}\| _{2}\lesssim\frac{1}{n\left(\gamma+\sigma\right)}\lambda, \qquad \left\Vert \hat{\H}\right\Vert _{F} \lesssim\frac{1}{n}\lambda.
$$  
\end{enumerate}

\subsection{Lower Bounds: Proof Outline}

The high-level ideas in the proofs of Theorems~\ref{thm:lower_bound_determ} and~\ref{thm:lower_bound_stochastic} are similar: we use a standard argument~\cite{yu1997assouad,yang1999minimax,birge1983approximation} to convert the estimation problem into a hypothesis testing problem, and then use information-theoretic inequalities to lower bound the error probability in hypothesis testing. In particular, recall the definition of the set~$ \Theta(\ugamma) $ of regressor pairs in~\eqref{eq:target}; we construct a $ \delta $-packing $ \Theta = \{\ttheta_1, \ldots,\ttheta_M\} $ of $ \Theta(\ugamma) $ in the metric $ \rho $, and use the following inequality: 
\begin{align}\label{eq:testing}
\inf_{\hat{\ttheta}} \sup_{\ttheta^*  \in \Theta(\ugamma)} \mathbb{E} \left[ \rho(\hat{\ttheta}, \ttheta^*)\right]
\ge \delta \inf_{\tilde{\ttheta}} \mathbb{P}\left( \tilde{\ttheta} \neq \ttheta^*\right),
\end{align}
where on the RHS $ \ttheta^* $ is assumed to be sampled uniformly at random from $ \Theta $. To lower-bound the minimax expected error by $ \frac{1}{2} \delta $, it suffices to show that the probability on the last RHS is at least $ \frac{1}{2}$. By Fano's inequality~\cite{cover2012information}, we have
\begin{align}\label{eq:fano}
\mathbb{P}\left( \tilde{\ttheta} \neq \ttheta^*\right) \ge 1- \frac{I\left(\y,\X; \ttheta^* \right)+\log 2}{\log M}.
\end{align}
It remains to construct a packing set $ \Theta $ with the appropriate separation~$ \delta $ and cardinality~$ M $, and to upper-bound the mutual information $ I\left(\y,\X; \ttheta^* \right) $. We show how to do this for Part 2 of Theorem~\ref{thm:lower_bound_stochastic}, for which the desired separation is $ \delta =  2c_0 \frac{\sigma^2}{\kappa} \sqrt{\frac
{p}{n}}$, where $ \kappa = \frac{\ugamma}{2} $. Let $ \{\xxi_1,\ldots,\xxi_M\} $ be a $ \frac{p-1}{16} $-packing of $ \{0,1\}^{p-1} $ in Hamming distance with  $ \log M \ge (p-1)/16  $, which exists by the Varshamov-Gilbert bound~\cite{tsybakov2009nonparm}. We construct $ \Theta $ by setting 
$\ttheta_{i}:=\left(\bbeta_{i},-\bbeta_{i}\right)$ for $ i=1,\ldots,M $ with 
\[
\bbeta_{i}=\kappa_{0}\eeps_{p}+\sum_{j=1}^{p-1}\left(2\xxi_{i}(j)-1\right)\tau\eeps_{j},
\]
where $ \tau = \frac{4\delta}{\sqrt{p-1}} $, $\kappa_{0}^{2}=\kappa^{2}-(p-1)\tau^{2}$, and $ \eeps_j $ is the $ j $-th standard basis in $ \mathbb{R}^{p} $. We verify that this $ \Theta $  indeed defines a $ \delta $-packing of $ \Theta(\ugamma) $, and moreover satisfies $\left\Vert \bbeta_{i}-\bbeta_{i'}\right\Vert ^{2}\le 16 \delta^2$ for all $ i\neq i' $. To bound the mutual information, we observe that by independence between $ \X $ and $ \ttheta^* $, we have
\begin{align*}
 I\left(\ttheta^{*};\X,\y\right)
\le  \frac{1}{M^{2}}\sum_{1\le i,i'\le M}D\left(\mathbb{P}_{i}\Vert\mathbb{P}_{i'}\right)
=  \frac{1}{M}\sum_{1\le i,i'\le M} \sum_{j=1}^n \mathbb{E}_{\X}\left[D\left(\mathbb{P}^{(j)}_{i,\X}\Vert\mathbb{P}^{(j)}_{i',\X}\right)\right],
\end{align*}
where $ \mathbb{P}^{(j)}_{i,\X} $ denotes the distribution of $ y_j $ conditioned on $ \X $ and $ \ttheta^* = \ttheta_i $.
The remaining and crucial step is to obtain sharp upper bounds on the above  KL-divergence between two mixtures of one-dimensional Gaussian distributions. This requires some technical calculations, from which we obtain
\[
\mathbb{E}_{\X}D\left(\mathbb{P}^{(j)}_{i,\X}\Vert\mathbb{P}^{(j)}_{i',\X}\right)\le \frac{c'\left\Vert \bbeta_{i}-\bbeta_{i'}\right\Vert ^{2}\kappa^{2}}{\sigma^{4}}.
\]
We conclude that $ I(\ttheta^*; \X, \y ) \le \frac{1}{4} \log M$. Combining with~\eqref{eq:testing} and~\eqref{eq:fano} proves Part 2 of Theorem~\ref{thm:lower_bound_stochastic}. Theorem~\ref{thm:lower_bound_determ} and Parts 1,  3 of Theorem~\ref{thm:lower_bound_stochastic} are proved in a similar manner.

\section{Conclusion}
This paper provides a computationally and statistically efficient algorithm for mixed regression with two components. To the best of our knowledge, the is the first efficient algorithm that can provide $O(p)$ sample complexity guarantees. Under certain conditions, we prove matching lower bounds, thus demonstrating our algorithm achieves the minimax optimal rates. There are several interesting open questions that remain. Most immediate is the issue of understanding the degree to which the assumptions currently required for minimax optimality can be removed or relaxed. The extension to more than two components is important, though how to do this within the current framework is not obvious. 

At its core, the approach here is a method of moments, as the convex optimization formulation produces an estimate of the cross moments, $(\bbeta_1^{\ast}\bbeta_2^{\ast \top} + \bbeta_2^{\ast}\bbeta_1^{\ast \top})$. An interesting aspect of these results is the significant improvement in sample complexity guarantees this tailored approach brings, compared to a more generic implementation of the tensor machinery which requires use of third order moments. Given the statistical and also computational challenges related to third order tensors, understanding the connections more carefully seems to be an important future direction. 

\section*{Acknowledgment}

We thank Yuxin Chen for illuminating conversations on the topic. We acknowledge support from NSF Grants EECS-1056028, CNS-1302435, CCF-1116955, and the USDOT UTC--D-STOP Center at UT-Austin.

\section*{Appendix}

\appendix


\section{Proofs of Theorems \ref{thm:determ_beta} and \ref{thm:random_beta}}
\label{sec:proofs_beta}
In this section, we show that an error bound on the input $(\hat{\K},\hat{\g})$ of Algorithm \ref{alg:get_beta} implies an error bound on its output $(\hat{\bbeta}_{1},\hat{\bbeta_{2}})$. Recall the quantities $\hat{\J}$, $\J^{*}$, $\hat{\lambda}$, $\lambda^{*}$,$\hat{\vv}$ and $\vv^{*}$ defined in Section~\ref{sec:setup} and in Algorithm \ref{alg:get_beta}. 

A key component of the proof involves some perturbation bounds. We prove these in the first section below, and then use them to prove Theorems \ref{thm:determ_beta} and \ref{thm:random_beta} in the two subsequent sections.

\subsection{Perturbation Bounds}
We require the following perturbation bounds.
\begin{lemma}
\label{lem:perturb}If 
$
\left\Vert \hat{\J}-\J^{*}\right\Vert _{F}\le\delta,
$
then 
\[
\left\Vert \sqrt{\hat{\lambda}}\hat{\vv}-\sqrt{\lambda^{*}}\vv^{*}\right\Vert_2 \le10\min\left\{ \frac{\delta}{\sqrt{\left\Vert \J^{*}\right\Vert }},\sqrt{\delta}\right\} .
\]
\end{lemma}
\proof
 By Weyl's inequality, we have
\[
\left|\hat{\lambda}-\lambda^{*}\right|\le\left\Vert \hat{\J}-\J^{*}\right\Vert \le\delta.
\]
This implies
\begin{equation}
\left|\sqrt{\hat{\lambda}}-\sqrt{\lambda^{*}}\right|=\left|\frac{\hat{\lambda}-\lambda^{*}}{\sqrt{\hat{\lambda}}+\sqrt{\lambda^{*}}}\right|\le2\min\left\{ \frac{\delta}{\sqrt{\lambda^{*}}},\sqrt{\delta}\right\}.
\label{eq:eval_bound}
\end{equation}
Using Weyl's inequality and Davis-Kahan's sine theorem, we obtain
\begin{equation}
\left|\sin\angle(\hat{\vv},\vv^{*})\right|
\le
\min\left\{ \frac{2\Vert \hat{\K}-\K^{*}\Vert }{\left\Vert \K^{*}\right\Vert },1\right\} \le\min\left\{ \frac{2\delta}{\lambda^{*}},1\right\}.
\label{eq:evec_bound}
\end{equation}
On the other hand, we have
\begin{align*}
\left\Vert \hat{\vv}\sqrt{\hat{\lambda}}-\vv^{*}\sqrt{\lambda^{*}}\right\Vert_2 
 & \le\left\Vert \hat{\vv}\sqrt{\hat{\lambda}}-\vv^{*}\sqrt{\hat{\lambda}}\right\Vert_2 +\left\Vert \vv^{*}\sqrt{\hat{\lambda}}-\vv^{*}\sqrt{\lambda^{*}}\right\Vert_2 \\
 & =\sqrt{\hat{\lambda}}\left\Vert \hat{\vv}-\vv^{*}\right\Vert_2 +\left\Vert \vv^{*}\right\Vert_2 \left|\sqrt{\hat{\lambda}}-\sqrt{\lambda^{*}}\right|\\
 & =\left(\sqrt{\lambda^{*}}+\sqrt{\hat{\lambda}}-\sqrt{\lambda^{*}}\right)\left\Vert \hat{\vv}-\vv^{*}\right\Vert_2 +\left\Vert \vv^{*}\right\Vert_2 \left|\sqrt{\hat{\lambda}}-\sqrt{\lambda^{*}}\right|\\
 & \le\sqrt{\lambda^{*}}\left\Vert \hat{\vv}-\vv^{*}\right\Vert_2 +3\left|\sqrt{\hat{\lambda}}-\sqrt{\lambda^{*}}\right|,
\end{align*}
where in the last inequality we use the fact that $\left\Vert \vv^{*}\right\Vert =\left\Vert \hat{\vv}\right\Vert =1$.
Elementary calculation shows that
\[
\left\Vert \hat{\vv}-\vv^{*}\right\Vert_2 
 =2\left|\sin\frac{1}{2}\angle(\hat{\vv},\vv^{*})\right|\le\sqrt{2}\left|\sin\angle(\hat{\vv},\vv^{*})\right|.
\]
It follows that
\begin{align*}
\left\Vert \hat{\vv}\sqrt{\hat{\lambda}}-\vv^{*}\sqrt{\lambda^{*}}\right\Vert_2 
& \le\sqrt{2}\sqrt{\lambda^{*}}\left|\sin\angle(\hat{\vv},\vv^{*})\right|+3\left|\sqrt{\hat{\lambda}}-\sqrt{\lambda^{*}}\right|\\
 & \le\sqrt{2}\min\left\{ \frac{2\delta}{\sqrt{\lambda^{*}} },\sqrt{\lambda^{*}} \right\} +6\min\left\{ \frac{\delta}{\sqrt{\lambda^{*}} },\sqrt{\delta}\right\} \\
 & \le10\min\left\{ \frac{\delta}{\sqrt{\lambda^{*}} },\sqrt{\delta}\right\} ,
\end{align*}
where we use (\ref{eq:eval_bound}) and (\ref{eq:evec_bound}) in
the second inequality.
We can now use this perturbation result to provide guarantees on recovering $\bbeta_1^{\ast}$ and $\bbeta_2^{\ast}$ given noisy versions of $\g^{\ast}$ and $\K^{\ast}$. To this end, suppose we are given $\hat{\K}$ and $\hat{\g}$ which satisfy
\begin{align*}
\left\Vert \hat{\K}-\K^{*}\right\Vert _{F}  \le\delta_{K},\qquad
\left\Vert \hat{\g}-\g^{*}\right\Vert _{2}  \le\delta_{g}.
\end{align*}
Then by triangle inequality we have
\[
\left\Vert \hat{\J}-\J^{*}\right\Vert _{F}\le\delta_{K}+2\delta_{g}\left\Vert \g^{*}\right\Vert _{2}+\delta_{g}^{2}.
\]
Therefore, up to relabeling $ b $, we have
\begin{align}
\left\Vert \hat{\bbeta}_b-\bbeta^{*}_b\right\Vert_2  & \le\left\Vert \hat{\g}-\g^{*}\right\Vert_2 +\left\Vert \sqrt{\hat{\lambda}}\hat{\vv}-\sqrt{\lambda^{*}}\vv^{*}\right\Vert_2 \nonumber\\
 & \lesssim\delta_{g}+\min\left\{ \frac{\delta_{K}+2\delta_{g}\left\Vert \g^{*}\right\Vert _{2}+\delta_{g}^{2}}{\left\Vert \bbeta_{1}^{*}-\bbeta_{2}^{*}\right\Vert _{2}},\sqrt{\delta_{K}+2\delta_{g}\left\Vert \g^{*}\right\Vert _{2}+\delta_{g}^{2}}\right\},\label{eq:beta_bound}
\end{align}
where the second inequality follows from Lemma~\ref{lem:perturb} and  $ \lambda^* = \frac{1}{4}\Vert \bbeta^*_1 - \bbeta^*_2\Vert_2^2 $.

We shall apply this result to the optimal solution $(\hat{\K},\hat{\g})$ obtained in the arbitrary noise setting, and in the stochastic noise setting, and thus prove Theorems \ref{thm:determ_beta} and \ref{thm:random_beta}.

\subsection{Proof of Theorem \ref{thm:determ_beta} (Arbitrary Noise)}

In the case of arbitrary noise, as set up above, Theorem \ref{thm:determ} guarantees the following:
\begin{align*}
\delta_{K} & \asymp\frac{\sqrt{n}\left\Vert \e\right\Vert _{2}\left\Vert \bbeta_{2}^{*}-\bbeta_{1}^{*}\right\Vert _{2}+\left\Vert \e\right\Vert _{2}^{2}}{\sqrt{\alpha}n}\lesssim\frac{1}{\sqrt{\alpha}}\frac{\left\Vert \e\right\Vert _{2}}{\sqrt{n}}\left\Vert \bbeta_{1}^{*}-\bbeta_{2}^{*}\right\Vert ,\\
\delta_{g} & \asymp\frac{\sqrt{n}\left\Vert \e\right\Vert _{2}\left\Vert \bbeta_{2}^{*}-\bbeta_{1}^{*}\right\Vert _{2}+\left\Vert \e\right\Vert _{2}^{2}}{\sqrt{\alpha}n\left(\left\Vert \bbeta_{1}^{*}\right\Vert _{2}+\left\Vert \bbeta_{2}^{*}\right\Vert _{2}\right)}\lesssim\frac{\left\Vert \e\right\Vert _{2}}{\sqrt{n}}.
\end{align*}
where we use the assumption $\left\Vert \e\right\Vert _{2}\le\frac{\sqrt{\alpha}}{c_{4}}\sqrt{n}\left(\left\Vert \bbeta_{1}^{*}\right\Vert_2 +\left\Vert \bbeta_{2}^{*}\right\Vert_2 \right)\asymp\frac{1}{c_{4}}\sqrt{n}\left\Vert \bbeta_{1}^{*}-\bbeta_{2}^{*}\right\Vert_2 $.
Using~\eqref{eq:beta_bound}, we get that up to relabeling $ b $,
\begin{align*}
\left\Vert \hat{\bbeta}_b-\bbeta^{*}_b\right\Vert_2  
& \lesssim\frac{\left\Vert \e\right\Vert_{2}}{\sqrt{n}}+\min\left\{ \frac{1}{\sqrt{\alpha}}\frac{\left\Vert \e\right\Vert _{2}}{\sqrt{n}}+\frac{\left\Vert \e\right\Vert _{2}^{2}}{n\left\Vert \bbeta_{1}^{*}-\bbeta_{2}^{*}\right\Vert_2 },\sqrt{\frac{1}{\sqrt{\alpha}}\frac{\left\Vert \e\right\Vert _{2}}{\sqrt{n}}\left\Vert \bbeta_{1}^{*}-\bbeta_{2}^{*}\right\Vert_2 +\frac{\left\Vert \e\right\Vert _{2}^{2}}{n}}\right\} \\
 & \lesssim\frac{1}{\sqrt{\alpha}}\frac{\left\Vert \e\right\Vert _{2}}{\sqrt{n}}+\min\left\{ \frac{\left\Vert \e\right\Vert _{2}^{2}}{n\left\Vert \bbeta_{1}^{*}-\bbeta_{2}^{*}\right\Vert_2 },\sqrt{\frac{1}{\sqrt{\alpha}}\frac{\left\Vert \e\right\Vert _{2}}{\sqrt{n}}\left\Vert \bbeta_{1}^{*}-\bbeta_{2}^{*}\right\Vert_2 }\right\} \\
 & \le\frac{1}{\sqrt{\alpha}}\frac{\left\Vert \e\right\Vert _{2}}{\sqrt{n}}.
\end{align*}

\subsection{Proof of Theorem \ref{thm:random_beta} (Stochastic Noise)}

Next consider the setting with stochastic noise. Under the assumption of Theorem \ref{thm:random_beta}, Theorem~\ref{thm:rand} guarantees the following bounds on the errors in recovering $ \K^* $ and $ \g^* $:
\begin{align*}
\delta_{K} & \asymp\sigma\left(\left\Vert \bbeta_{1}^{*}\right\Vert_2 +\left\Vert \bbeta_{2}^{*}\right\Vert_2 +\sigma\right)\sqrt{\frac{p}{n}}\log^{4}n,\\
\delta_{g} & \asymp\sigma\sqrt{\frac{p}{n}}\log^{4}n.
\end{align*}
If we let $ \gamma = \Vert \beta^*_1\Vert_2 + \Vert \beta^*_2\Vert_2 $, then this means
\begin{align*}
 \delta_K + 2\delta_g\|\g^*\|_2 + \delta_g^2 
\asymp & \sigma\gamma\sqrt{\frac{p}{n}}\log^4n + \sigma^2\sqrt{\frac{p}{n}}\log^4n + \sigma^2\frac{p}{n}\log^8n \\
\lesssim & \sigma\gamma\sqrt{\frac{p}{n}}\log^4n + \sigma^2\sqrt{\frac{p}{n}}\log^4n,
\end{align*}
where last inequality follows from the assumption that $n \geq p\log^8n$ for some $c > 1$.
Combining these with~\eqref{eq:beta_bound}, we obtain that up to relabeling of $ b $,
\begin{align*}
\label{uppbeta}
\left\Vert \hat{\bbeta}_b-\bbeta^{*}_b\right\Vert_2  
 & \lesssim \sigma\sqrt{\frac{p}{n}}\log^4n + \min\left\{ \frac{\sigma \gamma \sqrt{\frac{p}{n}}+\sigma^{2}\sqrt{\frac{p}{n}}}{\sqrt{\alpha}\gamma },\sqrt{\sigma \gamma \sqrt{\frac{p}{n}}+\sigma^{2}\sqrt{\frac{p}{n}}}\right\}\log^4n \notag\\
 & \lesssim \sigma\sqrt{\frac{p}{n}}\log^4n+\min\left\{ \frac{\sigma^{2}\sqrt{\frac{p}{n}}}{\gamma },\sqrt{\sigma \gamma \sqrt{\frac{p}{n}}+\sigma^{2}\sqrt{\frac{p}{n}}}\right\}\log^4n, 
\end{align*}
where the last inequality follows from $ \alpha  $ being lower-bounded by a constant. Observe that the minimization in the last RHS is no larger than $ \sigma\sqrt{\frac{p}{n}}  $ if $ \gamma \ge \sigma $, and equals $ 
\min\left\{ \frac{\sigma^{2}\sqrt{\frac{p}{n}}}{\gamma },\sigma\left(\frac{p}{n}\right)^{1/4}\right\}$ if $ \gamma < \sigma. $
It follows that
\[
\left\Vert \hat{\bbeta}_b-\bbeta^{*}_b\right\Vert_2 \lesssim \sigma\sqrt{\frac{p}{n}}\log^4n+\min\left\{ \frac{\sigma^{2}\sqrt{\frac{p}{n}}}{\gamma },  \sigma\left(\frac{p}{n}\right)^{1/4} \right\}\log^4n.
\]

\section{Proof of Theorem \ref{thm:determ}}
\label{sec:proofdeterm}
We now fill in the details for the proof outline given in Section~\ref{ssec:determoutline}, and complete the proof of Theorem~\ref{thm:determ} for the arbitrary noise setting. Some of the more technical or tedious proofs are relegated to the appendix. As in the proof outline, we assume the optimal solution to the optimization is $(\hat{\K},\hat{\g})=(\K^{*}+\hat{\H},\g^{*}+\hat{\h})$, and recall that $\hat{\H}_{T}:=\mathcal{P}_{T}\hat{\H}$ and $\hat{\H}_{T}^{\bot}:=\mathcal{P}_{T^{\bot}}\hat{\H}$. Note that $\hat{\H}_{T}$ has rank at most $4$ and $\hat{\H}_{T}^{\bot}$ has rank at most $p-4$. We have
\begin{align}
\left\Vert \hat{\K}\right\Vert _{*}-\left\Vert \K^{*}\right\Vert _{*} & \ge\left\Vert \K^{*}+\hat{\H}_{T}^{\bot}\right\Vert _{*}-\left\Vert \hat{\H}_{T}\right\Vert _{*}-\left\Vert \K^{*}\right\Vert _{*}=\left\Vert \hat{\H}_{T}^{\bot}\right\Vert _{*}-\left\Vert \hat{\H}_{T}\right\Vert _{*}.\label{eq:nuclear_decompose}
\end{align}

\subsection{Step (1): Consequence of Feasibility}
This step uses feasibility of the solution, to get a bound on $\BB$ in terms of the error parameter $\eta$.

For any $\left(\K,\g\right)=(\K^{*}+\H,\g^{*}+\h)$, it is easy to check that
\begin{equation}
-\left\langle \x_{b,i}\x_{b,i}^{\top},\K\right\rangle +2y_{b,i}\left\langle \x_{b,i},\g\right\rangle -y_{b,i}^{2}=-\left\langle \x_{b,i}\x_{b,i}^{\top},\H\right\rangle +2y_{b,i}\left\langle \x_{b,i},\h\right\rangle -e_{b,i}\x_{b,i}^{\top}\ddelta_{b}^{*}-e_{b,i}^{2}.\label{eq:expand}
\end{equation}
 Therefore, the constraint (\ref{eq:constraint}) is equivalent to
\[
\sum_{b=1}^{2}\sum_{i=1}^{n_{b}}\left|-\left\langle \x_{b,i}\x_{b,i}^{\top},\H\right\rangle +2\left(\x_{b,i}^{\top}\bbeta_{b}^{*}+e_{b,i}\right)\left\langle \x_{b,i},\h\right\rangle -e_{b,i}\x_{b,i}^{\top}\ddelta_{b}^{*}-e_{b,i}^{2}\right|\le\eta.
\]
Using the notation from Section~\ref{sub:notation}, this can be rewritten as
\begin{equation}
\sum_{b}\left\Vert n_{b}\AA_{b}\left(-\H+2\bbeta_{b}^{*}\h^{\top}\right)+2\e_{b}\circ\left(\X_{b}\h\right)-\e_{b}\circ\left(\X_{b}\ddelta_{b}^{*}\right)-\e_{b}^{2}\right\Vert _{1}\le\eta,\label{eq:equivalent_constraint}
\end{equation}
where $\circ$ denotes the element-wise product and $\e_{b}^{2}=\e_{b}\circ\e_{b}$.

First, note that $\K^{*}$ and $\g^{*}$ are feasible. By standard bounds on the spectral norm of random matrices~\cite{vershynin2010nonasym}, we know that with probability at least $1-2\exp(-cn_{b})$,
\[
\left\Vert \X_{b}\z\right\Vert _{2}\lesssim\sqrt{n_{b}}\left\Vert \z\right\Vert _{2},\forall\z\in\mathbb{R}^{p}.
\]
We thus have
\begin{align*}
\left\Vert -\e_{b}\circ\left(\X_{b}\ddelta_{b}^{*}\right)-\e_{b}^{2}\right\Vert _{1} & \le c_{1}\left(\sqrt{n_{b}}\left\Vert \e_{b}\right\Vert _{2}\left\Vert \ddelta_{b}^{*}\right\Vert _{2}+\left\Vert \e\right\Vert _{2}^{2}\right)\\
 & \overset{(a)}{\le}c_{1}\sqrt{n_{b}}\left\Vert \e\right\Vert _{2}\left\Vert \bbeta_{1}^{*}-\bbeta_{2}^{*}\right\Vert _{2}\overset{(b)}{\le}\eta,
\end{align*}
where we use the assumptions on $\e$ and $\eta$ in (a) and (b), respectively. This implies that (\ref{eq:equivalent_constraint}) holds with $\H=\zero$ and $\h=\zero$, thus showing the feasibility of $(\K^{*},\g^{*})$.

Since $\left(\hat{\K},\hat{\g}\right)$ is feasible by assumption, combining the last two display equations and (\ref{eq:equivalent_constraint}), we further have 
\begin{align}
\sum_{b}\left\Vert n_{b}\AA_{b}\left(-\hat{\H}+2\bbeta_{b}^{*}\hat{\h}^{\top}\right)\right\Vert _{1} & \le\sum_{b}\left\Vert 2\e_{b}\circ\left(\X_{b}\hat{\h}\right)\right\Vert _{1}+\sum_{b}\left\Vert -2\e_{b}\circ\left(\X_{b}\ddelta_{b}^{*}\right)-\e_{b}^{2}\right\Vert _{1}+\eta\nonumber \\
 & \le c_{2}\sum_{b}\sqrt{n_{b}}\left\Vert \e_{b}\right\Vert _{2}\|\hat{\h}\| _{2}+2\eta.\label{eq:from_optimality_determ}
\end{align}
Now from the definition of $\AA_{b}$ and $\BB_{b}$, we have
\begin{align*}
\left\lfloor n_{b}/2\right\rfloor \left\Vert \BB_{b}\left(-\hat{\H}+2\bbeta_{b}^{*}\hat{\h}^{\top}\right)\right\Vert _{1} & \le\sum_{j=1}^{\left\lfloor n_{b}/2\right\rfloor }\left\Vert \left\langle \A_{b,2j},-\hat{\H}+2\bbeta_{b}^{*}\hat{\h}^{\top}\right\rangle \right\Vert _{1}+\left\Vert \left\langle \A_{b,2j-1},-\hat{\H}+2\bbeta_{b}^{*}\hat{\h}^{\top}\right\rangle \right\Vert _{1}\\
 & \le n_{b}\left\Vert \AA_{b}\left(-\hat{\H}+2\bbeta_{b}^{*}\hat{\h}^{\top}\right)\right\Vert _{1}.
\end{align*}
It follows from (\ref{eq:from_optimality_determ}) and $n_{1}\asymp n_{2}\asymp n$ that 
\begin{align}
\sum_{b}n\left\Vert \BB_{b}\left(-\hat{\H}+2\bbeta_{b}^{*}\hat{h}^{\top}\right)\right\Vert _{1}-c_{2}\sum_{b}\sqrt{n}\left\Vert \e_{b}\right\Vert _{2}\|\hat{\h}\| _{2} \leq 2 \eta. \label{eq:determ1}
\end{align} 
This concludes Step (1) of the proof.

\subsection{Step (2): RIP and Lower Bounds}

The bound in (\ref{eq:determ1}) relates the $\ell_1$-norm of $\BB$ and $\eta$. Since we want a bound on the $\ell_2$ and Frobenius norms of $\hat{\h}$ and $\hat{\H}$ respectively, a major step is the proof of an RIP-like property for $\BB$:
\begin{lemma}
\label{lem:RIP}
The following holds for some numerical constants $c,\underbar{\ensuremath{\delta}},\bar{\delta}$.
For $b=1,2$, if $\mu>1$ and $n_{b}\ge c\rho p$, then with probability
$1-\exp(-n_{b})$, we have the following:
\[
\underbar{\ensuremath{\delta}}\left\Vert \Z\right\Vert _{F}\le\left\Vert \BB_{b}\Z\right\Vert _{1}\le\bar{\delta}\left\Vert \Z\right\Vert _{F},\quad\forall\Z\in\mathbb{R}^{p\times p}\textrm{ with rank}(\Z)\le\rho.
\]

\end{lemma}
We defer the proof of this lemma to the appendix, where in fact we show it is a special case of a similar result we use in Section \ref{sec:proofrand}. 

We now turn to the implications of this lemma, in order to get lower bounds on the term $\left\Vert \BB_{b}\left(-\hat{\H}+2\bbeta_{b}^{*}\h^{\top}\right)\right\Vert _{1}$ from the first term in (\ref{eq:determ1}), in terms of $\|\hat{\h}\|_2$ and $\|\hat{\H}\|_F$.

Since we have proved that $(\K^{*},\g^{*})$ is feasible, we have $\left\Vert \hat{\K}\right\Vert _{*}\le\left\Vert \K^{*}\right\Vert _{*}$
by optimality. It follows from (\ref{eq:nuclear_decompose}) that 
\begin{equation}
\left\Vert \hat{\H}_{T}^{\bot}\right\Vert _{*}\le\left\Vert \hat{\H}_{T}\right\Vert _{*}.\label{eq:cone_determ}
\end{equation}
Let $K=c\frac{1}{\alpha}$ for $c$ some numeric constant to be chosen later. We can partition $\hat{\H}_{T}^{\bot}$ into a sum of $M:=\frac{p-4}{K}$ matrices $\hat{\H}_{1},\ldots,\hat{\H}_{M}$ according to the SVD of $\hat{\H}_{T}^{\bot}$, such that $\text{rank}(\hat{\H}_{i})\le K$ and the smallest singular value of $\hat{\H}_{i}$ is larger than the largest singular value of $\hat{\H}_{i+1}$ (cf.~\cite{recht2010guaranteed}). By Lemma \ref{lem:RIP}, we get that for each $b=1,2$,
\begin{equation}
\sum_{i=2}^{M}\left\Vert \BB_{b}(\hat{\H}_{i})\right\Vert _{1}\le\bar{\delta}\sum_{i=2}^{M}\left\Vert \hat{\H}_{i}\right\Vert _{F}\le\bar{\delta}\sum_{i=2}^{M}\frac{1}{\sqrt{K}}\left\Vert \hat{\H}_{i-1}\right\Vert _{*}\le\frac{\bar{\delta}}{\sqrt{K}}\left\Vert \hat{\H}_{T^{\bot}}\right\Vert _{*}\overset{(a)}{\le}\frac{\bar{\delta}}{\sqrt{K}}\sqrt{4}\left\Vert \hat{\H}_{T}\right\Vert _{F},\label{eq:Tperp}
\end{equation}
where $(a)$ follows from (\ref{eq:cone_determ}) and the rank of
$\hat{\H}_{T}$. It follows that for $b=1,2$, 
\begin{align*}
\left\Vert \BB_{b}\left(\hat{\H}-2\bbeta_{b}^{*}\hat{h}^{\top}\right)\right\Vert _{1} & \overset{(a)}{\ge}\left\Vert \BB_{b}\left(\hat{\H}_{T}+\hat{\H}_{1}-2\bbeta_{b}^{*}\hat{\h}^{\top}\right)\right\Vert _{1}-\sum_{i=2}^{M}\left\Vert \BB(\hat{\H}_{i})\right\Vert _{1}\\
 & \overset{(b)}{\ge}\underbar{\ensuremath{\delta}}\left\Vert \hat{\H}_{T}+\hat{\H}_{1}-2\bbeta_{b}^{*}\hat{\h}^{\top}\right\Vert _{F}-2\bar{\delta}\sqrt{\frac{1}{K}}\left\Vert \hat{\H}_{T}\right\Vert _{F}\\
 & \overset{(c)}{\ge}\underbar{\ensuremath{\delta}}\left\Vert \hat{\H}_{T}-2\bbeta_{b}^{*}\hat{\h}^{\top}\right\Vert _{F}+\left\Vert \hat{\H}_{1}\right\Vert _{F}-2\bar{\delta}\sqrt{\frac{1}{K}}\left\Vert \hat{\H}_{T}\right\Vert _{F}\\
 & \ge\underbar{\ensuremath{\delta}}\left\Vert \hat{\H}_{T}-2\bbeta_{b}^{*}\hat{\h}^{\top}\right\Vert _{F}-2\bar{\delta}\sqrt{\frac{1}{K}}\left\Vert \hat{\H}_{T}\right\Vert _{F},
\end{align*}
where $(a)$ follows from the triangle inequality, $(b)$ follows from Lemma \ref{lem:RIP} and (\ref{eq:Tperp}), and $(c)$ follows from the fact that $\hat{\H}_{T}-\beta_{b}\hat{\h}^{\top}\in T$ and $\hat{\H}_{1}\in T^{\bot}$. Summing the above inequality for $b=1,2$, we obtain
\begin{equation}
\sum_{b}\left\Vert \BB_{b}\left(\hat{\H}-2\bbeta_{b}^{*}\hat{\h}^{\top}\right)\right\Vert _{1}\ge\underbar{\ensuremath{\delta}}\sum_{b}\left\Vert \hat{\H}_{T}-2\bbeta_{b}^{*}\hat{\h}^{\top}\right\Vert _{F}-4\bar{\delta}\sqrt{\frac{1}{K}}\left\Vert \hat{\H}_{T}\right\Vert _{F}.\label{eq:squeeze}
\end{equation}
The first term in the RHS of (\ref{eq:squeeze}) can be bounded using the following lemma, whose proof is deferred to the appendix.
\begin{lemma}
\label{claim}We have
\begin{align*}
\sum_{b}\left\Vert \hat{\H}_{T}-2\bbeta_{b}^{*}\hat{\h}^{\top}\right\Vert _{F} & \ge\sqrt{\alpha}\left\Vert \hat{\H}_{T}\right\Vert _{F},\\
\sum_{b}\left\Vert \hat{\H}_{T}-2\bbeta_{b}^{*}\hat{\h}^{\top}\right\Vert _{F} & \ge\sqrt{\alpha}\left(\left\Vert \bbeta_{1}^{*}\right\Vert _{2}+\left\Vert \bbeta_{2}^{*}\right\Vert _{2}\right)\|\hat{\h}\| _{2}.
\end{align*}

\end{lemma}
Combining (\ref{eq:squeeze}) and the lemma, we obtain
\begin{align*}
\sum_{b}\left\Vert \BB_{b}\left(\hat{\H}-2\bbeta_{b}^{*}\hat{\h}^{\top}\right)\right\Vert _{1} & \ge\left(\underbar{\ensuremath{\delta}}\sqrt{\alpha}-4\bar{\delta}\sqrt{\frac{1}{K}}\right)\left\Vert \hat{\H}_{T}\right\Vert _{F}
\end{align*}
and
\begin{align*}
\sum_{b}\left\Vert \BB_{b}\left(\hat{\H}-2\bbeta_{b}^{*}\hat{\h}^{\top}\right)\right\Vert _{1} & \ge\left(\underbar{\ensuremath{\delta}}-4\bar{\delta}\sqrt{\frac{1}{\alpha K}}\right)\sum_{b}\left\Vert \hat{\H}_{T}-\bbeta_{b}\hat{\h}^{\top}\right\Vert _{F}\\
&\ge\left(\underbar{\ensuremath{\delta}}-4\bar{\delta}\sqrt{\frac{1}{\alpha K}}\right)\sqrt{\alpha}\left(\left\Vert \bbeta_{1}^{*}\right\Vert _{2}+\left\Vert \bbeta_{2}^{*}\right\Vert _{2}\right)\|\hat{\h}\| _{2}.
\end{align*}
Recall that $K=c\frac{1}{\alpha}$. When $c$ is sufficiently large,
the above inequalities imply that for some numeric constant $c'$,
\begin{align}
\sum_{b}\left\Vert \BB_{b}\left(\hat{\H}-2\bbeta_{b}^{*}\hat{\h}^{\top}\right)\right\Vert _{1} & \ge\frac{\sqrt{\alpha}}{c''}\left\Vert \hat{\H}_{T}\right\Vert _{F}\overset{(d)}{\ge}\frac{\sqrt{\alpha}}{c'}\left\Vert \hat{\H}\right\Vert _{F},\label{eq:RSC_det_1}\\
\sum_{b}\left\Vert \BB_{b}\left(\hat{\H}-2\bbeta_{b}^{*}\hat{\h}^{\top}\right)\right\Vert _{1} & \ge\frac{\sqrt{\alpha}}{c'}\left(\left\Vert \bbeta_{1}^{*}\right\Vert _{2}+\left\Vert \bbeta_{2}^{*}\right\Vert _{2}\right)\|\hat{\h}\| _{2},\label{eq:RSC_det_2}
\end{align}
where the inequality (d) follows from (\ref{eq:cone_determ}) and $\text{rank}(\hat{\H}_{T})\le4$. This concludes the proof of Step (2).

\subsection{Step (3): Producing Error Bounds}
We now combine the result of the three steps, in order to obtain bounds on $\|\hat{\h}\|_2$ and $\|\hat{\H}\|_F$ in terms of $\eta$, and the other parameters of the problem, hence concluding the proof of Theorem \ref{thm:determ}. 

From Step (1), we concluded the bound (\ref{eq:determ1}), which we reproduce:
$$
\sum_{b}n\left\Vert \BB_{b}\left(-\hat{\H}+2\bbeta_{b}^{*}\hat{\h}^{\top}\right)\right\Vert _{1}-c_{2}\sum_{b}\sqrt{n}\left\Vert \e_{b}\right\Vert _{2}\|\hat{\h}\| _{2} \leq 2 \eta. 
$$
Applying (\ref{eq:RSC_det_2}) to the LHS above, we get 
\begin{align*}
\sqrt{n}\sum_{b}\left(\sqrt{\alpha}\sqrt{n}\left\Vert \bbeta_{b}^{*}\right\Vert _{2}-\left\Vert \e_{b}\right\Vert _{2}\right)\|\hat{\h}\| _{2} \lesssim 2 \eta.
\end{align*}
Under the assumption $\left\Vert \e\right\Vert_{2}\le\frac{1}{c_{5}}\sqrt{\alpha}\sqrt{n}\left(\left\Vert \bbeta_{1}^{*}\right\Vert _{2}+\left\Vert \bbeta_{2}^{*}\right\Vert_{2}\right)$
for some $c_{5}$ sufficiently large, we obtain the following bound
for $\|\hat{\h}\|_2$:
\[
\|\hat{\h}\| _{2}\lesssim\frac{1}{\sqrt{\alpha}n\left(\left\Vert \bbeta_{1}^{*}\right\Vert _{2}+\left\Vert \bbeta_{2}^{*}\right\Vert _{2}\right)}\eta.
\]

To obtain a bound on $\left\Vert \hat{\H}\right\Vert _{F}$, we note
that 
\[
\sum_{b}\left\Vert \e_{b}\right\Vert _{2}\|\hat{\h}\| _{2}\le\frac{1}{c_{5}}\sqrt{n}\sum_{b}\sqrt{\alpha}\left\Vert \bbeta_{b}^{*}\right\Vert _{2}\|\hat{\h}\| _{2}\le\frac{c'}{c_{5}}\sqrt{n}\sum_{b}\left\Vert \BB_{b}\left(\hat{\H}-2\bbeta_{b}^{*}\hat{\h}^{\top}\right)\right\Vert _{1},
\]
where we use the assumption on $\left\Vert \e\right\Vert $ and (\ref{eq:RSC_det_2})
in the two inequalities, respectively. When $c_{5}$ is large, we
combine the last display equation with (\ref{eq:determ1}) to obtain
\begin{align*}
n\sqrt{\alpha}\left\Vert \hat{\H}\right\Vert _{F} \lesssim n\sum_{b}\left\Vert \BB_{b}\left(\hat{\H}_{T}-2\bbeta_{b}^{*}\hat{\h}^{\top}\right)\right\Vert _{1} \lesssim 2 \eta,
\end{align*}
where we use (\ref{eq:RSC_det_1}) in the last inequality. This implies
\[
\left\Vert \hat{\H}\right\Vert _{F}\lesssim\frac{1}{n\sqrt{\alpha}}\eta,
\]
completing the proof of Step (3) and thus Theorem \ref{thm:determ}.

\section{Proof of Theorem \ref{thm:rand}}
\label{sec:proofrand}
We follow the three steps from the proof outline in Section \ref{ssec:stochasticoutline}, to give the proof of Theorem \ref{thm:rand} for the stochastic noise setting. We continue to use the notation given in Section \ref{sub:notation}. For each $b=1,2$, we define the vector $\dd_{b,j}=e_{b,2j}\x_{b,2j}-e_{b,2j-1}\x_{b,2j-1}$ for $j=1,\ldots,\left\lfloor n_{b}/2\right\rfloor $, as well as the vectors $\cc_{b,i}:=y_{b,i}\x_{b,i}$ for $i\in[n_{b}]$. We let $\D_{b}:=\left(\left\lfloor n_{b}/2\right\rfloor \right)^{-1}\left[\dd_{b,1},\ldots,\dd_{b,\left\lfloor n_{b}/2\right\rfloor }\right]^{\top}\in\mathbb{R}^{\left\lfloor n_{b}/2\right\rfloor \times p}$. We also define the shorthand
\begin{align*}
\gamma & :=\left\Vert \bbeta_{1}^{*}\right\Vert _{2}+\left\Vert \bbeta_{2}^{*}\right\Vert _{2}.
\end{align*}

Since the $\left\{ \x_{i}\right\} $ are assumed to be Gaussian with i.i.d. entries, the statement of the theorem is invariant under rotation of the $\bbeta_{b}^{*}$'s. Therefore, it suffices to prove the theorem assuming $\bbeta_{1}^{*}-\bbeta_{2}^{*}$ is supported on the first coordinate.
The follow lemma shows that we can further assume $\left\{ \x_{i}\right\} $ and $\e$ have bounded entries, since we are interested in results that hold with high probability. This simplifies the subsequent analysis.
\begin{lemma}
\label{lem:truncation}There exists an absolute constant $c>0$ such
that, if the conclusion of Theorem \ref{thm:rand} holds w.h.p. with
the additional assumption that 
\begin{align*}
\x_{i}(l) & \le c\sqrt{\log n},\forall i\in[n],l\in[p],\\
e_{i} & \le c\sigma\sqrt{\log n},\forall i\in[n],
\end{align*}
then it also holds w.h.p. without this assumption.
\end{lemma}
We prove this lemma in the appendix. In the sequel, we therefore assume $\text{support}\left(\bbeta_{1}^{*}-\bbeta_{2}^{*}\right)=\{1\}$, and the $\left\{ \x_{i}\right\} $ and $\{\e_{i}\}$ satisfy the bounds in the above lemma.

\subsection{Step (1): Consequence of Optimality}
This step uses optimality of the solution $(\hat{\K},\hat{\g}`)=(\K^{*}+\hat{\H},\g^{*}+\hat{\h})$, to get a bound on $\AA$. By optimality, we have
\begin{align*}
&\sum_{i=1}^{n}\left(-\left\langle \x_{i}\x_{i}^{\top},\hat{\K}\right\rangle +2y_{i}\left\langle \x_{i},\hat{\g}\right\rangle -y_{i}^{2}+\sigma^{2}\right)^{2}+\lambda\left\Vert \hat{\K}\right\Vert _{*}\\
\le&\sum_{i=1}^{n}\left(-\left\langle \x_{i}\x_{i}^{\top},\K^{*}\right\rangle +2y_{i}\left\langle \x_{i},\g^{*}\right\rangle -y_{i}^{2}+\sigma^{2}\right)^{2}+\lambda\left\Vert \K^{*}\right\Vert _{*}.
\end{align*}
Using the expression (\ref{eq:expand}), we have
\begin{align*}
&\sum_{i=1}^{n}\left(-\left\langle \x_{i}\x_{i}^{\top},\hat{\H}\right\rangle +2(\x_{i}^{\top}\bbeta_{b}^{*}+e_{i})\left\langle \x_{i},\hat{\h}\right\rangle -e_{i}\x_{i}^{\top}\ddelta_{b}^{*}-(e_{i}^{2}-\sigma^{2})\right)^{2}+\lambda\left\Vert \hat{\K}\right\Vert _{*}\\
\le&\sum_{i=1}^{n}\left(-e_{i}\x_{i}^{\top}\ddelta_{b}^{*}-(e_{i}^{2}-\sigma^{2})\right)^{2}+\lambda\left\Vert \K^{*}\right\Vert _{*}.
\end{align*}
Defining the noise vectors $\w_{1,b}:=-\e_{b}\circ\left(\X\ddelta_{b}^{*}\right)$, $\w_{2,b}:=-\left(\e_{b}^{2}-\sigma^{2}\mathbf{1}\right)$ and $\w_{b}=\w_{1,b}-\w_{2,b}$, we can rewrite the display equation above as 
\[
\sum_{b}\left\Vert n_{b}\AA_{b}\left(-\hat{\H}+2\bbeta_{b}^{*}\hat{\h}^{\top}\right)+2\e_{b}\circ(\X_{b}\hat{\h})+\w_{b}\right\Vert _{2}^{2}+\lambda\left\Vert \hat{\K}\right\Vert _{*}\lesssim\sum_{b=1,2}\left\Vert \w_{b}\right\Vert _{2}^{2}+\lambda\left\Vert \hat{\K}\right\Vert _{*}.
\]
Expanding the squares and rearranging terms, we obtain
\begin{align*}
 & \sum_{b}\left\Vert n_{b}\AA_{b}\left(-\hat{\H}+2\bbeta_{b}^{*}\hat{\h}^{\top}\right)+2\e_{b}\circ(\X_{b}\hat{\h})\right\Vert _{2}^{2}\\
\le & \sum_{b}\left\langle -\hat{\H}+2\bbeta_{b}^{*}\hat{\h}^{\top},n_{b}\AA_{b}^{*}\w_{b}\right\rangle +\sum_{b}\left\langle \hat{\h},2\X_{b}^{\top}\text{diag}(\e_{b})\w_{b}\right\rangle +\lambda\left(\left\Vert \K^{*}\right\Vert _{*}-\left\Vert \hat{\K}\right\Vert _{*}\right) \\
\overset{(a)}{\le} & \left(\left\Vert \hat{\H}_{T}\right\Vert _{*}+\left\Vert \hat{\H}_{T}^{\bot}\right\Vert _{*}\right)\cdot P+\|\hat{\h}\| _{2}\cdot Q+\lambda\left(\left\Vert \K^{*}\right\Vert _{*}-\left\Vert \hat{\K}\right\Vert _{*}\right)\\
\overset{(b)}{\le} & \left(\left\Vert \hat{\H}_{T}\right\Vert _{*}+\left\Vert \hat{\H}_{T}^{\bot}\right\Vert _{*}\right)\cdot P+\|\hat{\h}\| _{2}\cdot Q+\lambda\left(\left\Vert \hat{\H}_{T}\right\Vert _{*}-\left\Vert \hat{\H}_{T}^{\bot}\right\Vert _{*}\right),
\end{align*}
where $\AA_{b}^{*}$ is the adjoint operator of $\AA_{b}$ and in
(a) we have defined 
\begin{align*}
P & :=2\sum_{b}\left\Vert n_{b}\AA_{b}^{*}\w_{b}\right\Vert ,\\
Q & :=\sum_{b}\left\Vert \bbeta_{b}^{*}\right\Vert _{2}\left\Vert n_{b}\AA_{b}^{*}\w_{b}\right\Vert +\sqrt{p}\left\Vert \sum_{b}2\X_{b}^{\top}\text{diag}(\e_{b})\w_{b}\right\Vert _{\infty},
\end{align*}
and (b) follows from (\ref{eq:nuclear_decompose}). We need the following lemma, which bounds the noise terms $ P $ and $ Q $. Its proof is a substantial part of the proof to the main result, but quite lengthy. We therefore defer it to Section \ref{sub:proof_noise_bound}.
\begin{lemma}
\label{lem:noise_bound_rand}Under the assumption of the theorem,
we have $\lambda\ge2P$ and $\lambda\ge\frac{1}{\sigma+\gamma}Q$
with high probability. 
\end{lemma}
Applying the lemma, we get
\begin{equation}
\sum_{b}\left\Vert n_{b}\AA_{b}\left(-\hat{\H}+2\bbeta_{b}^{*}\hat{\h}^{\top}\right)+2\e_{b}\circ(\X_{b}\hat{\h})\right\Vert _{2}^{2}\le\lambda\left(\frac{3}{2}\left\Vert \hat{\H}_{T}\right\Vert _{*}-\frac{1}{2}\left\Vert \hat{\H}_{T}^{\bot}\right\Vert _{*}\right)+\lambda\left(\gamma+\sigma\right)\|\hat{\h}\| _{2}.\label{eq:from_opt}
\end{equation}
Since the right hand side of (\ref{eq:from_opt}) is non-negative, we obtain the following cone constraint for the optimal solution:
\begin{equation}
\left\Vert \hat{\H}_{T}^{\bot}\right\Vert _{*}\le\frac{5}{2}\left\Vert \hat{\H}_{T}\right\Vert _{*}+\left(\gamma+\sigma\right)\|\hat{\h}\| _{2}.\label{eq:cone_rand}
\end{equation}

This concludes the proof of Step (1) of the proof. 

\subsection{Step (2): RIP and Lower Bounds}
We can get a lower bound to the expression in the LHS of (\ref{eq:from_opt}) using $\BB$, as follows. Similarly as before, let $K$ be some numeric constant to be chosen later; we partition $\hat{\H}_{T}^{\bot}$ into a sum of $M:=\frac{p-4}{K}$ matrices $\hat{\H}_{1},\ldots,\hat{\H}_{M}$ according to the SVD of $\hat{\H}_{T}^{\bot}$, such that $\text{rank}(\hat{\H}_{i})\le K$ and the smallest singular value of $\hat{\H}_{i}$ is larger than the largest singular value of $\hat{\H}_{i+1}$. Then we have the following chain of inequalities:
\begin{align}
 & \sum_{b}\left\Vert n_{b}\AA_{b}\left(-\hat{\H}+2\bbeta_{b}^{*}\hat{\h}^{\top}\right)+2\e_{b}\circ\left(\X_{b}\hat{\h}\right)\right\Vert _{2}^{2} \nonumber \\
\overset{(a)}{\ge} & \sum_{b}\left\Vert n_{b}\BB_{b}\left(-\hat{\H}+2\bbeta_{b}^{*}\hat{\h}^{\top}\right)+2n_{b}\D_{b}\hat{\h}\right\Vert _{2}^{2} \nonumber \\
\overset{(b)}{\ge} & \sum_{b}n_{b}\left\Vert \BB_{b}\left(-\hat{\H}+2\bbeta_{b}^{*}\hat{\h}^{\top}\right)+2\D_{b}\hat{\h}\right\Vert _{1}^{2} \nonumber \\
\overset{(c)}{\gtrsim} & n\left(\sum_{b}\left\Vert \BB_{b}\left(-\hat{\H}+2\bbeta_{b}^{*}\hat{\h}^{\top}\right)+2\D_{b}\hat{\h}\right\Vert _{1}\right)^{2} \nonumber \\
\overset{(d)}{\ge} & n\left(\sum_{b}\left\Vert \BB_{b}\left(-\hat{\H}_{T}+2\bbeta_{b}^{*}\hat{\h}^{\top}+\hat{\H}_{1}\right)+2\D_{b}\hat{\h}\right\Vert _{1}-\sum_{b}\sum_{i=2}^{M}\left\Vert \BB_{b}(\hat{\H}_{i})\right\Vert _{1}\right)^{2}. \label{eq:last}
\end{align}
Here (a) follows from the definitions of $\AA_{b}$ and $\BB_{b}$ and the triangle inequality, (b) follows from $\left\Vert \u\right\Vert _{2}\ge\frac{1}{n_{b}}\left\Vert \u\right\Vert _{1}$ for all $\u\in\mathbb{R}^{n_{b}}$, (c) follows from $n_{1}\approx n_{2}$, and (d) follows from the triangle inequality. 

We see that in order to obtain lower bounds on~\eqref{eq:last} in terms of $\|\hat{\h}\|_2$ and $\|\hat{\H}\|_F$, we need an extension of the previous RIP-like result from Lemma \ref{lem:RIP}, in order to deal with the first term in~(\ref{eq:last}). The following lemma is proved in the appendix.
\begin{lemma}
\label{lem:RIP2}The following holds for some numerical constants $c,\underbar{\ensuremath{\delta}},\bar{\delta}$. For $b=1,2$, if $\mu>1$ and $n_{b}\ge cpr$, then with probability $1-\exp(-n_{b})$, we have the following RIP-2:
\begin{align*}
\underbar{\ensuremath{\delta}}\left(\left\Vert \Z\right\Vert _{F}+\sigma\left\Vert \z\right\Vert _{2}\right)
\le\left\Vert \BB_{b}\Z-\D_{b}\z\right\Vert _{1}
&\le\bar{\delta}\left(\left\Vert \Z\right\Vert _{F}+\sigma\left\Vert \z\right\Vert _{2}\right),\\
&\forall\z\in\mathbb{R}^{p},\forall\Z\in\mathbb{R}^{p\times p}\textrm{ with rank}(\Z)\le r.
\end{align*}
\end{lemma}
Using this we can now bound the last inequality in (\ref{eq:last}) above. First, note that for each $b=1,2$,
\begin{equation}
\sum_{i=2}^{M}\left\Vert \BB_{b}(\hat{\H}_{i})\right\Vert _{1}\overset{(a)}{\le}\bar{\delta}\sum_{i=2}^{M}\left\Vert \hat{\H}_{i}\right\Vert _{F}\le\bar{\delta}\sum_{i=2}^{M}\frac{1}{\sqrt{K}}\left\Vert \hat{\H}_{i-1}\right\Vert _{*}\le\frac{\bar{\delta}}{\sqrt{K}}\left\Vert \hat{\H}_{T^{\bot}}\right\Vert _{*},\label{eq:Tperp2}
\end{equation}
where (a) follows from the upper bound in Lemma \ref{lem:RIP2} with $\sigma$ set to
$0$. Then, applying the lower-bound in Lemma \ref{lem:RIP2} to the first term in the parentheses in~\eqref{eq:last}, and (\ref{eq:Tperp2}) to the second term, we obtain
\begin{align*}
 & \sum_{b}\left\Vert n_{b}\AA_{b}\left(-\hat{\H}+2\bbeta_{b}^{*}\hat{\h}^{\top}\right)+2\e_{b}\circ\left(\X_{b}\hat{\h}\right)\right\Vert _{2}^{2}\\
\ge & n\left(\sum_{b}\underbar{\ensuremath{\delta}}\left\Vert \hat{\H}_{T}-2\bbeta_{b}^{*}\hat{\h}^{\top}\right\Vert _{F}+2\underbar{\ensuremath{\delta}}\sigma\|\hat{\h}\| _{2}-\bar{\delta}\sqrt{\frac{1}{K}}\left\Vert \hat{\H}_{T^{\bot}}\right\Vert _{*}\right)^{2}\\
\gtrsim & n\left(\sum_{b}\underbar{\ensuremath{\delta}}^{2}\left\Vert \hat{\H}_{T}-2\bbeta_{b}^{*}\hat{\h}^{\top}\right\Vert _{F}^{2}+\underbar{\ensuremath{\delta}}^{2}\sigma^{2}\|\hat{\h}\| _{2}^{2}-\bar{\delta}^{2}\frac{1}{K}\left\Vert \hat{\H}_{T^{\bot}}\right\Vert _{*}^{2}\right).
\end{align*}
Choosing $K$ to be sufficiently large, and applying Lemma \ref{claim},
we obtain 
\begin{align*}
\sum_{b}\left\Vert n_{b}\AA_{b}\left(-\hat{\H}+2\bbeta_{b}^{*}\hat{\h}^{\top}\right)+2\e_{b}\circ\left(\X_{b}\hat{\h}\right)\right\Vert _{2}^{2}\gtrsim & n\left(\left\Vert \hat{\H}_{T}\right\Vert _{F}^{2}+\gamma^{2}\|\hat{\h}\| _{2}^{2}+\sigma^{2}\|\hat{\h}\| _{2}^{2}-\frac{1}{100}\left\Vert \hat{\H}_{T^{\bot}}\right\Vert _{*}^{2}\right).
\end{align*}
Using (\ref{eq:cone_rand}), we further get
\begin{align}
 & \sum_{b}\left\Vert n_{b}\AA_{b}\left(-\hat{\H}+2\bbeta_{b}^{*}\hat{\h}^{\top}\right)+2\e_{b}\circ\left(\X_{b}\hat{\h}\right)\right\Vert _{2}^{2}\nonumber \\
\gtrsim & n\left[\left\Vert \hat{\H}_{T}\right\Vert _{F}^{2}+\gamma^{2}\|\hat{\h}\| _{2}^{2}+\sigma^{2}\|\hat{\h}\| _{2}^{2}-\frac{1}{8}\left\Vert \hat{\H}_{T}\right\Vert _{*}^{2}-\frac{1}{25}\left(\gamma^{2}+\sigma^{2}\right)\|\hat{\h}\| _{2}^{2}\right]\nonumber \\
\gtrsim & \frac{1}{8}n\left(\left\Vert \hat{\H}_{T}\right\Vert _{F}+\left(\gamma+\sigma\right)\|\hat{\h}\| _{2}\right)^{2}.\label{eq:RSC_rand}
\end{align}

This completes Step (2), and we are ready to combine the results to obtain error bounds, as promised in Step (3) and by the theorem.

\subsection{Step (3): Producing Error bounds}

Combining (\ref{eq:from_opt}) and (\ref{eq:RSC_rand}), we get
\begin{align*}
n\left(\left\Vert \hat{\H}_{T}\right\Vert _{F}+\left(\gamma+\sigma\right)\|\hat{\h}\| _{2}\right)^{2} & \lesssim\lambda\left\Vert \H_{T}\right\Vert _{F}+\lambda(\gamma+\sigma)\|\hat{\h}\| _{2},
\end{align*}
which implies $\left\Vert \hat{\H}_{T}\right\Vert _{F}+\left(\gamma+\sigma\right)\|\hat{\h}\| _{2}\lesssim\frac{\lambda}{n}.$
It follows that $\|\hat{\h}\| _{2}\lesssim\frac{1}{n\left(\gamma+\sigma\right)}\lambda$
and 
\begin{align*}
\left\Vert \hat{\H}\right\Vert _{F} & \le\left\Vert \hat{\H}_{T}\right\Vert _{*}+\left\Vert \hat{\H}_{T}^{\bot}\right\Vert _{*}\\
 & \overset{(a)}{\le}\frac{7}{2}\left\Vert \hat{\H}_{T}\right\Vert _{*}+\left(\gamma+\sigma\right)\|\hat{\h}\| _{2}\\
 & \overset{(b)}{\le}\frac{7}{2}\cdot\sqrt{4}\left\Vert \hat{\H}_{T}\right\Vert _{F}+\left(\gamma+\sigma\right)\|\hat{\h}\| _{2}\\
 & \lesssim\frac{1}{n}\lambda,
\end{align*}
where we use (\ref{eq:cone_rand}) in (a) and $\text{rank}\left(\hat{\H}_{T}\right)\le4$
in (b). This completes Step (3) and the proof of the theorem.

\subsection{Proof of Lemma \ref{lem:noise_bound_rand}\label{sub:proof_noise_bound}}
We now move to the proof of Lemma \ref{lem:noise_bound_rand}, which bounds the noise terms $ P $ and $ Q $.  Note that 
\[
P=2\sum_{b}\left\Vert n_{b}\AA_{b}^{*}\w_{b}\right\Vert \le\underbrace{2\sum_{b}\left\Vert n_{b}\AA_{b}^{*}\w_{1,b}\right\Vert }_{S_{1}}+\underbrace{2\sum_{b}\left\Vert n_{b}\AA_{b}^{*}\w_{2,b}\right\Vert }_{S_{2}},
\]
and
\begin{align*}
Q & =\sum_{b}\left\Vert \bbeta_{b}^{*}\right\Vert _{2}\left\Vert n_{b}\AA_{b}^{*}\w_{b}\right\Vert +\sqrt{p}\left\Vert \sum_{b}2\X_{b}^{\top}\text{diag}(\e_{b})\w_{b}\right\Vert _{\infty}\\
 & \le\gamma P+\underbrace{\sqrt{p}\left\Vert \sum_{b}2\X_{b}^{\top}\text{diag}(\e_{b})\w_{1,b}\right\Vert _{\infty}}_{S_{3}}+\underbrace{\sqrt{p}\left\Vert \sum_{b}2\X_{b}^{\top}\text{diag}(\e_{b})\w_{2,b}\right\Vert _{\infty}}_{S_{4}}.
\end{align*}
So the lemma is implied if we can show 
\[
S_{1}+S_{2}\le\frac{\lambda}{2}, \quad S_{3}+S_{4}\le\sigma\lambda, \quad \mbox{ w.h.p.}
\]
But $\lambda\gtrsim\sigma\left(\gamma+\sigma\right)\left(\sqrt{np}+\left|n_{1}-n_{2}\right|\sqrt{p}\right)\log^{3}n$
by assumption of Theorem \ref{thm:rand}. Therefore, the lemma follows
if each of the following bounds holds w.h.p.
\begin{align*}
S_{1} & \lesssim\sigma\gamma\sqrt{np}\log^{3}n,\\
S_{2} & \lesssim\sigma^{2}\sqrt{np}\log^{3}n,\\
S_{3} & \lesssim\sigma^{2}\gamma\left(\sqrt{np}+\left|n_{1}-n_{2}\right|\sqrt{p}\right)\log^{2}n,\\
S_{4} & \lesssim\sigma^{3}\sqrt{np}\log^{2}n.
\end{align*}
We now prove these bounds.

\paragraph{Term $S_{1}$:}

Note that $\gamma\ge\left\Vert \bbeta_{1}^{*}-\bbeta_{2}^{*}\right\Vert _{2}$,
so the desired bound on $S_{1}$ follows from the lemma below, which
is proved in the appendix.
\begin{lemma}
\label{lem:Gaussian_S1}Suppose $\bbeta_{1}^{*}-\bbeta_{2}^{*}$ is
supported on the first coordinate. Then w.h.p.
\[
\left\Vert S_{1}\right\Vert \lesssim\left\Vert \bbeta_{1}^{*}-\bbeta_{2}^{*}\right\Vert _{2}\sigma\sqrt{np}\log^{3}n.
\]

\end{lemma}

\paragraph{Term $S_{2}$:}

By definition, we have
\[
S_{2}=2\sum_{b}\left\Vert \sum_{i=1}^{n_{b}}\left(\e_{b,i}^{2}-\sigma^{2}\right)\x_{b,i}\x_{b,i}^{\top}\right\Vert .
\]
Here each $\e_{b,i}^{2}-\sigma^{2}$ is zero-mean, $\lesssim\sigma^{2}\log n$ almost surely,
and has variance $\lesssim\sigma^{4}$. The quantity inside the spectral norm is the sum of independent zero-mean bounded matrices. An application of the Matrix Bernstein inequality~\cite{tropp2010matrixmtg}
gives
\[
\left\Vert \sum_{i=1}^{n_{b}}\left(\e_{b,i}^{2}-\sigma^{2}\right)\x_{b,i}\x_{b,i}^{\top}\right\Vert \lesssim\sigma^{2}\sqrt{np}\log^{3}n,
\]
for each $b=1,2$. The desired bound follows.

\paragraph{Term $S_{3}$:}

We have
\begin{align*}
S_{3}= & \sqrt{p}\left\Vert \sum_{b}\X_{b}^{\top}\textrm{diag}\left(\e_{b}\right)\left(-\e_{b}\circ(\X_{b}\ddelta_{b}^{*})\right)\right\Vert _{\infty}\\
= & \sqrt{p}\left\Vert \sum_{b}\X_{b}^{\top}\textrm{diag}\left(\e_{b}^{2}\right)\X_{b}\ddelta_{b}^{*}\right\Vert _{\infty}\\
= & \sqrt{p}\max_{l\in[p]}\left|\sum_{b}\left(\e_{b}^{2}\circ\X_{b,l}\right)^{\top}\X_{b}\ddelta_{b}^{*}\right|,
\end{align*}
where $\X_{b,l}$ is the $l$-th column of $\X_{b}$. WLOG, we assume
$n_{1}\ge n_{2}$. Observe that for each $l\in[p]$, 
\[
\sum_{b}\left(\e_{b}^{2}\circ\X_{b,l}\right)^{\top}\X_{b}\ddelta_{b}^{*}=\underbrace{\sum_{i=1}^{n_{2}}\left(e_{1,i}^{2}\x_{1,i}(l)\x_{1,i}^{\top}-e_{2,i}^{2}\x_{2,i}(l)\x_{2,i}^{\top}\right)\ddelta_{1}^{*}}_{S_{3,1,l}}+\underbrace{\sum_{i=n_{2}+1}^{n_{1}}e_{1,i}^{2}\x_{1,i}(l)\x_{1,i}^{\top}\ddelta_{1}^{*}}_{S_{3,2,l}}.
\]
Let $\eeps_{i}$ be the $i$-th standard basis vector in $\mathbb{R}^{n}$.
The term $S_{3,1,l}$ can be written as
\begin{align*}
S_{3,1,l}= & \sum_{i=1}^{n_{2}}\left(\x_{1,i}^{\top}\left(e_{1,i}^{2}\eeps_{l}\ddelta_{1}^{*\top}\right)\x_{1,i}-\x_{2,i}^{\top}\left(e_{2,i}^{2}\eeps_{l}\ddelta_{1}^{*\top}\right)\x_{2,i}\right)\\
= & \cchi^{\top}\G\cchi,
\end{align*}
where
\begin{align*}
\cchi^{\top} & :=\left[\begin{array}{cccccccc}
e_{1,1}\x_{1,1}^{\top} & e_{1,2}\x_{1,2}^{\top} & \cdots & e_{1,n_{2}}\x_{1,n_{2}}^{\top} & e_{2,1}\x_{2,1}^{\top} & e_{2,2}\x_{2,2}^{\top} & \cdots & e_{2,n_{2}}\x_{2,n_{2}}^{\top}\end{array}\right]\in\mathbb{R}^{2n_{2}p}\\
\G & :=\textrm{diag}\left(\eeps_{l}\ddelta_{1}^{*\top},\eeps_{l}\ddelta_{1}^{*\top},\ldots,\eeps_{l}\ddelta_{1}^{*\top},-\eeps_{l}\ddelta_{1}^{*\top},-\eeps_{l}\ddelta_{1}^{*\top},\ldots,-\eeps_{l}\ddelta_{1}^{*\top}\right)\in\mathbb{R}^{2n_{2}p\times2n_{2}p};
\end{align*}
in other words, $\G$ is the block-diagonal matrix with $\left\{ \pm\eeps_{l}\ddelta_{1}^{*\top}\right\}$ on
its diagonal. Note that $\mathbb{E}S_{3,1,l}=0$, and the entries
of $\cchi$ are i.i.d.\ sub-Gaussian with parameter bounded by $\sigma\sqrt{\log n}$.
Using the Hanson-Wright inequality (e.g., \cite{rudelson2013hansonwright}), we obtain w.h.p.\
\[
\max_{l\in[p]}\left|S_{3,1,l}\right|\lesssim\left\Vert \G\right\Vert _{F}\sigma^{2}\log^{2}n\le\sigma^{2}\sqrt{2n}\gamma\log^{2}n.
\]
Since $\ddelta_{1}^{*}$ is supported on the first coordinate, the
term $S_{3,2,l}$ can be bounded w.h.p. by 
\[
\max_{l\in[p]}\left|S_{3,2,l}\right|=\max_{l\in[p]}\left|\sum_{i=n_{2}+1}^{n_{1}}e_{1,i}^{2}\x_{1,i}(l)\x_{1,i}(1)\ddelta_{1}^{*}(1)\right|\lesssim\left(n_{1}-n_{2}\right)\sigma^{2}\gamma\log^{2}n
\]
using the Hoeffding's inequality.
It follows that w.h.p.
\[
S_{3}\le\sqrt{p}\max_{l\in[p]}\left(\left|S_{3,1,l}\right|+\left|S_{3,2,l}\right|\right)\lesssim\sigma^{2}\gamma\left(\sqrt{np}+\left|n_{1}-n_{2}\right|\sqrt{p}\right)\log^{2}n.
\]

\paragraph{Term $S_{4}$:}

We have w.h.p.
\begin{align*}
S_{4}\le & 2\sqrt{p}\sum_{b}\left\Vert \X^{\top}\left(\e_{b}\circ\w_{2,b}\right)\right\Vert _{\infty}\\
\overset{(a)}{\lesssim} & \sqrt{p\log n}\sum_{b}\left\Vert \e_{b}\circ\w_{2,b}\right\Vert _{2}\\
= & \sqrt{p\log n}\sum_{b}\left\Vert \e_{b}^{3}-\sigma^{2}\e_{b}\right\Vert _{2}\\
\overset{(b)}{\lesssim} & \sigma^{3}\sqrt{np}\log^{2}n,
\end{align*}
where in (a) we use the independence between $\X$ and $\e_{b}\circ\w_{2,b}$
and the standard sub-Gaussian concentration inequality~(e.g., \cite{vershynin2010nonasym}), and (b) follows from the boundedness of
$\e$.

\section{Proof of Theorem~\ref{thm:lower_bound_determ}}

We need some additional notation. Let $\z:=\left(z_{1},z_{2},\ldots,z_{n}\right)^{\top}\in\left\{ 0,1\right\} ^{n}$
be the vector of hidden labels with $z_{i}=1$ if and only if $i\in\mathcal{I}_{1}$.
We use $\y(\ttheta^{*},\X,\e,\z)$ to denote the value of the response
vector $\y$ given $\ttheta^{*}$, $\X$, $\e$ and $\z$, i.e.,
\[
\y\left(\ttheta^{*},\X,\e,\z\right)=\z\circ\left(\X\bbeta_{1}^{*}\right)+(\mathbf{1}-\z)\circ\left(\X\bbeta_{2}^{*}\right)+\e,
\]
where $\mathbf{1}$ is the all-one vector in $\mathbb{R}^{n}$ and
$\circ$ denotes element-wise product.

By standard results, we know that with probability $1-n^{-10}$, 
\begin{equation}
\left\Vert \X\aalpha\right\Vert _{2}\le2\sqrt{n}\left\Vert \aalpha\right\Vert _{2},\forall\aalpha\in\mathbb{R}^{p}.\label{eq:rip}
\end{equation}
Hence it suffices to proves~(\ref{eq:determ_lower}) in the theorem
statement assuming~(\ref{eq:rip}) holds.

Let $\vv$ be an arbitrary unit vector in $\mathbb{R}^{p}$. We define
$\delta:=c_{0}\frac{\epsilon}{\sqrt{n}}$, $\ttheta_{1}:=\left(\frac{1}{2}\ugamma\vv,-\frac{1}{2}\ugamma\vv\right)$
and $\ttheta_{2}=\left(\frac{1}{2}\ugamma\vv+\delta\vv,-\frac{1}{2}\ugamma\vv-\delta\vv\right)$.
Note that $\ttheta_{1},\ttheta_{2}\in\Theta(\ugamma)$ as long as
$c_{0}$ is sufficiently small, and $\rho\left(\ttheta_{1},\ttheta_{2}\right)=2\delta$.
We further define $\e_{1}:=\zero$ and $\e_{2}:=-\delta\left(2\z-\mathbf{1}\right)\circ(\X\vv)$.
Note that $\left\Vert \e_{2}\right\Vert \le2\sqrt{n}\delta\le\epsilon$
by~(\ref{eq:rip}), so $\e_{1},\e_{2}\in\mathbb{B}(\epsilon)$. If
we set $\y_{i}=\y\left(\ttheta_{i},\X,\e_{i},\z\right)$ for $i=1,2$,
then we have 
\begin{align*}
\y_{2} & =\z\circ\left(\X(\frac{1}{2}\ugamma\vv+\delta\vv)\right)+(\mathbf{1}-\z)\circ\left(\X(-\frac{1}{2}\ugamma\vv-\delta\vv)\right)+\e_{2}\\
 & =(2\z-\mathbf{1})\circ\left(\X(\frac{1}{2}\ugamma\vv+\delta\vv)\right)-\delta\left(2\z-\mathbf{1}\right)\circ(\X\vv)\\
 & =(2\z-\mathbf{1})\circ\left(\X(\frac{1}{2}\ugamma\vv)\right)+\e_{1}\\
 & =\y_{1},
\end{align*}
which holds for any $\X$ and $\z$. $ $Therefore, for any $\hat{\ttheta}$,
we have
\begin{align*}
\sup_{\ttheta^{*}\in\Theta(\ugamma)}\sup_{\e\in\mathbb{B}(\epsilon)}\rho\left(\hat{\ttheta}(\X,\y),\ttheta^{*}\right) & \ge\frac{1}{2}\rho\left(\hat{\ttheta}\left(\X,\y_{1}\right),\ttheta_{1}\right)+\frac{1}{2}\rho\left(\hat{\ttheta}\left(\X,\y_{2}\right),\ttheta_{2}\right)\\
 & =\frac{1}{2}\rho\left(\hat{\ttheta}\left(\X,\y_{1}\right),\ttheta_{1}\right)+\frac{1}{2}\rho\left(\hat{\ttheta}\left(\X,\y_{1}\right),\ttheta_{2}\right)\\
 & \ge\frac{1}{2}\rho\left(\ttheta_{1},\ttheta_{2}\right)\\
 & =\delta,
\end{align*}
where the second inequality holds because $\rho$ is a metric and
satisfies the triangle inequality. Taking the infimum over $\hat{\ttheta}$
proves the theorem.

\section{Proof of Theorem~\ref{thm:lower_bound_stochastic}}

Throughout the proof we set $\kappa:=\frac{1}{2}\ugamma$.

\subsection{Part 1 of the Theorem}

We prove the first part of the theorem by establishing a lower-bound
for standard linear regression. Set $\delta_{1}:=c_{0}\sigma\sqrt{\frac{p-1}{n}}$,
and define the (semi)-metric $\rho_{1}\left(\cdot,\cdot\right)$ by
$\rho_{1}(\bbeta,\bbeta')=\min\left\{ \left\Vert \bbeta-\bbeta'\right\Vert ,\left\Vert \bbeta+\bbeta'\right\Vert \right\} $.
We begin by constructing a $\delta_{1}-$packing set $\Phi_{1}:=\left\{ \bbeta_{1},\ldots,\bbeta_{M}\right\} $
of $\mathbb{G}^{p}\left(\kappa\right):=\left\{ \bbeta\in\mathbb{R}^{p}:\left\Vert \bbeta\right\Vert \ge\kappa\right\} $
in the metric $\rho_{1}$. We need a packing set of the hypercube
$\{0,1\}^{p-1}$ in the Hamming distance. 
\begin{lemma}
\label{lem:packing}For $p\ge16$, there exists $\left\{ \xxi_{1},\ldots,\xxi_{M}\right\} \subset\left\{ 0,1\right\} ^{p-1}$
such that 
\begin{align*}
M & \ge2^{(p-1)/16},\\
\min\left\{ \left\Vert \xxi_{i}-\xxi_{j}\right\Vert _{0},\left\Vert \xxi_{i}+\xxi_{j}\right\Vert _{0}\right\}  & \ge\frac{p-1}{16},\forall1\le i<j\le M.
\end{align*}
\end{lemma}
Let $\tau:=2c_{0}\sigma\sqrt{\frac{1}{n}}$ for some absolute
constant $c_{0}>0$ that is sufficiently small, and $\kappa_{0}^{2}:=\kappa^{2}-(p-1)\tau^{2}.$
Note that $\kappa_{0}\ge0$ since $\ugamma\ge\sigma$ by assumption.
For $i=1,\ldots,M$, we set 
\[
\bbeta_{i}=\kappa_{0}\eeps_{p}+\sum_{j=1}^{p-1}\left(2\xxi_{i}(j)-1\right)\tau\eeps_{j},
\]
where $\eeps_{j}$ is the $j$-th standard basis in $\mathbb{R}^{p}$
and $\xxi_{i}(j)$ is the $j$-th coordinate of $\xxi_{i}$.
Note that $\left\Vert \bbeta_{i}\right\Vert _{2}=\kappa,\forall i\in[M]$,
so $\Phi_{1}=\left\{ \bbeta_{1},\ldots,\bbeta_{M}\right\} \subset\mathbb{G}^{p}(\kappa)$.
We also have that for all $1\le i<j\le M$, 
\begin{equation}
\left\Vert \bbeta_{i}-\bbeta_{j}\right\Vert _{2}^{2}\le(p-1)\tau^{2}=4c_{0}^{2}\frac{\sigma^{2}(p-1)}{n}.\label{eq:distance_standard}
\end{equation}
Moreover, we have 
\begin{align}
\rho^{2}\left(\bbeta_{i},\bbeta_{j}\right) & =\min\left\{ \left\Vert \bbeta_{i}-\bbeta_{j}\right\Vert _{2}^{2},\left\Vert \bbeta_{i}+\bbeta_{j}\right\Vert _{2}^{2}\right\} \nonumber \\
 & \ge4\tau^{2}\min\left\{ \left\Vert \xxi_{i}-\xxi_{j}\right\Vert _{0},\left\Vert \xxi_{i}+\xxi_{j}\right\Vert _{0}\right\} \ge4\cdot4c_{0}^{2}\frac{\sigma^{2}}{n}\cdot\frac{p-1}{16}=\delta_{1}^{2}.\label{eq:packing_standard}
\end{align}
so $\Phi_{1}=\left\{ \bbeta_{1},\ldots,\bbeta_{M}\right\} $ is a
$\delta_{1}$-packing of $\mathbb{G}^{p}(\kappa)$ in the metric
$\rho_{1}$.

Suppose $\bbeta^{*}$ is sampled uniformly at random from the set
$\Phi_{1}$. For $i=1,\ldots,M$, let $\mathbb{P}_{i,\X}$ denote
the distribution of $\y$ conditioned on $\bbeta^{*}=\bbeta_{i}$
and $\X$, and $\mathbb{P}_{i}$ denote the joint distribution of
$\X$ and $\y$ conditioned on $\bbeta^{*}=\bbeta_{i}$. Because $\X$
are independent of $\z$,$\e$ and $\bbeta^{*}$, we have 
\begin{align*}
D\left(\mathbb{P}_{i}\Vert\mathbb{P}_{i'}\right) & =\mathbb{E}_{\mathbb{P}_{i}(\X,\y)}\log\frac{p_{i}(\X,\y)}{p_{i'}(\X,\y)}\\
 & =\mathbb{E}_{\mathbb{P}_{i}(\X,\y)}\log\frac{p_{i}(\y\vert\X)}{p_{i'}(\y\vert\X)}\\
 & =\mathbb{E}_{\mathbb{P}(\X)}\left[\mathbb{E}_{\mathbb{P}_{i}(\y\vert\X)}\left[\log\frac{p_{i}(\y\vert\X)}{p_{i'}(\y\vert\X)}\right]\right]\\
 & =\mathbb{E}_{\X}\left[D\left(\mathbb{P}_{i,\X}\Vert\mathbb{P}_{i',\X}\right)\right].
\end{align*}
Using the above equality and the convexity of the mutual information,
we get that
\begin{align*}
I\left(\bbeta^{*};\X,\y\right)\le\frac{1}{M^{2}}\sum_{1\le i,i'\le M}D\left(\mathbb{P}_{i}\Vert\mathbb{P}_{i'}\right)= & \frac{1}{M^{2}}\sum_{1\le i,i'\le M}\mathbb{E}_{\X}\left[D\left(\mathbb{P}_{i,\X}\Vert\mathbb{P}_{i',\X}\right)\right]\\
= & \frac{1}{M^{2}}\sum_{1\le i,i'\le M}\mathbb{E}_{\X}\frac{\left\Vert \X\bbeta_{i}-\X\bbeta_{i'}\right\Vert ^{2}}{2\sigma^{2}}\\
= & \frac{1}{M^{2}}\sum_{1\le i,i'\le M}\frac{n\left\Vert \bbeta_{i}-\bbeta_{i'}\right\Vert ^{2}}{2\sigma^{2}}.
\end{align*}
It follows from~(\ref{eq:distance_standard}) that
\[
I\left(\bbeta^{*};\X,\y\right)\le8c_{0}^{2}p\le\frac{1}{2}\left(\log_{2}M\right)/\left(\log_{2}e\right)=\frac{1}{4}\log M
\]
provided $c_{0}$ is sufficiently small. Following a standard argument~\cite{yu1997assouad,yang1999minimax,birge1983approximation}
to transform the estimation problem into a hypothesis testing problem (cf. Eq.~\eqref{eq:testing} and~\eqref{eq:fano}),
we obtain
\begin{align*}
\inf_{\hat{\bbeta}}\sup_{\bbeta^{*}\in\mathbb{G}^{p}(\kappa)}\mathbb{E}_{\X,\z,\e}\left[\rho_{1}\left(\hat{\bbeta},\bbeta^{*}\right)\right] & \ge\delta_{1}\left(1-\frac{I\left(\bbeta^{*};\X,\y\right)+\log2}{\log M}\right)\\
 & \ge\frac{1}{2}\delta_{1}=\frac{1}{2}c_{0}\sigma\sqrt{\frac{p}{n}}.
\end{align*}
This establishes a minimax lower bound for standard linear regression.
Now observe that given any standard linear regression problem with
regressor $\bbeta^{*}\in\mathbb{G}^{p}\left(\kappa\right)$, we can
reduce it to a mixed regression problem with $\ttheta^{*}=\left(\bbeta^{*},-\bbeta^{*}\right)\in\Theta(\ugamma)$
by multiplying each $y_{i}$ by a Rademacher $\pm1$ variable. Part
1 of the theorem hence follows.

\subsection{Part 2 of the Theorem}

Let $\delta_{2}:=2c_{0}\frac{\sigma^{2}}{\kappa}\sqrt{\frac{p-1}{n}}$.
We first construct a $\delta_{2}-$packing set $\Theta_{2}:=\left\{ \ttheta_{1},\ldots,\ttheta_{M}\right\} $
of $\Theta(\ugamma)$ in the metric $\rho(\cdot,\cdot)$. Set $\tau:=2c_{0}\frac{\sigma^{2}}{\kappa}\sqrt{\frac{1}{n}}$
and $\kappa_{0}^{2}:=\kappa^{2}-(p-1)\tau^{2}$. Note that
$\kappa_{0}\ge0$ under the assumption $\kappa\ge c_{1}\sigma\left(\frac{p}{n}\right)^{1/4}$
provided that $c_{0}$ is small enough. For $i=1,\ldots,M$, we set
$\ttheta_{i}:=\left(\bbeta_{i},-\bbeta_{i}\right)$ with 
\[
\bbeta_{i}=\kappa_{0}\eeps_{p}+\sum_{j=1}^{p-1}\left(2\xxi_{i}(j)-1\right)\tau\eeps_{j},
\]
where $\left\{ \xxi_{i}\right\} $ are the vectors in Lemma~\ref{lem:packing}.
Note that $\left\Vert \bbeta_{i}\right\Vert =\kappa$ for all $i$,
so $\Theta_{2}=\left\{ \ttheta_{1},\ttheta_{2},\ldots,\ttheta_{M}\right\} \subset\Theta(\ugamma)$.
We also have that for all $1\le i<i'\le M$,
\begin{equation}
\left\Vert \bbeta_{i}-\bbeta_{i'}\right\Vert ^{2}\le p\tau^{2}=4c_{0}^{2}\frac{\sigma^{4}p}{\kappa^{2}n}.\label{eq:distance_3}
\end{equation}
Moreover, we have 
\begin{align}
\rho^{2}\left(\ttheta_{i},\ttheta_{i'}\right) & =4\min\left\{ \left\Vert \bbeta_{i}-\bbeta_{i'}\right\Vert ^{2},\left\Vert \bbeta_{i}+\bbeta_{i'}\right\Vert ^{2}\right\} \nonumber \\
 & \ge16\tau^{2}\min\left\{ \left\Vert \xxi_{i}-\xxi_{i'}\right\Vert _{0},\left\Vert \xxi_{i}+\xxi_{i'}\right\Vert _{0}\right\} \ge16\cdot4c_{0}^{2}\frac{\sigma^{4}}{\kappa^{2}n}\cdot\frac{p-1}{16}=\delta_{2}^{2},\label{eq:packing_2}
\end{align}
so $\Theta_{2}=\left\{ \ttheta_{1},\ldots,\ttheta_{M}\right\} $ forms
a $\delta_{2}$-packing of the $\Theta(\ugamma)$ in the metric $\rho$.

Suppose $\ttheta^{*}$ is sampled uniformly at random from the set
$\Theta_{2}$. For $i=1,\ldots,M$, let $\mathbb{P}_{i,\X}^{(j)}$
denote the distribution of $\y_{j}$ conditioned on $\ttheta^{*}=\ttheta_{i}$
and $\X$, $\mathbb{P}_{i,\X}$ denote the distribution of $\y$ conditioned
on $\ttheta^{*}=\ttheta_{i}$ and $\X$, and $\mathbb{P}_{i}$ denote
the joint distribution of $\X$ and $\y$ conditioned on $\ttheta^{*}=\ttheta_{i}$.
We need the following bound on the KL divergence between two mixtures
of univariate Gaussians. For any $a>0$, we use $\mathbb{Q}_{a}$
to denote the distribution of the equal-weighted mixture of two Gaussian
distributions $\mathcal{N}(a,\sigma^{2})$ and $\mathcal{N}(-a,\sigma^{2})$.
\begin{lemma}
\label{lem:KL_bound} The following bounds holds for any $u,v\ge0$:
\[
D\left(\mathbb{Q}_{u}\Vert\mathbb{Q}_{v}\right)\le\frac{u^{2}-v^{2}}{2\sigma^{4}}u^{2}+\frac{v^{3}\max\left\{ 0,v-u\right\} }{2\sigma^{8}}\left(u^{4}+6u^{2}\sigma^{2}+3\sigma^{4}\right).
\]

\end{lemma}
Note that $\mathbb{P}_{i,\X}^{(j)}=\mathbb{Q}_{\left|\x_{j}^{\top}\bbeta_{i}\right|}$. Using $\mathbb{P}_{i,\X}=\otimes_{j=1}^{n}\mathbb{P}_{i,\X}^{(j)}$
and the above lemma, we have
\begin{align*}
 & \mathbb{E}_{\X}D\left(\mathbb{P}_{i,\X}\Vert\mathbb{P}_{i',\X}\right)\\
= & \sum_{j=1}^{n}\mathbb{E}_{\X}D\left(\mathbb{P}_{i,\X}^{(j)}\Vert\mathbb{P}_{i',\X}^{(j)}\right)\\
\le & n\mathbb{E}\frac{\left|\x_{1}^{\top}\bbeta_{i}\right|^{2}-\left|\x_{1}^{\top}\bbeta_{i'}\right|^{2}}{2\sigma^{4}}\left|\x_{j}^{\top}\bbeta_{i}\right|^{2}\\
&+n\mathbb{E}_{\X}\frac{\left|\x_{1}^{\top}\bbeta_{i'}\right|^{3}\max\left\{ 0, \left|\x_{1}^{\top}\bbeta_{i'}\right|-\left|\x_{1}^{\top}\bbeta_{i}\right|\right\} }{2\sigma^{8}}\left(\left|\x_{1}^{\top}\bbeta_{i}\right|^{4}+6\left|\x_{1}^{\top}\bbeta_{i}\right|^{2}\sigma^{2}+3\sigma^{4}\right).
\end{align*}
To bound the expectations in the last RHS, we need a simple technical
lemma.
\begin{lemma}
\label{lem:moment_bound}Suppose $\x\in\mathbb{R}^{p}$ has i.i.d.
standard Gaussian components, and $\aalpha,\bbeta\in\mathbb{R}^{p}$
are any fixed vectors with $\left\Vert \aalpha\right\Vert _{2}=\left\Vert \bbeta\right\Vert _{2}$.
There exists an absolute constant $\bar{c}$ such that for any non-negative
integers $k,l$ with $k+l\le8$,
\begin{align*}
\mathbb{E}\left|\x^{\top}\aalpha\right|^{k}\left|\x^{\top}\bbeta\right|^{l} & \le\bar{c}\left\Vert \aalpha\right\Vert ^{k}\left\Vert \bbeta\right\Vert ^{l}.
\end{align*}
Moreover, we have
\[
\mathbb{E}_{\X}\left[\left(\left|\x^{\top}\aalpha\right|^{2}-\left|\x^{\top}\bbeta\right|^{2}\right)\left|\x^{\top}\aalpha\right|^{2}\right]\le2\left\Vert \aalpha\right\Vert \left\Vert \aalpha-\bbeta\right\Vert ^{2}.
\]
\[
\mathbb{E}\left(\left|\x^{\top}\aalpha\right|^{2}-\left|\x^{\top}\bbeta\right|^{2}\right)^{2}\le\left\Vert \aalpha-\bbeta\right\Vert ^{4}.
\]
\end{lemma}
Using the above lemma and the fact that $\left\Vert \bbeta_{i}\right\Vert _{2}=\left\Vert \bbeta_{i'}\right\Vert _{2}=\kappa$
for all $1\le i<i'\le M$, we have
\[
\mathbb{E}_{\X}\frac{\left|\x_{1}^{\top}\bbeta_{i}\right|^{2}-\left|\x_{1}^{\top}\bbeta_{i'}\right|^{2}}{2\sigma^{4}}\left|\x_{1}^{\top}\bbeta_{i}\right|^{2}\le\frac{1}{2\sigma^{4}}\kappa^{2}\left\Vert \bbeta_{i}-\bbeta_{i'}\right\Vert ^{2}
\]
and for some universal constant $c'>0$, 
\begin{align*}
 & \mathbb{E}_{\X}\frac{\left|\x_{1}^{\top}\bbeta_{i'}\right|^{3}\max\left\{ 0,\left|\x_{1}^{\top}\bbeta_{i'}\right|-\left|\x_{1}^{\top}\bbeta_{i}\right|\right\} }{2\sigma^{8}}\left(\left|\x_{1}^{\top}\bbeta_{i}\right|^{4}+6\left|\x_{1}^{\top}\bbeta_{i}\right|^{2}\sigma^{2}+3\sigma^{4}\right)\\
\le & \frac{1}{2\sigma^{8}}\mathbb{E}_{\X}\max\left\{ 0,\left|\x_{1}^{\top}\bbeta_{i'}\right|^{2}-\left|\x_{1}^{\top}\bbeta_{i}\right|^{2}\right\} \left|\x_{1}^{\top}\bbeta_{i'}\right|^{2}\left(\left|\x_{1}^{\top}\bbeta_{i}\right|^{4}+6\left|\x_{1}^{\top}\bbeta_{i}\right|^{2}\sigma^{2}+3\sigma^{4}\right)\\
\overset{(a)}{\le} & \frac{1}{2\sigma^{4}}\sqrt{\mathbb{E}_{\X}\left(\left|\x_{1}^{\top}\bbeta_{i'}\right|^{2}-\left|\x_{1}^{\top}\bbeta_{i}\right|^{2}\right)^{2}\cdot\frac{1}{\sigma^{8}}\mathbb{E}_{\X}\left|\x_{1}^{\top}\bbeta_{i'}\right|^{4}\left(\left|\x_{1}^{\top}\bbeta_{i}\right|^{4}+6\left|\x_{1}^{\top}\bbeta_{i}\right|^{2}\sigma^{2}+3\sigma^{4}\right)^{2}}\\
\overset{(b)}{\le} & \frac{1}{2\sigma^{4}}\sqrt{\left\Vert \bbeta_{i}-\bbeta_{i}\right\Vert ^{4}\cdot c'^{2}\left\Vert \bbeta_{i'}\right\Vert ^{4}}=\frac{c'}{2\sigma^{4}}\left\Vert \bbeta_{i}-\bbeta_{i}\right\Vert ^{2}\kappa^{2},
\end{align*}
where (a) follows from Cauchy-Schwarz inequality, and (b) follows
from the first and third inequalities in Lemma~\ref{lem:moment_bound}
as well as $\left\Vert \bbeta_{i}\right\Vert =\left\Vert \bbeta_{i'}\right\Vert =\kappa\le\sigma$.
It follows that

\[
\mathbb{E}_{\X}D\left(\mathbb{P}_{i,\X}\Vert\mathbb{P}_{i',\X}\right)\le n\cdot\frac{c'\left\Vert \bbeta_{i}-\bbeta_{i'}\right\Vert ^{2}\kappa^{2}}{\sigma^{4}}\le c''p,
\]
where the last inequality follows from~(\ref{eq:distance_3}) and
$c''$ can be made sufficiently small by choosing $c_{0}$ small enough.
We therefore obtain
\begin{align*}
 & I\left(\ttheta^{*};\X,\y\right)\\
\le & \frac{1}{M^{2}}\sum_{1\le i,i'\le M}D\left(\mathbb{P}_{i}\Vert\mathbb{P}_{i'}\right)\\
= & \frac{1}{M}\sum_{1\le i,i'\le M}\mathbb{E}_{\X}\left[D\left(\mathbb{P}_{i,\X}\Vert\mathbb{P}_{i',\X}\right)\right]\\
\le & c''p\le\frac{1}{4}\log M
\end{align*}
using $M\ge2^{(p-1)/16}$. Following a standard argument~\cite{yu1997assouad,yang1999minimax,birge1983approximation}
to transform the estimation problem into a hypothesis testing problem (cf. Eq.~\eqref{eq:testing} and~\eqref{eq:fano}),
we obtain
\begin{align*}
\inf_{\hat{\ttheta}}\sup_{\ttheta^{*}\in \Theta(\ugamma)}\mathbb{E}_{\X,\z,\e}\left[\rho\left(\hat{\ttheta},\ttheta^{*}\right)\right] & \ge\delta_{2}\left(1-\frac{I\left(\ttheta^{*};\X,\y\right)+\log2}{\log M}\right)\\
 & \ge\frac{1}{2}\delta_{2}=c_{0}\frac{\sigma^{2}}{\kappa}\sqrt{\frac{p}{n}}.
\end{align*}

\subsection{Part 3 of the Theorem}

The proof follows similar lines as Part 2. Let $\delta_{3}:=2c_{0}\sigma\left(\frac{p}{n}\right)^{1/4}$.
Again we first construct a $\delta_{3}-$packing set $\Theta_{3}:=(\ttheta_{1},\ldots,\ttheta_{M})$
of $\Theta(\ugamma)$ in the metric $\rho(\cdot,\cdot)$. Set $\tau:=\frac{2c_{0}\sigma}{\sqrt{p-1}}\left(\frac{p}{n}\right)^{1/4}$.
For $i=1,\ldots,M$, we set $\ttheta_{i}=(\bbeta_{i},-\bbeta_{i})$
with 
\[
\bbeta_{i}=\sum_{j=1}^{p-1}\left(2\xxi_{i}(j)-1\right)\tau\eeps_{j},
\]
where $\left\{ \xxi_{i}\right\} $ are the vectors from Lemma~\ref{lem:packing}.
Note that $\left\Vert \bbeta_{i}\right\Vert _{2}=\sqrt{p-1}\tau=2c_{0}\sigma\left(\frac{p}{n}\right)^{1/4}\ge c_{1}\sigma\left(\frac{p}{n}\right)^{1/4}\ge\kappa$
provided $c_{1}$ is sufficiently small, so $\Theta_{3}=\left\{ \ttheta_{1},\ldots,\ttheta_{M}\right\} \subset\Theta(\ugamma).$
We also have for all $1\le i<i'\le M$, 
\begin{align}
\rho^{2}\left(\bbeta_{i},\bbeta_{i'}\right) & =4\min\left\{ \left\Vert \bbeta_{i}-\bbeta_{i'}\right\Vert _{2}^{2}\left\Vert \bbeta_{i}+\bbeta_{i'}\right\Vert _{2}^{2}\right\} \nonumber \\
 & \ge16\tau^{2}\min\left\{ \left\Vert \xxi_{i}-\xxi_{i'}\right\Vert _{0},\left\Vert \xxi_{i}+\xxi_{i'}\right\Vert _{0}\right\} =16\cdot\frac{4c_{0}^{2}\sigma^{2}}{p-1}\sqrt{\frac{p}{n}}\cdot\frac{p-1}{16}\ge\delta_{3}^{2},\label{eq:packing_3}
\end{align}
so $\Theta_{3}=\left\{ \ttheta_{1},\ldots,\ttheta_{M}\right\} $ is
a $\delta_{3}$-packing of $\Theta(\ugamma)$ in the metric $\rho$.

Suppose $\ttheta^{*}$ is sampled uniformly at random from the set
$\Theta_{2}$. Define $\mathbb{P}_{i,\X},\mathbb{P}_{i,\X}^{(j)}$
and $\mathbb{P}_{i}$ as in the proof of Part 2 of the theorem. We
have
\begin{align*}
 & \mathbb{E}_{\X}D\left(\mathbb{P}_{i,\X}\vert\mathbb{P}_{i',\X}\right)\\
= & \sum_{j=1}^{n}\mathbb{E}_{\X}D\left(\mathbb{P}_{i,\X}^{(j)}\Vert\mathbb{P}_{i',\X}^{(j)}\right)\\
\overset{(a)}{\le} & n\mathbb{E}_{\X}\frac{\left|\x_{1}^{\top}\bbeta_{i}\right|^{2}-\left|\x_{1}^{\top}\bbeta_{i'}\right|^{2}}{2\sigma^{4}}\left|\x_{1}^{\top}\bbeta_{i}\right|^{2}\\
&+n\mathbb{E}_{\X}\frac{\left|\x_{1}^{\top}\bbeta_{i'}\right|^{3}\max\left\{ 0,\left|\x_{1}^{\top}\bbeta_{i'}\right|-\left|\x_{1}^{\top}\bbeta_{i}\right|\right\} }{2\sigma^{8}}\left(\left|\x_{1}^{\top}\bbeta_{i}\right|^{4}+6\left|\x_{1}^{\top}\bbeta_{i}\right|^{2}\sigma^{2}+3\sigma^{4}\right)\\
\le & \frac{n}{2\sigma^{4}}\mathbb{E}_{\X}\left|\x_{1}^{\top}\bbeta_{i}\right|^{4}+\frac{n}{2\sigma^{8}}\mathbb{E}_{\X}\left|\x_{1}^{\top}\bbeta_{i'}\right|^{4}\left(\left|\x_{1}^{\top}\bbeta_{i}\right|^{4}+6\left|\x_{1}^{\top}\bbeta_{i}\right|^{2}\sigma^{2}+3\sigma^{4}\right)\\
\overset{(b)}{\le} & \frac{n}{2\sigma^{4}}\bar{c}\left\Vert \bbeta_{i}\right\Vert ^{4}+\frac{n}{2\sigma^{8}}\bar{c}\left\Vert \bbeta_{i'}\right\Vert ^{4}\left(\left\Vert \bbeta_{i}\right\Vert ^{4}+6\sigma^{2}\left\Vert \bbeta_{i}\right\Vert ^{2}+9\sigma^{4}\right)\\
\overset{(c)}{\le} & c'p.
\end{align*}
where (a) follows from Lemma~\ref{lem:KL_bound}, (b) follows from Lemma~\ref{lem:moment_bound},
(c) follows from $ $$\left\Vert \bbeta_{i}\right\Vert =2c_{0}\sigma\left(\frac{p}{n}\right)^{1/4}\le\sigma,\forall i$,
and $c'$ is a sufficiently small absolute constant. It follows that
\[
I\left(\ttheta^{*};\X,\y\right)\le\frac{1}{M}\sum_{1\le i,i'\le M}\mathbb{E}_{\X}D\left(\mathbb{P}_{i}\Vert\mathbb{P}_{i'}\right)\le c'p\le\frac{1}{4}\log M
\]
since $M\ge2^{(p-1)/8}$. Following a standard argument~\cite{yu1997assouad,yang1999minimax,birge1983approximation}
to transform the estimation problem into a hypothesis testing problem (cf. Eq.~\eqref{eq:testing} and~\eqref{eq:fano}),
we obtain
\begin{align*}
\inf_{\hat{\ttheta}}\sup_{\ttheta^{*}\in \Theta(\ugamma)}\mathbb{E}_{\X,\z,\e}\left[\rho\left(\hat{\ttheta},\ttheta^{*}\right)\right] & \ge\delta_{3}\left(1-\frac{I\left(\ttheta^{*};\X,\y\right)+\log2}{\log M}\right)\\
 & \ge\frac{1}{2}\delta_{3}=c_{0}\sigma\left(\frac{p}{n}\right)^{1/4}.
\end{align*}

\section{Proofs of Technical Lemmas}

\subsection{Proof of Lemma \ref{claim} }

Simple algebra shows that
\begin{align*}
\sum_{b}\left\Vert \hat{\H}_{T}-2\bbeta_{b}^{*}\hat{\h}^{\top}\right\Vert _{F}^{2} & =2\left\Vert \hat{\H}_{T}-(\bbeta_{1}^{*}+\bbeta_{2}^{*})\hat{\h}^{\top}\right\Vert _{F}^{2}+2\left\Vert \bbeta_{1}^{*}-\bbeta_{2}^{*}\right\Vert _{2}^{2}\|\hat{\h}\| _{2}^{2}\\
 & \ge2\left\Vert \bbeta_{1}^{*}-\bbeta_{2}^{*}\right\Vert _{2}^{2}\|\hat{\h}\| _{2}^{2}\ge\alpha\left(\left\Vert \bbeta_{1}^{*}\right\Vert _{2}+\left\Vert \bbeta_{2}^{*}\right\Vert \right)^{2}\|\hat{\h}\| _{2}^{2},
\end{align*}
and
\begin{align*}
 & \sum_{b}\left\Vert \hat{\H}_{T}-2\bbeta_{b}^{*}\hat{\h}\right\Vert _{F}^{2}\\
= & 4\left(\left\Vert \bbeta_{1}^{*}\right\Vert _{2}^{2}+\left\Vert \bbeta_{2}^{*}\right\Vert ^{2}\right)\left\Vert \hat{\h}-\frac{\hat{\H}_{T}(\bbeta_{1}^{*}+\bbeta_{2}^{*})}{2\left\Vert \bbeta_{1}^{*}\right\Vert _{2}^{2}+2\left\Vert \bbeta_{2}^{*}\right\Vert ^{2}}\right\Vert _{2}^{2}\\
&+\frac{2\left(\left\Vert \bbeta_{1}^{*}\right\Vert _{2}^{2}+\left\Vert \bbeta_{2}^{*}\right\Vert ^{2}\right)\left\Vert \hat{\H}_{T}\right\Vert _{F}^{2}-\left\Vert \hat{\H}_{T}\left(\bbeta_{1}^{*}+\bbeta_{2}^{*}\right)\right\Vert _{2}^{2}}{\left\Vert \bbeta_{1}^{*}\right\Vert _{2}^{2}+\left\Vert \bbeta_{2}^{*}\right\Vert ^{2}}\\
\overset{(a)}{\ge} & \frac{2\left(\left\Vert \bbeta_{1}^{*}\right\Vert _{2}^{2}+\left\Vert \bbeta_{2}^{*}\right\Vert ^{2}\right)\left\Vert \hat{\H}_{T}\right\Vert _{F}^{2}-\left\Vert \hat{\H}_{T}\right\Vert _{F}^{2}\left\Vert \bbeta_{1}^{*}+\bbeta_{2}^{*}\right\Vert _{2}^{2}}{\left\Vert \bbeta_{1}^{*}\right\Vert _{2}^{2}+\left\Vert \bbeta_{2}^{*}\right\Vert ^{2}}=\alpha\left\Vert \hat{\H}_{T}\right\Vert _{F}^{2},
\end{align*}
where the inequality (a) follows from $\left\Vert \hat{\H}_{T}\right\Vert \le\left\Vert \hat{\H}_{T}\right\Vert _{F}$.
Combining the last two display equations with the simple inequality
\[
\sum_{b}\left\Vert \hat{\H}_{T}-2\bbeta_{b}^{*}\hat{\h}\right\Vert _{F}\ge\sqrt{\sum_{b}\left\Vert \hat{\H}_{T}-2\bbeta_{b}^{*}\hat{\h}\right\Vert _{F}^{2}},
\]
we obtain
\begin{align*}
\sum_{b}\left\Vert \hat{\H}_{T}-2\bbeta_{b}^{*}\hat{\h}\right\Vert _{F} & \ge\sqrt{\alpha}\left(\left\Vert \bbeta_{1}^{*}\right\Vert _{2}+\left\Vert \bbeta_{2}^{*}\right\Vert _{2}\right)\|\hat{\h}\| _{2},\\
\sum_{b}\left\Vert \hat{\H}_{T}-2\bbeta_{b}^{*}\hat{\h}\right\Vert _{F} & \ge\sqrt{\alpha}\left\Vert \hat{\H}_{T}\right\Vert _{F}.
\end{align*}

\subsection{Proof of Lemmas \ref{lem:RIP} and \ref{lem:RIP2}}

Setting $\sigma=0$ in Lemma \ref{lem:RIP2} recovers Lemma \ref{lem:RIP}.
So we only need to prove Lemma \ref{lem:RIP2}.
The proofs for $b=1$ and $2$ are identical, so we omit the subscript
$b$. WLOG we may assume $\sigma=1$. Our proof generalizes the proof of an RIP-type result in~\cite{chen2013covariance}

Fix $\Z$ and $\z$. Let $\xi_{j}:=\left\langle \B_{j},\Z\right\rangle $ and $\nu:=\left\Vert \Z\right\Vert _{F}$. We already know that $\xi_{j}$ is a sub-exponential random variable with $\left\Vert \xi_{j}\right\Vert _{\psi_{1}}\le c_{1}\nu$ and $\left\Vert \xi_{j}-\mathbb{E}\left[\xi_{j}\right]\right\Vert _{\psi_{1}}\le2c_{1}\nu.$ 

On the other hand, let $\gamma_{j}=\left\langle \dd_{j},\z\right\rangle $
and $\omega:=\left\Vert \z\right\Vert _{2}$. It is easy to check
that $\gamma_{j}$ is sub-Gaussian with $\left\Vert \gamma_{j}\right\Vert _{\psi_{2}}\le c_{1}\mu$. It follows that $\left\Vert \xi_{j}-\gamma_{j}\right\Vert _{\psi_{1}}\le c_{1}\left(\nu+\omega\right).$

Note that 
\[
\left\Vert \BB\Z-\D\z\right\Vert _{1}=\sum_{j=1}^{n/2}\frac{2}{n}\left|\xi_{j}-\gamma_{j}\right|.
\]
Therefore, applying the Bernstein-type inequality for the sum of sub-exponential
variables~\cite{vershynin2010nonasym}, we obtain
\[
\mathbb{P}\left[\left|\left\Vert \BB\Z-\D\z\right\Vert _{1}-\mathbb{E}\left|\xi_{j}-\gamma_{j}\right|\right|\ge t\right]\le2\exp\left[-c\min\left\{ \frac{t^{2}}{c_{2}(\nu+\mu)^{2}/n},\frac{t}{c_{2}(\nu+\mu)/n}\right\} \right].
\]
Setting $t=(\nu+\sigma\omega)/c_{3}$ for any $c_{3}>1$, we get
\begin{equation}
\mathbb{P}\left[\left|\left\Vert \BB\Z-\D\z\right\Vert _{1}-\mathbb{E}\left|\xi_{j}-\gamma_{j}\right|\right|\ge\frac{\nu+\omega}{c_{3}}\right]\le2\exp\left[-c_{4}n\right].\label{eq:concentrate2}
\end{equation}

But sub-exponentiality implies
\[
\mathbb{E}\left[\left|\xi_{j}-\gamma_{j}\right|\right]\le\left\Vert \xi_{j}-\gamma_{j}\right\Vert _{\psi_{1}}\le c_{2}\left(\nu+\mu\right).
\]
Hence
\[
\mathbb{P}\left[\left\Vert \BB\Z-\D\z\right\Vert _{1}\ge\left(c_{2}+\frac{1}{c_{3}}\right)(\nu+\omega)\right]\le2\exp\left[-c_{4}n\right].
\]

On the other hand, note that 
\[
\mathbb{E}\left[\left|\xi_{j}-\gamma_{j}\right|\right]\ge\sqrt{\frac{\left(\mathbb{E}\left[(\xi_{j}-\gamma_{j})^{2}\right]\right)^{3}}{\mathbb{E}\left[(\xi_{j}-\gamma_{j})^{4}\right]}}.
\]
 We bound the numerator and denominator. By sub-exponentiality, we
have $\mathbb{E}\left[\left(\xi_{j}-\gamma_{j}\right)^{4}\right]\le c_{5}(\nu+\omega)^{4}.$
On the other hand, note that 
\begin{align*}
& \;\quad\mathbb{E}\left(\xi_{j}-\gamma_{j}\right){}^{2}\\
 & =\mathbb{E}\left(\left\langle \B_{j},\Z\right\rangle -\left\langle \dd_{j},\z\right\rangle \right)^{2}\\
 & =\mathbb{E}\left\langle \B_{j},\Z\right\rangle ^{2}+\mathbb{E}\left\langle \dd_{j},\z\right\rangle ^{2}-2\mathbb{E}\left[\left\langle \B_{j},\Z\right\rangle \left\langle \dd_{j},\z\right\rangle \right]\\
 & =\mathbb{E}\left\langle \B_{j},\Z\right\rangle ^{2}+\mathbb{E}\left\langle \dd_{j}\dd_{j}^{\top},\z\z^{\top}\right\rangle -2\mathbb{E}\left[\left\langle \B_{j},\Z\right\rangle \left\langle e_{2j}\x_{2j}-e_{2j-1}\x_{2j-1},\z\right\rangle \right]\\
 & =\mathbb{E}\left\langle \B_{j},\Z\right\rangle ^{2}+\mathbb{E}\left\langle \dd_{j}\dd_{j}^{\top},\z\z^{\top}\right\rangle -2\mathbb{E}\left[e_{2j}\right]\mathbb{E}\left[\left\langle \B_{j},\Z\right\rangle \left\langle \x_{2j},\z\right\rangle \right]-2\mathbb{E}\left[e_{2j-1}\right]\mathbb{E}\left[\left\langle \B_{j-1},\Z\right\rangle \left\langle \x_{2j-1},\z\right\rangle \right]\\
 & =\mathbb{E}\left\langle \B_{j},\Z\right\rangle ^{2}+\mathbb{E}\left\langle \dd_{j}\dd_{j}^{\top},\z\z^{\top}\right\rangle ,
\end{align*}
where in the last equality we use the fact that $\left\{ e_{i}\right\} $
are independent of $\left\{ \x_{i}\right\} $ and $\mathbb{E}\left[e_{i}\right]=0$
for all $i$. We already know 
\[
\mathbb{E}\left\langle \B_{j},\Z\right\rangle ^{2}=\left\langle \mathbb{E}\left[\left\langle \B_{j},\Z\right\rangle \B_{j}\right],\Z\right\rangle =4\left\Vert \Z\right\Vert _{F}^{2}+2(\mu-3)\left\Vert \textrm{diag}\left(\Z\right)\right\Vert _{F}^{2}\ge2(\mu-1)\left\Vert \Z\right\Vert _{F}^{2}.
\]
Some calculation shows that 
\[
\mathbb{E}\left\langle \dd_{j}\dd_{j}^{\top},\z\z^{\top}\right\rangle =\left\langle \mathbb{E}\left[e_{2j}^{2}\x_{2j}\x_{2j}^{\top}+e_{2j}^{2}\x_{2j}\x_{2j}^{\top}\right],\z\z^{\top}\right\rangle =2\left\langle \I,\z\z^{\top}\right\rangle =2\left\Vert \z\right\Vert ^{2}.
\]
It follows that 
\[
\mathbb{E}\left(\xi_{j}-\gamma_{j}\right){}^{2}\ge2(\mu-1)\left\Vert \Z\right\Vert _{F}^{2}+2\left\Vert \z\right\Vert ^{2}\ge c_{6}\left(\nu^{2}+\omega^{2}\right),
\]
where the inequality holds when $\mu>1$. We therefore obtain
\[
\mathbb{E}\left[\left|\xi_{j}-\gamma_{j}\right|\right]\ge c_{7}\frac{\sqrt{\left(\nu^{2}+\omega^{2}\right)^{3}}}{(\nu+\omega)^{2}}\ge c_{8}(\nu+\omega).
\]
Substituting back to (\ref{eq:concentrate2}), we get
\[
\mathbb{P}\left[\left\Vert \BB\Z-\D\z\right\Vert _{1}\le\left(c_{8}-\frac{1}{c_{3}}\right)(\nu+\omega)\right]\le2\exp\left[-c_{4}n\right].
\]

To complete the proof of the lemma, we use an $ \epsilon $-net argument.
Define the set $$\mathcal{S}_{r}:=\left\{ (\Z,\z)\in\mathbb{R}^{p\times p}\times\mathbb{R}^{p}:\text{rank}(\Z)\le r,\left\Vert \Z\right\Vert _{F}^{2}+\left\Vert \z\right\Vert _{2}^{2}=1\right\}. $$
We need the following lemma, which is proved in Appendix \ref{sub:proof_covering}.
\begin{lemma}
\label{lem:covering}For each $\epsilon>0$ and $r\ge1$, there exists
a set $\mathcal{N}_{r}(\epsilon)$ with $\left|\mathcal{N}_{r}(\epsilon)\right|\le\left(\frac{40}{\epsilon}\right)^{10pr}$
which is an $\epsilon$-covering of $\mathcal{S}_{r}$, meaning that for all $\left(\Z,\z\right)\in\mathcal{S}_{r}$, there exists $\left(\tilde{\Z},\tilde{\z}\right)\in\mathcal{N}_{r}(\epsilon)$ such that
\[
\sqrt{\left\Vert \tilde{\Z}-\Z\right\Vert _{F}^{2}+\left\Vert \tilde{\z}-\z\right\Vert _{2}^{2}}\le\epsilon.
\]

\end{lemma}
Note that $\frac{1}{\sqrt{2}}\left(\left\Vert \Z\right\Vert _{F}+\left\Vert \z\right\Vert _{2}\right)\le\sqrt{\left\Vert \Z\right\Vert _{F}^{2}+\left\Vert \z\right\Vert _{2}^{2}}\le\left\Vert \Z\right\Vert _{F}+\left\Vert \z\right\Vert _{2}$
for all $\Z$ and $\z$. Therefore, up to a change of constant, it
suffices to prove Lemma~\ref{lem:RIP2} for all $\left(\Z,\z\right)$ in $\mathcal{S}_{r}$.
By the union bound and Lemma~\ref{lem:covering}, we have
\[
\mathbb{P}\left(\max_{\left(\tilde{\Z},\tilde{\z}\right)\in\mathcal{N}_{r}(\epsilon)}\left\Vert \BB\tilde{\Z}-\D\tilde{\z}\right\Vert _{1}\le2\left(c_{2}+\frac{1}{c_{3}}\right)\right)\ge1-\left|\mathcal{N}_{r}(\epsilon)\right|\cdot\exp\left(-c_{4}n\right)\ge1-\exp(-c_{4}n/2),
\]
when $n\ge\left(2/c_{4}\right)\cdot10pr\log(40/\epsilon)$. On this event, we have
\begin{align*}
\bar{M} & :=\sup_{(\Z,\z)\in\mathcal{S}_{r}}\left\Vert \BB\Z-\D\z\right\Vert _{1}\\
 & \le\max_{\left(\tilde{\Z},\tilde{\z}\right)\in\mathcal{N}_{r}(\epsilon)}\left\Vert \BB\tilde{\Z}-\D\tilde{\z}\right\Vert _{1}+\sup_{(\Z,\z)\in\mathcal{S}_{r}}\left\Vert \BB(\Z-\tilde{\Z})-\D(\z-\tilde{\z})\right\Vert _{1}\\
 & \le2\left(c_{2}+\frac{1}{c_{3}}\right)+\sup_{\Z\in S_{r}}\sqrt{\left\Vert \Z-\tilde{\Z}\right\Vert _{F}^{2}+\left\Vert \z-\tilde{\z}\right\Vert _{2}^{2}}\sup_{(\Z',\z')\in\mathcal{S}_{2r}}\left\Vert \BB\Z'-\D\z'\right\Vert _{1}\\
 & \le2\left(c_{2}+\frac{1}{c_{3}}\right)+\epsilon\sup_{(\Z',\z')\in\mathcal{S}_{2r}}\left\Vert \BB\Z'-\D\z'\right\Vert _{1}.
\end{align*}
Note that for $(\Z',\z')\in\mathcal{S}_{2r}$, we can write $\Z'=\Z'_{1}+\Z'_{2}$
such that $\Z'_{1},\Z'_{2}$ has rank $r$ and $1=\left\Vert \Z'\right\Vert _{F}\ge\max\left\{ \left\Vert \Z'_{1}\right\Vert _{F},\left\Vert \Z'_{2}\right\Vert _{F}\right\} $.
So 
\begin{equation}
\sup_{(\Z',\z')\in\mathcal{S}_{2r}}\left\Vert \BB\Z'-\D\z'\right\Vert _{1}\le\sup_{\Z'\in\mathcal{S}_{2r}}\left\Vert \BB\Z'_{1}-\D\z'\right\Vert _{1}+\sup_{\Z'\in\mathcal{S}_{2r}}\left\Vert \BB\Z'_{2}\right\Vert _{1}\le2\bar{M}.\label{eq:2r_to_r}
\end{equation}
Combining the last two display equations and choosing $\epsilon=\frac{1}{4}$,
we obtain 
\[
\bar{M}\le\bar{\delta}:=\frac{2}{1-2\epsilon}\left(c_{2}+\frac{1}{c_{3}}\right),
\]
with probability at least $1-\exp(-c_{9}n)$. Note that $\bar{\delta}$
is a constant independent of $p$ and $r$ (but might depend on $\mu:=\mathbb{E}\left[\left(\x_{i}\right)_{l}^{4}\right]$).

For a possibly different $\epsilon'$, we have
\[
\inf_{(\Z,\z)\in\mathcal{S}_{r}}\left\Vert \BB\Z-\D\z\right\Vert _{1}\ge\min_{\left(\tilde{\Z},\tilde{\z}\right)\in\mathcal{N}_{r}(\epsilon)}\left\Vert \BB\tilde{\Z}-\tilde{\z}\right\Vert _{1}-\sup_{(\Z,\z)\in\mathcal{S}_{r}}\left\Vert \BB(\Z-\tilde{\Z})-\D(\z-\tilde{\z})\right\Vert _{1}.
\]
By the union bound, we have 
\begin{align*}
\mathbb{P}\left(\min_{\left(\tilde{\Z},\tilde{\z}\right)\in\mathcal{N}_{r}(\epsilon)}\left\Vert \BB\tilde{\Z}-\tilde{\z}\right\Vert _{1}\ge\left(c_{7}-\frac{1}{c_{3}}\right)\right)
&\ge1-\exp\left(-c_{4}n+10pr\log(40/\epsilon')\right)\\
&\ge1-\exp(-c_{4}n/2),
\end{align*}
provided $n\ge(2/c_{4})\cdot10pr\log(40/\epsilon')$. On this event,
we have
\[
\inf_{(\Z,\z)\in\mathcal{S}_{r}}\left\Vert \BB\Z-\D\z\right\Vert _{1}\overset{(a)}{\ge}\left(c_{7}-\frac{1}{c_{3}}\right)-2\epsilon'\bar{M}\overset{(b)}{\ge}\left(c_{7}-\frac{1}{c_{3}}\right)-2\epsilon'\bar{\delta},
\]
where (a) follows from (\ref{eq:2r_to_r}) and (b) follows from the
the upper-bound on $\bar{M}$ we just established. We complete the
proof by choosing $\epsilon'$ to be a sufficiently small constant
such that $\underbar{\ensuremath{\delta}}:=\left(c_{7}-\frac{1}{c_{3}}\right)-2\epsilon'\bar{\delta}>0$.

\subsubsection{Proof of Lemma \ref{lem:covering}\label{sub:proof_covering}}
\proof
Define the sphere 
$$
\mathcal{T}_{r}(b):=\left\{ \Z\in\mathbb{R}^{p\times p}:\text{rank}(\Z)\le r,\left\Vert \Z\right\Vert _{F}=b\right\}.
$$
Let $\mathcal{M}_{r}(\epsilon/2,1)$ be the smallest $\epsilon/2$-net of $\mathcal{T}'_{r}(1)$. We know $\left|\mathcal{M}_{r}(\epsilon/2,1)\right|\le\left(\frac{20}{\epsilon}\right)^{6pr}$ by~\cite{candes2011oracle}. For any $0\le b\le1$, we know $\mathcal{M}_{r}(\epsilon/2,b):=\left\{ b\Z:\Z\in\mathcal{M}(\epsilon/2,1)\right\}$ is an $\epsilon/2$-net of $\mathcal{T}'_{r}(b)$, with $\left|\mathcal{M}_{r}(\epsilon/2,b)\right|=\left|\mathcal{M}_{r}(\epsilon/2,1)\right|\le\left(\frac{20}{\epsilon}\right)^{6pr}$.
Let $k:=\left\lfloor 2/\epsilon\right\rfloor \le2/\epsilon$. Consider
the set $\bar{\mathcal{M}}_{r}(\epsilon)=\left\{ 0\right\} \cup\bigcup_{i=1}^{k}\mathcal{M}_{r}(\epsilon/2,i\epsilon/2)$.
We claim that $\bar{\mathcal{M}}_{r}(\epsilon)$ is an $\epsilon$-net
of the ball $\bar{\mathcal{T}}_{r}:=\left\{ \Z\in\mathbb{R}^{p\times p}:\text{rank}(\Z)\le r,\left\Vert \Z\right\Vert _{F}\le1\right\} $,
with the additional property that every $\Z$'s nearest neighbor $\tilde{\Z}$
in $\bar{\mathcal{M}}_{r}(\epsilon)$ satisfies $\left\Vert \tilde{\Z}\right\Vert _{F}\le\left\Vert \Z\right\Vert _{F}$.
To see this, note that for any $\Z\in\bar{\mathcal{T}}(r)$, there
must be some $0\le i\le k$ such that $i\epsilon/2\le\left\Vert \Z\right\Vert _{F}\le(i+1)\epsilon/2$.
Define$\Z':=i\epsilon\Z/(2\left\Vert \Z\right\Vert _{F})$, which
is in $\mathcal{T}_{r}(i\epsilon/2)$. We choose $\tilde{\Z}$ to
be the point in $\mathcal{M}_{r}(\epsilon/2,i\epsilon/2)$ that is
closest to $\Z'$. We have
\[
\left\Vert \tilde{\Z}-\Z\right\Vert _{F}\le\left\Vert \tilde{\Z}-\Z'\right\Vert _{F}+\left\Vert \Z'-\Z\right\Vert _{F}\le\epsilon/2+\left(\left\Vert \Z\right\Vert _{F}-i\epsilon/2\right)\le\epsilon,
\]
and $\left\Vert \tilde{\Z}\right\Vert _{F}=i\epsilon/2\le\left\Vert \Z\right\Vert _{F}.$
The cardinality of $\bar{\mathcal{M}}_{r}(\epsilon)$ satisfies 
$$
\left|\bar{\mathcal{M}}_{r}(\epsilon)\right|\le1+\sum_{i=1}^{k}\left|\mathcal{M}_{r}(\epsilon/2,k\epsilon/2)\right|\le1+\frac{1}{\epsilon}\left(\frac{20}{\epsilon}\right)^{6pr}\le\left(\frac{20}{\epsilon}\right)^{7pr}.
$$

We know that the smallest $\epsilon/2$-net $\mathcal{M}'(\epsilon/2,1)$
of the sphere $\mathcal{T}'(1):=\left\{ \z\in\mathbb{R}^{p}:\left\Vert \z\right\Vert =1\right\} $
satisfies $\left|\mathcal{M}'(\epsilon/2,1)\right|\le\left(\frac{20}{\epsilon}\right)^{p}$.
It follows from an argument similar to above that there is an $\epsilon$-covering$\bar{\mathcal{M}}'(\epsilon)$
of the ball $\bar{\mathcal{T}}':=\left\{ \z\in\mathbb{R}^{p}:\left\Vert \z\right\Vert \le1\right\} $
with cardinality $\left|\mathcal{\bar{M}}'(\epsilon)\right|\le\left(\frac{20}{\epsilon}\right)^{2p}$
and the property that every $\z$'s nearest neighbor $\tilde{\z}$
in $\bar{\mathcal{M}}'(\epsilon)$ satisfies $\left\Vert \tilde{\z}\right\Vert _{2}\le\left\Vert \z\right\Vert _{2}$. 

Define the ball $\bar{\mathcal{S}}_{r}:=\left\{ (\Z,\z)\in\mathbb{R}^{p\times p}\times\mathbb{R}^{p}:\text{rank}(\Z)\le r,\left\Vert \Z\right\Vert _{F}^{2}+\left\Vert \z\right\Vert _{2}^{2}\le1\right\} $.
We claim that $\bar{\mathcal{N}}_{r}(\sqrt{2}\epsilon):=\left(\mathcal{\bar{M}}_{r}(\epsilon)\times\mathcal{\bar{M}}'(\epsilon)\right)\cap\bar{\mathcal{S}}_{r}$
is an $\sqrt{2}\epsilon$-net of $\bar{\mathcal{S}}_{r}$. To see
this, for any $(\Z,\z)\in\bar{\mathcal{S}}_{r}\subset\bar{\mathcal{T}}(r)\times\bar{\mathcal{T}}'$,
we let $\tilde{\Z}$ ($\tilde{\z}$, resp.) be the point in $\bar{\mathcal{M}}_{r}(\epsilon)$
($\bar{\mathcal{M}}'(\epsilon)$, resp.) closest to $\Z$ ($\z$,
resp.) We have 
\[
\sqrt{\left\Vert \tilde{\Z}-\Z\right\Vert _{F}^{2}+\left\Vert \tilde{\z}-\z\right\Vert _{2}^{2}}\le\sqrt{\epsilon^{2}+\epsilon^{2}}=\sqrt{2}\epsilon,
\]
and $\left\Vert \tilde{\Z}\right\Vert _{F}^{2}+\left\Vert \tilde{\z}\right\Vert _{2}^{2}\le\left\Vert \Z\right\Vert _{F}^{2}+\left\Vert \z\right\Vert _{2}^{2}\le1$. 

Let $\mathcal{N}_{r}(\sqrt{2}\epsilon)$ be the projection of the
set $\bar{\mathcal{N}}_{r}(\sqrt{2}\epsilon)$ onto the sphere $\mathcal{S}_{r}$.
Since projection does not increase distance, we are guaranteed that
$\mathcal{N}_{r}(\sqrt{2}\epsilon)$ is an $\sqrt{2}\epsilon$-net
of $\mathcal{S}_{r}$. Moreover,
\[
\left|\mathcal{N}_{r}(\sqrt{2}\epsilon)\right|\le\left|\bar{\mathcal{N}}_{r}(\sqrt{2}\epsilon)\right|\le\left|\mathcal{\bar{M}}_{r}(\epsilon)\right|\times\left|\mathcal{\bar{M}}'(\epsilon)\right|\le\left(\frac{20}{\epsilon}\right)^{10pr}.
\]

\subsection{Proof of Lemma \ref{lem:truncation}}

Without loss of generality, we may assume $\sigma=1$. Set $L:=\sqrt{c\log n}$ for some
$c$ sufficiently large. For each $i\in[n]$, we define the event
$\mathcal{E}_{i}=\left\{ \left|e_{i}\right|\le L\right\} $ and the
truncated random variables 
\begin{align*}
\bar{e}_{i} & =e_{i}\mathbf{1}\left(\mathcal{E}_{i}\right),
\end{align*}
where $\mathbf{1}(\cdot)$ is the indicator function and $c$ is some
sufficiently large numeric constant. Let $m_i:=\mathbb{E}\left[e_{i}\mathbf{1}\left(\mathcal{E}_i^{c}\right)\right]$
and $s_i:=\sqrt{\mathbb{E}\left[e_{i}^{2}\mathbf{1}\left(\mathcal{E}_i^{c}\right)\right]}$.
WLOG we assume $m_i\ge0$.
Note that the following equation holds almost surely:
\[
e_{i}^{2}\mathbf{1}\left(\mathcal{E}_i^{c}\right)
=\left|e_{i}\right|\cdot\left|e_{i}\right|\mathbf{1}\left(\mathcal{E}_i^{c}\right)
\ge L\cdot\left|e_{i}\right|\mathbf{1}\left(\mathcal{E}_i^{c}\right)
\ge L\cdot e_{i}\mathbf{1}\left(\mathcal{E}_i^{c}\right).
\]
Taking the expectation of both sides gives $s_i^{2}\ge Lm_i$. We further define
\[
\tilde{e}_{i}:=\bar{e}_{i}+L\epsilon_{i}^{+}-L\epsilon_{i}^{-},
\]
where $\epsilon_{i}^{+}$ and $\epsilon_{i}^{-}$ are independent random variables  distributed as Ber($\nu_{i}^{+}$) and Ber($\nu_{i}^{-}$), respectively, with
\[
\nu_{i}^{+}:=\frac{1}{2}\left(\frac{m_i}{L}+\frac{s_i^{2}}{L^{2}}\right),\quad\nu_{i}^{-}:=\frac{1}{2}\left(-\frac{m_i}{L}+\frac{s_i^{2}}{L^{2}}\right).
\]
Note that $ m_i\ge 0 $ and $s_i^{2}\ge Lm_i$ implies that $\nu_{i}^{+},\nu_{i}^{-}\ge0$.
We show below that $\nu_{i}^{+},\nu_{i}^{-}\le1$ so the random
variables $\epsilon_{i}^{+}$ and $\epsilon_{i}^{-}$ are well-defined.

With this setup, we now characterize the distribution of $\tilde{e}_{i}$.
Note that
\begin{align*}
\mathbb{E}\left[L\epsilon_{i}^{+}-L\epsilon_{i}^{-}\right] & =m_i,\\
\mathbb{E}\left[(L\epsilon_{i}^{+})^{2}+(L\epsilon_{i}^{-})^{2}\right] & =s_i^{2},
\end{align*}
which means 
\begin{align*}
\mathbb{E}\left[\tilde{e}_{i}\right] & =\mathbb{E}\left[\bar{e}_{i}\right]+\mathbb{E}\left[e_{i}\mathbf{1}\left(\mathcal{E}_i^{c}\right)\right]=\mathbb{E}\left[e_{i}\right]=0.\\
Var\left[\tilde{e}_{i}^{2}\right] & =\mathbb{E}\left[\bar{e}_{i}^{2}\right]+\mathbb{E}\left[e_{i}^{2}\mathbf{1}\left(\mathcal{E}_i^{c}\right)\right]=\mathbb{E}\left[e_{i}^{2}\right]=1.
\end{align*}
Moreover, $\tilde{e}_{i}$ is bounded by $3L$ almost surely,
which means it is sub-Gaussian with sub-Gaussian norm at most $ 3L $. Also note that 
\begin{align*}
m_i & \le\mathbb{E}\left[\left|e_{i}\mathbf{1}\left(\mathcal{E}_i^{c}\right)\right|\right]\\
 & =\int_{0}^{\infty}\mathbb{P}\left(\left|e_{i}\mathbf{1}\left(\mathcal{E}_i^{c}\right)\right|\ge t\right)dt\\
 & =L\cdot \mathbb{P}(|e_i|\ge L) + \int_{L}^{\infty}\mathbb{P}\left(\left|e_{i}\right|\ge t\right)dt\\
 & \le \sqrt{c\log n} \frac{1}{n^{c_1}} +  \int_{L}^{\infty}e^{1-t^{2}}dt\le\frac{4}{n^{c_2}}
\end{align*}
for some large constant $ c_1$ and $ c_2 $ by sub-Gaussianity of $ e_i $.
A similar calculation gives 
\[
s_i^{2}=\mathbb{E}\left[e_{i}^{2}\mathbf{1}\left(\mathcal{E}^{c}\right)\right]\lesssim\frac{1}{n^{c_2}}.
\]
This implies $\nu_{i}^{+},\nu_{i}^{-}\lesssim\frac{1}{n^{c_2}}$, or
equivalently $L\epsilon_{i}^{+}-L\epsilon_{i}^{-}=0$ w.h.p. We also
have $\bar{e}_{i}=e_{i}$ w.h.p. by sub-Gaussianity of $e_{i}$. It
follows that $\tilde{e}_{i}=\bar{e}_{i}+L\epsilon_{i}^{+}-L\epsilon_{i}^{-}=e_{i}$
w.h.p. Moreover, $ \tilde{e}_i $ and $ e_i $ have the same mean and variance.

We define the variables $\left\{ (\tilde{\x}{}_{i})_{l},i\in[n],l\in[p]\right\} $
in a similar manner. Each $\left(\tilde{\x}_{i}\right)_{l}$ is sub-Gaussian,
bounded by $L$ a.s., has mean $0$ and variance $1$, and equals
$\left(\x_{i}\right)_{l}$ w.h.p. 

Now suppose the conclusion of Theorem \ref{thm:rand} holds w.h.p.
for the program~\eqref{eq:L1_regularized_LS} with $\left\{ \left(\tilde{\x}_{i},\tilde{y}_{i}\right)\right\} $
as the input, where $\tilde{y}_{i}=\tilde{\x}_{i}^{\top}\bbeta_{b}^{*}+\tilde{e}_{i}$
for all $i\in\mathcal{I}_{b}$ and $b=1,2$. We know that $\e=\tilde{\e}$
and $\x_{i}=\tilde{\x}_{i},\forall i$ with high probability. On this
event, the program above is identical to the original program with
$\left\{ \left(\x_{i},y_{i}\right)\right\} $ as the input. Therefore,
the conclusion of the theorem also holds w.h.p. for the original program.

\subsection{Proof of Lemma \ref{lem:Gaussian_S1}}
\proof
We need to bound
\[
S_{1,1}=2\sum_{b}\left\Vert \sum_{i=1}^{n_{b}}e_{b,i}\x_{b,i}\x_{b,i}^{\top}\cdot\x_{b,i}^{\top}\left(\bbeta_{b}^{*}-\bbeta_{-b}^{*}\right)\right\Vert,
\]
where $\bbeta_{b}^{*}-\bbeta_{-b}^{*}$ is supported on the first coordinate. Because $n_{1}\asymp n_{2}\asymp n$ and $\left\{ (e_{b,i},\x_{b,i})\right\} $ are identically distributed, it suffices to prove w.h.p.
\begin{equation}
\left\Vert \E\right\Vert :=\left\Vert \sum_{i=1}^{n}e_{i}\x_{i}\x_{i}^{\top}\cdot\x_{i}^{\top}\ddelta_{1}^{*}\right\Vert \lesssim\sigma\left\Vert \ddelta_{1}^{*}\right\Vert _{2}\sqrt{np}\log^{3}n.\label{eq:gaussian_S1_a}
\end{equation}

Let $\bar{\x}_{i}\in\mathbb{R}^{1}$ and $\underbar{\ensuremath{\x}}_{i}\in\mathbb{R}^{p-1}$ be the subvectors of $\x_{i}$ corresponding to the first and the last $p-1$ coordinates, respectively. We define $\bar{\ddelta}_{1}^{*}$ similarly; note that $\left\Vert \bar{\ddelta}_{1}^{*}\right\Vert =\left\Vert \ddelta_{1}^{*}\right\Vert .$

Note that $\E:=\sum_{i}e_{i}\x_{i}\x_{i}^{\top}\cdot\bar{\x}_{i}^{\top}\bar{\ddelta}_{1}^{*}$
due to the support of $\ddelta_{1}^{*}$. We partition $\E\in\mathbb{R}^{p\times p}$
as
\[
\E=\left[\begin{array}{cc}
\E_{1} & \E_{12}\\
\E_{12}^{\top} & \E_{2}
\end{array}\right],
\]
where $\E_{1}\in\mathbb{R}^{1\times1}$, $\E_{2}\in\mathbb{R}^{(p-1)\times(p-1)}$
and $\E_{12}\in\mathbb{R}^{1\times p}$. We have 
\[
\left\Vert \E\right\Vert \le\left\Vert \E_{1}\right\Vert +\left\Vert \E_{2}\right\Vert +2\left\Vert \E_{12}\right\Vert .
\]
We bound each term separately.

Consider $\E_{1}=\sum_{i}e_{i}\bar{\x}_{i}\bar{\x}_{i}^{\top}\cdot\bar{\x}_{i}^{\top}\bar{\ddelta}_{1}^{*}$.
We condition on $\left\{ \bar{\x}_{i}\right\} $. Note that $\left\Vert \bar{\x}_{i}\right\Vert _{2}\lesssim\sqrt{\log n}$
and $\left|\bar{\x}_{i}^{\top}\bar{\ddelta}_{1}^{*}\right|\lesssim\left\Vert \ddelta_{1}^{*}\right\Vert \sqrt{\log n}$
a.s. by boundedness of $\x_{i}$. Since $\{e_{i}\}$ are independent
of $\left\{ \bar{\x}_{i}\right\} $, we have
\[
\mathbb{P}\left[\left\Vert \E_{1}\right\Vert \lesssim\sigma\left\Vert \ddelta_{1}^{*}\right\Vert \sqrt{n}\log^{2}n\vert\left\{ \bar{\x}_{i}\right\} \right]\ge1-n^{-10},
\]
w.h.p. using Hoeffding's inequality. Integrating over $\left\{ \bar{\x}_{i}\right\} $
proves $\left\Vert \E_{1}\right\Vert \lesssim\sigma\left\Vert \ddelta_{1}^{*}\right\Vert \sqrt{n}\log^{2}n$,
w.h.p.

Consider $\E_{2}=\sum_{i}e_{i}\underbar{\ensuremath{\x}}_{i}\underbar{\ensuremath{\x}}_{i}^{\top}\cdot\bar{\x}_{i}^{\top}\bar{\ddelta}_{1}^{*}$.
We condition on the event $\mathcal{F}:=\left\{ \forall i:\left|\bar{\x}_{i}^{\top}\bar{\ddelta}_{1}^{*}\right|\lesssim\left\Vert \ddelta_{1}^{*}\right\Vert \sqrt{\log n}\right\} $,
which occurs with high probability and is independent of $e_{i}$
and $\underbar{\ensuremath{\x}}_{i}$. We shall apply the matrix Bernstein inequality~\cite{tropp2010matrixmtg}; to this end, we compute:
\[
\left\Vert e_{i}\underbar{\ensuremath{\x}}_{i}\underbar{\ensuremath{\x}}_{i}^{\top}\cdot\bar{\x}_{i}^{\top}\bar{\ddelta}_{1}^{*}\right\Vert \lesssim\sigma p\left\Vert \ddelta_{1}^{*}\right\Vert \log^{2}n, \quad\text{a.s.}
\]
by boundedness, and
\[
\left\Vert \sum_{i}\mathbb{E}e_{i}^{2}\left(\underbar{\ensuremath{\x}}_{i}\underbar{\ensuremath{\x}}_{i}^{\top}\right)^{2}\cdot\left(\bar{\x}_{i}^{\top}\bar{\ddelta}_{1}^{*}\right)^{2}\right\Vert \le n\sigma^{2}\max_{i}\left|\bar{\x}_{i}^{\top}\bar{\ddelta}_{1}^{*}\right|^{2}\left\Vert \mathbb{E}\left(\underbar{\ensuremath{\x}}_{i}\underbar{\ensuremath{\x}}_{i}^{\top}\right)^{2}\right\Vert \le np\sigma^{2}\left\Vert \ddelta_{1}^{*}\right\Vert ^{2}\log n.
\]
Applying the Matrix Bernstein inequality then gives
\[
\left\Vert \E_{2}\right\Vert \lesssim\sigma\left\Vert \ddelta_{1}^{*}\right\Vert \left(p+\sqrt{np}\right)\log^{2}n\le\sigma\left\Vert \ddelta_{1}^{*}\right\Vert \sqrt{np}\log^{3}n,
\]
w.h.p., where we use $n\gtrsim p$ in the last inequality.

Consider $\E_{12}=\sum_{i}e_{i}\bar{\x}_{i}\underbar{\ensuremath{\x}}_{i}^{\top}\cdot\bar{\x}_{i}^{\top}\bar{\ddelta}_{1}^{*}$.
We again condition on the event $\mathcal{F}$ and use the matrix Bernstein inequality. Observe that
\[
\left\Vert e_{i}\bar{\x}_{i}\underbar{\ensuremath{\x}}_{i}^{\top}\cdot\bar{\x}_{i}^{\top}\bar{\ddelta}_{1}^{*}\right\Vert \lesssim\sigma\sqrt{p}\left\Vert \ddelta_{1}^{*}\right\Vert \log^{2}n, \quad\textrm{a.s.}
\]
by boundedness, and
\begin{align*}
\left\Vert \sum_{i}\mathbb{E}e_{i}^{2}\left(\bar{\x}_{i}^{\top}\bar{\ddelta}_{b}\right)^{2}\left(\underbar{\ensuremath{\x}}_{i}\bar{\x}_{i}^{\top}\right)\left(\bar{\x}_{i}\underbar{\ensuremath{\x}}_{i}^{\top}\right)\right\Vert  & \le n\sigma^{2}\max_{i}\left|\bar{\x}_{i}^{\top}\bar{\ddelta}_{1}^{*}\right|^{2}\left\Vert \bar{\x}_{i}\right\Vert ^{2}\left\Vert \mathbb{E}\underbar{\ensuremath{\x}}_{i}\underbar{\ensuremath{\x}}_{i}^{\top}\right\Vert \lesssim n\sigma^{2}\left\Vert \ddelta_{1}^{*}\right\Vert ^{2}\log^{2}n\\
\left\Vert \sum_{i}\mathbb{E}e_{i}^{2}\left(\bar{\x}_{i}^{\top}\bar{\ddelta}_{b}\right)^{2}\left(\bar{\x}_{i}\underbar{\ensuremath{\x}}_{i}^{\top}\right)\left(\underbar{\ensuremath{\x}}_{i}\bar{\x}_{i}^{\top}\right)\right\Vert  & \le n\sigma^{2}\max_{i}\left|\bar{\x}_{i}^{\top}\bar{\ddelta}_{1}^{*}\right|^{2}\left\Vert \bar{\x}_{i}\bar{\x}_{i}^{\top}\right\Vert \mathbb{E}\left[\underbar{\ensuremath{\x}}_{i}^{\top}\underbar{\ensuremath{\x}}_{i}\right]\lesssim np\sigma^{2}\left\Vert \ddelta_{1}^{*}\right\Vert ^{2}\log^{2}n.
\end{align*}
Applying the Matrix Bernstein inequality then gives
\[
\left\Vert \E_{12}\right\Vert \lesssim\sigma\left\Vert \ddelta_{1}^{*}\right\Vert \sqrt{np}\log^{3}n.
\]
Combining these bounds on $\left\Vert \E_{i}\right\Vert $, $i=1,2,3$, we conclude that (\ref{eq:gaussian_S1_a}) holds w.h.p., which completes the proves of the lemma.

\subsection{Proof of Remark~\ref{rem:metric}}

$\rho(\cdot,\cdot)$ satisfies the triangle inequality because 
\begin{align*}
 & \rho\left(\ttheta,\ttheta'\right)+\rho\left(\ttheta,\ttheta''\right)\\
= & \min\left\{ \left\Vert \bbeta_{1}-\bbeta_{1}'\right\Vert _{2}+\left\Vert \bbeta_{2}-\bbeta_{2}'\right\Vert _{2},\left\Vert \bbeta_{1}-\bbeta_{2}'\right\Vert _{2}+\left\Vert \bbeta_{2}-\bbeta_{1}'\right\Vert _{2}\right\} \\
 & +\min\left\{ \left\Vert \bbeta_{1}-\bbeta_{1}''\right\Vert _{2}+\left\Vert \bbeta_{2}-\bbeta_{2}''\right\Vert _{2},\left\Vert \bbeta_{1}-\bbeta_{2}''\right\Vert _{2}+\left\Vert \bbeta_{2}-\bbeta_{1}''\right\Vert _{2}\right\} \\
= & \min\left\{ \left\Vert \bbeta_{1}-\bbeta_{1}'\right\Vert _{2}+\left\Vert \bbeta_{2}-\bbeta_{2}'\right\Vert _{2}+\min\left\{ \left\Vert \bbeta_{1}-\bbeta_{1}''\right\Vert _{2}+\left\Vert \bbeta_{2}-\bbeta_{2}''\right\Vert _{2},\left\Vert \bbeta_{1}-\bbeta_{2}''\right\Vert _{2}+\left\Vert \bbeta_{2}-\bbeta_{1}''\right\Vert _{2}\right\} ,\right.\\
 & \;\left.\qquad\left\Vert \bbeta_{1}-\bbeta_{2}'\right\Vert _{2}+\left\Vert \bbeta_{2}-\bbeta_{1}'\right\Vert _{2}+\min\left\{ \left\Vert \bbeta_{1}-\bbeta_{1}''\right\Vert _{2}+\left\Vert \bbeta_{2}-\bbeta_{2}''\right\Vert _{2},\left\Vert \bbeta_{1}-\bbeta_{2}''\right\Vert _{2}+\left\Vert \bbeta_{2}-\bbeta_{1}''\right\Vert _{2}\right\} \right\} \\
\ge & \min\left\{ \min\left\{ \left\Vert \bbeta_{1}'-\bbeta_{1}''\right\Vert _{2}+\left\Vert \bbeta_{2}'-\bbeta_{2}''\right\Vert _{2},\left\Vert \bbeta_{1}'-\bbeta_{2}''\right\Vert _{2}+\left\Vert \bbeta_{2}'-\bbeta_{1}''\right\Vert _{2}\right\} ,\right.\\
 & \;\left.\qquad+\min\left\{ \left\Vert \bbeta_{2}'-\bbeta_{1}''\right\Vert _{2}+\left\Vert \bbeta_{1}'-\bbeta_{2}''\right\Vert _{2},\left\Vert \bbeta_{2}'-\bbeta_{2}''\right\Vert _{2}+\left\Vert \bbeta_{1}'-\bbeta_{1}''\right\Vert _{2}\right\} \right\} \\
= & \min\left\{ \left\Vert \bbeta_{1}'-\bbeta_{1}''\right\Vert _{2}+\left\Vert \bbeta_{2}'-\bbeta_{2}''\right\Vert _{2},\left\Vert \bbeta_{1}'-\bbeta_{2}''\right\Vert _{2}+\left\Vert \bbeta_{2}'-\bbeta_{1}''\right\Vert _{2}\right\} .
\end{align*}

\subsection{Proof of Lemma~\ref{lem:packing}}

We need a standard result on packing the unit hypercube.
\begin{lemma}
[Varshamov-Gilbert Bound, \cite{tsybakov2009nonparm}]\label{lem:varshamov}For
$p\ge15$, there exists a set $\Omega_{0}=\left\{ \xxi_{1},\ldots,\xxi_{M_{0}}\right\} \subset\left\{ 0,1\right\} ^{p-1}$
such that $M\ge2^{(p-1)/8}$ and $\left\Vert \xxi_{i}-\xxi_{j}\right\Vert _{0}\ge\frac{p-1}{8}$,
$\forall1\le i<j\le M_{0}$.
\end{lemma}
We claim that for $i\in[M_{0}]$, there is at most one $\bar{i}\in[M_{0}]$
with $\bar{i}\neq i$ such that 
\begin{equation}
\left\Vert \xxi_{i}-(-\xxi_{\bar{i}})\right\Vert _{0}<\frac{p-1}{16};\label{eq:negative_ball}
\end{equation}
otherwise if there are two distinct $i_{1}$, $i_{2}$ that satisfy
the above inequality, then they also satisfy 
\[
\left\Vert \xxi_{i_{1}}-\xxi_{i_{2}}\right\Vert _{0}\le\left\Vert \xxi_{i_{1}}-(-\xxi_{i})\right\Vert _{0}+\left\Vert \xxi_{i_{2}}-(-\xxi_{i})\right\Vert _{0}<\frac{p-1}{8},
\]
which contradicts Lemma~\ref{lem:varshamov}. Consequently, for each
$i\in[M_{0}]$, we use $\bar{i}$ to denote the unique index in $[M_{0}]$
that satisfies~(\ref{eq:negative_ball}) if such an index exists.

We construct a new set $\Omega\subseteq\Omega_{0}$ by deleting elements
from $\Omega_{0}$: Sequentially for $i=1,2,\ldots,M$, we delete
$\xxi_{\bar{i}}$ from $\Omega_{0}$ if $\bar{i}$ exists and both
$\xxi_{i}$ and $\xxi_{\bar{i}}$ have not been deleted. Note
that at most half of the elements in $\Omega$ are deleted in this
procedure. The resulting $\Omega=\left\{ \xxi_{1},\xxi_{2},\ldots,\xxi_{M}\right\} $
thus satisfies
\begin{align*}
M & \ge2^{(p-1)/16},\\
\min\left\{ \left\Vert \xxi_{i}-\xxi_{j}\right\Vert _{0},\left\Vert \xxi_{i}+\xxi_{j}\right\Vert _{0}\right\}  & \ge\frac{p-1}{16},\forall1\le i<j\le M.
\end{align*}

\subsection{Proof of Lemma \ref{lem:KL_bound}}
\begin{proof}
By rescaling, it suffices to prove the lemma for $\sigma=1$. Let
$\psi(x):=\frac{1}{\sqrt{2\pi}}\exp\left(-\frac{x{}^{2}}{2}\right)$
be the density function of the standard Normal distribution. The density
function of $\mathbb{Q}_{u}$ is 
\[
f_{u}(x)=\frac{1}{2}\psi(x-u)+\frac{1}{2}\psi(x+u),
\]
and the density of $\mathbb{Q}_{v}$ is given similarly. We compute
\begin{align}
D\left(\mathbb{Q}_{u}\Vert\mathbb{Q}_{v}\right) & =\int_{-\infty}^{\infty}f_{u}(x)\log\frac{f_{u}(x)}{f_{v}(x)}dx\nonumber \\
 & =\frac{1}{2}\int_{-\infty}^{\infty}\left[\psi\left(x-u\right)+\psi\left(x-u\right)\right]\log\left[\frac{\exp\left(-\frac{(x-u)^{2}}{2}\right)+\exp\left(-\frac{(x+u)^{2}}{2}\right)}{\exp\left(-\frac{(x-v)^{2}}{2}\right)+\exp\left(-\frac{(x+v)^{2}}{2}\right)}\right]dx\nonumber \\
 & =\frac{1}{2}\int_{-\infty}^{\infty}\left[\psi\left(x-u\right)+\psi\left(x-u\right)\right]\log\left[\frac{\exp\left(xu-\frac{u^{2}}{2}\right)+\exp\left(-xu-\frac{u^{2}}{2}\right)}{\exp\left(xv-\frac{v^{2}}{2}\right)+\exp\left(-xv-\frac{v^{2}}{2}\right)}\right]dx\nonumber \\
 & =\frac{1}{2}\int_{-\infty}^{\infty}\left[\psi\left(x-u\right)+\psi\left(x-u\right)\right]\log\left[\exp\left(-\frac{u^{2}-v^{2}}{2}\right)\frac{\exp\left(xu\right)+\exp\left(-xu\right)}{\exp\left(xv\right)+\exp\left(-xv\right)}\right]dx\nonumber \\
 & =\frac{1}{2}\int_{-\infty}^{\infty}\left[\psi\left(x-u\right)+\psi\left(x-u\right)\right]\left[-\frac{u^{2}-v^{2}}{2}+\log\frac{\cosh\left(xu\right)}{\cosh\left(xv\right)}\right]dx\nonumber \\
 & =-\frac{u^{2}-v^{2}}{2}+\frac{1}{2}\int_{-\infty}^{\infty}\left[\psi\left(x-u\right)+\psi\left(x-u\right)\right]\log\frac{\cosh\left(xu\right)}{\cosh\left(xv\right)}dx\label{eq:KL}
\end{align}
By Taylor's Theorem, the expansion of $\log\cosh(y)$ at the point
$a$ satisfies 
\[
\log\cosh(y)=\log\cosh(a)+(y-a)\tanh(a)+\frac{1}{2}(y-a)^{2}\sech^{2}(u)-\frac{1}{3}\left(y-a\right)^{3}\tanh(\xi)\sech^{2}(\xi)
\]
for some number $\xi$ between $a$ and $y$. Let $w:=\frac{u+v}{2}$.
We expand $\log\cosh(xu)$ and $\log\cosh(xv)$ separately using the
above equation, which gives that for some $\xi_{1}$ between $u$
and $w$, and some $\xi_{2}$ between $v$ and $w$, 
\begin{align}
 & \log\cosh\left(xu\right)-\log\cosh\left(xv\right)\nonumber \\
= & x(u-v)\tanh\left(xw\right)+\frac{x^{2}\left[\left(u-w\right){}^{2}-\left(v-w\right)^{2}\right]}{2}\sech^{2}\left(xw\right)\nonumber \\
 & -\frac{x^{3}\left(u-w\right)^{3}}{3}\tanh(x\xi_{1})\sech^{2}(x\xi_{1})+\frac{x^{3}\left(v-w\right)^{3}}{3}\tanh(x\xi_{2})\sech^{2}(x\xi_{2})\nonumber \\
= & x(u-v)\tanh\left(\frac{x(u+v)}{2}\right)+\frac{-x^{3}}{3}\left(\frac{u-v}{2}\right)^{3}\left[\tanh(x\xi_{1})\sech^{2}(x\xi_{1})+\tanh(x\xi_{2})\sech^{2}(x\xi_{2})\right],\label{eq:taylor}
\end{align}
where the last equality follows from $u-w=w-v=\frac{u-v}{2}$. We
bound the RHS of (\ref{eq:taylor}) by distinguishing two cases.

\paragraph*{Case 1: $u\ge v\ge0.$}

Because $\tanh(x\xi_{1})$ and $\tanh(x\xi_{2})$ have the same sign
as $x^{3}$, the second term in~(\ref{eq:taylor}) is negative. Moreover,
$ $we have $x\tanh\left(\frac{x(u+v)}{2}\right)\le x\cdot\frac{x(u+v)}{2}$
since $\frac{u+v}{2}\ge0$. It follows that
\begin{align*}
\log\cosh\left(xu\right)-\log\cosh\left(xv\right)\le & \frac{x^{2}(u-v)(u+v)}{2},
\end{align*}
Substituting back to~(\ref{eq:KL}), we obtain 
\begin{align*}
D\left(\mathbb{Q}_{u}\Vert\mathbb{Q}_{v}\right) & \le-\frac{u^{2}-v^{2}}{2}+\frac{1}{2}\int_{-\infty}^{\infty}\left[\psi(x-u)+\psi(x+u)\right]\cdot\frac{x^{2}(u^{2}-v^{2})}{2}dx\\
 & =-\frac{u^{2}-v^{2}}{2}+\frac{u^{2}-v^{2}}{2}(u^{2}+1) =\frac{u^{2}-v^{2}}{2}u^{2}.
\end{align*}

\paragraph*{Case 2: $v\ge u\ge0$.}

Let $h(y):=\tanh(y)-y+\frac{y^{3}}{3}.$ Taking the first order taylor's
expansion at the origin, we know that for any $y\ge0$ and some $0\le\xi\le y$,
$h(y)=-2\left(\tanh(\xi)\sech^{2}(\xi)-\xi\right)y^{2}\ge0$ since
$\tanh(\xi)\sech^{2}(\xi)\le\xi\cdot1^{2}$ for all $\xi\ge0$. This
means $\tanh(y)\ge y-\frac{y^{3}}{3},\forall y\ge0$. Since $u-v\le0$
and $\tanh(\cdot)$ is an odd function, we have
\[
x(u-v)\tanh\left(x(u+v)\right)\le x(u-v)\left[x(u+v)-\frac{1}{3}\left(xx(u+v)\right)^{3}\right].
\]
On the other hand, we have
\[
x\left[\tanh(x\xi_{1})\sech^{2}(x\xi_{1})+\tanh(x\xi_{2})\sech^{2}(x\xi_{2})\right]\overset{(a)}{\le}x(x\xi_{1}+x\xi_{2})\overset{(b)}{\le}x\cdot2vx,
\]
where (a) follows from $\sech^{2}(y)\le1$ and $0\le y\tanh(y)\le y^{2}$
for all $y$, and (b) follows from $\xi_{1},\xi_{2}\le v$ since $v\ge w\ge u\ge0$.
Combining the last two display equations with~(\ref{eq:taylor}),
we obtain
\[
\log\cosh\left(xu\right)-\log\cosh\left(xv\right)\le x(u-v)\left[\frac{x(u+v)}{2}-\frac{1}{3}\left(\frac{x(u+v)}{2}\right)^{3}\right]+\frac{x^{3}}{3}\left(\frac{v-u}{2}\right)^{3}\left(2vx\right).
\]
 when $a\le b$, we get
\begin{align*}
 & D\left(\mathbb{Q}_{u}\Vert\mathbb{Q}_{v}\right)\\
\le & -\frac{u^{2}-v^{2}}{2} \\
     &+\frac{1}{2}\int_{-\infty}^{\infty}\left[\psi(x\!-\!u)+\psi(x\!+\!v)\right]\cdot\left[\frac{u^{2}\!-\!v^{2}}{2}x^{2}+\frac{(v-u)}{3}\left(\frac{u+v}{2}\right)^{3}x^{4}+\frac{2v}{3}\left(\frac{v-u}{2}\right)^{3}x^{4}\right]dx\\
= & -\frac{u^{2}\!-\!v^{2}}{2}+\frac{u^{2}\!-\!v^{2}}{2}(u^{2}\!+\!1)+\left[\frac{(v\!-\!u)(u\!+\!v)^{3}}{48}+\frac{v\left(v\!-\!u\right)^{3}}{24}\right]\int_{-\infty}^{\infty}\left[\psi(x\!-\!u)+\psi(x\!+\!u)\right] x^{4}dx\\
= & \frac{u^{2}-v^{2}}{2}u^{2}+\left[\frac{(v-u)(u+v)^{3}}{24}+\frac{2v\left(v-u\right)^{3}}{24}\right]\left(u^{4}+6u^{2}+3\right)\\
\le & \frac{u^{2}-v^{2}}{2}u^{2}+(v-u)\left[\frac{(2v)^{3}}{24}+\frac{2v\left(v\right)^{2}}{24}\right]\left(u^{4}+6u^{2}+3\right)\\
\le & \frac{u^{2}-v^{2}}{2}u^{2}+(v-u)\frac{v{}^{3}}{2}\left(u^{4}+6u^{2}+3\right).
\end{align*}

Combining the two cases, we conclude that
\[
D\left(\mathbb{Q}_{u}\Vert\mathbb{Q}_{v}\right)\le\frac{u^{2}-v^{2}}{2}u^{2}+\frac{v^{3}\max\left\{ 0,v-u\right\} }{2}(u^{4}+6u^{2}+3).
\]

\end{proof}

\subsection{Proof of Lemma~\ref{lem:moment_bound}}

We recall that for any standard Gaussian variable $z\sim\mathcal{N}(0,1)$,
there exists a universal constant $\bar{c}$ such that $\mathbb{E}\left[\left|z\right|^{k}\right]\le\bar{c}$
for all $k\le16$. Now observe that $\mu:=\x^{\top}\aalpha\sim\mathcal{N}(0,\left\Vert \aalpha\right\Vert ^{2})$
and $\nu:=\x^{\top}\bbeta\sim\mathcal{N}(0,\left\Vert \bbeta\right\Vert ^{2})$.
Because $\x^{\top}\aalpha/\left\Vert \aalpha\right\Vert \sim\mathcal{N}\left(0,1\right)$
and $\x^{\top}\bbeta/\left\Vert \bbeta\right\Vert \sim\mathcal{N}\left(0,1\right)$,
it follows from the Cauchy-Schwarz inequality,
\[
\mathbb{E}\left[\left|\x^{\top}\aalpha\right|^{k}\left|\x^{\top}\bbeta\right|^{l}\right]\le\left\Vert \aalpha\right\Vert ^{k}\left\Vert \bbeta\right\Vert ^{l}\sqrt{\mathbb{E}\left|\frac{\x^{\top}\aalpha}{\left\Vert \aalpha\right\Vert }\right|^{2k}\mathbb{E}_{\X}\left|\frac{\x^{\top}\bbeta}{\left\Vert \bbeta\right\Vert }\right|^{2l}}\le\bar{c}\left\Vert \aalpha\right\Vert ^{k}\left\Vert \bbeta\right\Vert ^{l}.
\]
This proves the first inequality in the lemma.

For the second inequality in the lemma, note that 
\begin{align*}
\mathbb{E}\left[\left(\left|\x^{\top}\aalpha\right|^{2}-\left|\x^{\top}\bbeta\right|^{2}\right)\left|\x^{\top}\aalpha\right|^{2}\right] & =\mathbb{E}\left|\x^{\top}\aalpha\right|^{4}-\mathbb{E}\left|\x^{\top}\aalpha\right|^{2}\left|\x^{\top}\bbeta\right|^{2}\\
 & =3\left\Vert \aalpha\right\Vert ^{4}-\mathbb{E}\left|\x^{\top}\aalpha\right|^{2}\left|\x^{\top}\bbeta\right|^{2}.
\end{align*}
But
\begin{align}
\mathbb{E}\left|\x^{\top}\aalpha\right|^{2}\left|\x^{\top}\bbeta\right|^{2} & =\mathbb{E}\left(\alpha_{1}x_{1}+\cdots+\alpha_{p}x_{p}\right)^{2}\left(x_{1}\beta_{1}+\cdots+x_{p}\beta_{p}\right)^{2}\nonumber \\
 & =\mathbb{E}\sum_{i=1}^{p}x_{i}^{4}\alpha_{i}^{2}\beta_{i}^{2}+\mathbb{E}\sum_{i\neq j}x_{i}^{2}x_{j}^{2}\alpha_{i}^{2}\beta_{j}^{2}+2\mathbb{E}\sum_{i\neq j}x_{i}^{2}x_{j}^{2}\alpha_{i}\alpha_{j}\beta_{i}\beta_{j}\nonumber \\
 & =3\sum_{i=1}^{p}\alpha_{i}^{2}\beta_{i}^{2}+\sum_{i\neq j}\alpha_{i}^{2}\beta_{j}^{2}+2\sum_{i\neq j}\alpha_{i}\alpha_{j}\beta_{i}\beta_{j}\nonumber \\
 & =2\sum_{i=1}^{p}\alpha_{i}^{2}\beta_{i}^{2}+\sum_{i,j}\alpha_{i}^{2}\beta_{j}^{2}+2\sum_{i\neq j}\alpha_{i}\alpha_{j}\beta_{i}\beta_{j}\nonumber \\
 & =\left\Vert \aalpha\right\Vert ^{2}\left\Vert \bbeta\right\Vert ^{2}+2\sum_{i,j}\alpha_{i}\alpha_{j}\beta_{i}\beta_{j}\nonumber \\
 & =\left\Vert \aalpha\right\Vert ^{2}\left\Vert \bbeta\right\Vert ^{2}+2\left\langle \aalpha,\bbeta\right\rangle ^{2}.\label{eq:22}
\end{align}
It follows that
\begin{align*}
&\mathbb{E}\left[\left(\left|\x^{\top}\aalpha\right|^{2}-\left|\x^{\top}\bbeta\right|^{2}\right)\left|\x^{\top}\aalpha\right|^{2}\right]  =3\left\Vert \aalpha\right\Vert ^{4}-\left\Vert \aalpha\right\Vert ^{2}\left\Vert \bbeta\right\Vert ^{2}-2\left\langle \aalpha,\bbeta\right\rangle ^{2}\\
 = & 2\left\Vert \aalpha\right\Vert ^{4}-2\left\langle \aalpha,\bbeta\right\rangle ^{2}
  \le2\left\Vert \aalpha\right\Vert ^{4}+2\left(\left\Vert \aalpha\right\Vert ^{2}-\left\langle \aalpha,\bbeta\right\rangle \right)^{2}-2\left\langle \aalpha,\bbeta\right\rangle ^{2}\\
 = & 4\left\Vert \aalpha\right\Vert ^{4}-4\left\Vert \aalpha\right\Vert ^{2}\left\langle \aalpha,\bbeta\right\rangle 
 = 2\left\Vert \aalpha\right\Vert ^{2}\left(\left\Vert \aalpha\right\Vert ^{2}-2\left\langle \aalpha,\bbeta\right\rangle +\left\Vert \bbeta\right\Vert ^{2}\right)
 \le2\left\Vert \aalpha\right\Vert ^{2}\left\Vert \aalpha-\bbeta\right\Vert ^{2}.
\end{align*}

For the third inequality in the lemma, we use the equality~(\ref{eq:22})
to obtain 
\begin{align*}
& \mathbb{E}\left(\left\Vert \aalpha\right\Vert ^{2}-\left\Vert \bbeta\right\Vert ^{2}\right)^{2}  =\mathbb{E}\left\Vert \aalpha\right\Vert ^{4}-2\mathbb{E}\left\Vert \aalpha\right\Vert ^{2}\left\Vert \bbeta\right\Vert ^{2}+\mathbb{E}\left\Vert \bbeta\right\Vert ^{4}\\
  = & 6\left\Vert \aalpha\right\Vert ^{4}-2\left\Vert \aalpha\right\Vert ^{2}\left\Vert \bbeta\right\Vert ^{2}-4\left\langle \aalpha,\bbeta\right\rangle ^{2}.
 = 4\left\Vert \aalpha\right\Vert ^{4}-4\left\langle \aalpha,\bbeta\right\rangle ^{2}\\
 \le& 4\left\Vert \aalpha\right\Vert ^{4}-4\left\langle \aalpha,\bbeta\right\rangle ^{2}+2\left(\left\Vert \aalpha\right\Vert ^{2}-2\left\langle \aalpha,\bbeta\right\rangle \right)^{2}
 =5\left\Vert \aalpha\right\Vert ^{4}+4\left\langle \aalpha,\bbeta\right\rangle ^{2}-8\left\Vert \aalpha\right\Vert ^{2}\left\langle \aalpha,\bbeta\right\rangle \\
 \le & 4\left[\left\Vert \aalpha\right\Vert ^{4}+\left\langle \aalpha,\bbeta\right\rangle ^{2}-2\left\Vert \aalpha\right\Vert ^{2}\left\langle \aalpha,\bbeta\right\rangle \right]
 =\left(2\left\Vert \aalpha\right\Vert ^{2}-2\left\langle \aalpha,\bbeta\right\rangle \right)^{2}\\
 =& \left(\left\Vert \aalpha\right\Vert ^{2}-2\left\langle \aalpha,\bbeta\right\rangle +\left\Vert \bbeta\right\Vert ^{2}\right)^{2}
 =\left\Vert \aalpha-\bbeta\right\Vert ^{4}.
\end{align*}

\newpage
\bibliographystyle{plain}
\bibliography{mixed_regression}

\end{document}